\documentclass[twoside]{article}

\usepackage[accepted]{aistats2026}

\usepackage[utf8]{inputenc} 
\usepackage[T1]{fontenc}    
\usepackage{hyperref}       
\usepackage{url}            
\usepackage{booktabs}       
\usepackage{amsfonts}       
\usepackage{nicefrac}       
\usepackage{microtype}      
\usepackage{xcolor}         

\usepackage{amsmath}
\usepackage{enumerate} 

\usepackage{amsmath}
\usepackage{amssymb}
\usepackage{mathtools}
\usepackage{amsthm}

\usepackage{cancel}

\usepackage{tcolorbox}

\usepackage{chngcntr} 
\usepackage{etoc} 

\newcommand{\RVecInputValue}[2]{\mathbf{X}_{#1 #2}^{(0)}}

\newcommand{\RCompInputValue}[2]{\mathrm{X}_{#1 #2}^{(0)}}

\newcommand{\RVecDeepInputValue}[3]{\mathbf{X}_{#1 #2}^{(#3)}}

\newcommand{\RCompDeepInputValue}[3]{\mathrm{X}_{#1 #2}^{(#3)}}

\newcommand{\RVecHidNodeBAF}[2]{\mathbf{Z}_{#1}^{(#2)}}
\newcommand{\RCompHidNodeBAF}[2]{\mathrm{Z}_{#1}^{(#2)}}

\newcommand{\ArgHidNodeBAF}[0]{\mathrm{z}}

\newcommand{\RVecHidNodeAAF}[2]{\mathbf{Y}_{#1}^{(#2)}}
\newcommand{\RCompHidNodeAAF}[2]{\mathrm{Y}_{#1}^{(#2)}}

\newcommand{\RNNCompHidNodeAAF}[2]{\overbar{\mathrm{Y}}_{#1}^{(#2)}}

\newcommand{\ArgHidNodeAAF}[0]{\mathrm{y}}

\newcommand{\AFOperator}[1]{\phi \left({#1} \right)}
\newcommand{\AFOperatorNoArg}[0]{\phi }
\newcommand{\BNOperator}[1]{\mathrm{B} \left({#1} \right)}
\newcommand{\BNOperatorNoArg}[0]{\mathrm{B} }
\newcommand{\LNOperator}[1]{\mathrm{L} \left({#1} \right)}
\newcommand{\LNOperatorNoArg}[0]{\mathrm{L} }

\newcommand{\RVecHidNodeABN}[2]{\mathbf{B}^{(#1)}_{#2}}
\newcommand{\RCompHidNodeABN}[2]{\mathrm{B}^{(#1)}_{#2}}

\newcommand{\ArgHidNodeABN}[0]{\mathrm{b}}
\newcommand{\ArgHidNodeCusABN}[0]{\tilde{\mathrm{b}}}

\newcommand{\ArgHidNodeALN}[0]{\mathrm{l}}

\newcommand{\RVecHidNodeALN}[2]{\mathbf{L}^{(#1)}_{#2}}
\newcommand{\RCompHidNodeALN}[2]{\mathrm{L}^{(#1)}_{#2}}

\newcommand{\ArgHidNodeALF}[0]{l}

\newcommand{\BNMean}[2]{\hat{\mu}_{#1}^{(#2)}}
\newcommand{\BNVar}[2]{\left(\hat{\sigma}_{#1}^{(#2)} \right)^2}
\newcommand{\BNStd}[2]{\hat{\sigma}_{#1}^{(#2)}}
\newcommand{\BNCusMean}[2]{\tilde{\mu}_{#1}^{(#2)}}
\newcommand{\BNCusVar}[2]{\left(\tilde{\sigma}_{#1}^{(#2)} \right)^2}
\newcommand{\BNCusStd}[2]{\tilde{\sigma}_{#1}^{(#2)}}

\newcommand{\BatchSize}[0]{|B|}
\newcommand{\Batch}[0]{B}

\newcommand{\RCusCompHidNodeABN}[2]{\tilde{\mathrm{B}}^{(#1)}_{#2}}

\newcommand{\LNMean}[2]{\hat{\mu}_{#1}^{(#2)}}
\newcommand{\LNVar}[2]{\left(\hat{\sigma}_{#1}^{(#2)} \right)^2}
\newcommand{\LNStd}[2]{\hat{\sigma}_{#1}^{(#2)}}

\newcommand{\Layerperator}[1]{X \left({#1} \right)}

\newcommand{\ROutNode}[2]{\mathrm{O}_{#1 #2}}
\newcommand{\OutNode}[2]{\mathrm{o}_{#1 #2}}

\newcommand{\RMean}[1]{\mu_{#1}}
\newcommand{\Mean}[1]{m_{#1}}

\newcommand{\RStd}[1]{\sigma_{#1}}

\newcommand{\RProxyStd}[1]{\hat{\sigma}_{#1}}

\newcommand{\RCusProxyMean}[1]{\tilde{\mu}_{#1}}

\newcommand{\RCusProxyStd}[1]{\tilde{\sigma}_{#1}}

\newcommand{\RDiff}[1]{\Delta_{#1}}
\newcommand{\Diff}[1]{\delta_{#1}}

\newcommand{\Dataset}[0]{\chi}
\newcommand{\LabelValue}[0]{y}
\newcommand{\Weights}[2]{w^{(#1)}_{#2}}
\newcommand{\RClassFraction}[1]{G_{#1}}
\newcommand{\ClassFraction}[1]{g_{#1}}
\newcommand{\RMultiClassFraction}[2]{G_{#1}^{ #2 }}
\newcommand{\RMultiClassRankedFraction}[2]{G_{\tilde #1}^{ #2 }}

\newcommand{\CE}[2]{\mathbb{E}_{#1} \left( {#2}  \right)}

\newcommand{\Einput}[1]{\CE{\Dataset}{#1}} 

\newcommand{\overbar}[1]{\mkern 1.5mu\overline{\mkern-1.5mu#1\mkern-1.5mu}\mkern 1.5mu}

\newcommand{\CVar}[2]{\textrm{Var}_{#1} \left( {#2}  \right)}
\newcommand{\NumberClasses}[0]{N_{C}}
\newcommand{\LayerNumNodes}[1]{N_{#1}}

\newcommand{\WeightsMat}[1]{\boldsymbol{W}^{(#1)}}
\newcommand{\Heavyside}[1]{\Theta \left( #1 \right)}
\newcommand{\Dirac}[1]{\delta \left( #1 \right)}
\newcommand{\DatasetSize}[0]{D}
\newcommand{\WeightSet}[1]{\mathcal{W}^{#1}}
\newcommand{\Prob}[2]{\mathbb{P}^{#1} \left( #2 \right) }
\newcommand{\NormalDistr}[2]{\mathcal{N}\left( #1 , #2 \right) }
\newcommand{\NormalDens}[3]{\mathcal{N}\left( #1 ;  #2 , #3 \right) }

\newcommand{\Arch}[0]{\mathcal{A}}
\newcommand{\PreprData}[0]{\psi (\chi)}

\newcommand{\VarRatio}[0]{\gamma (\Arch, \PreprData)}
\newcommand{\LayerVarRatio}[1]{\gamma^{(#1)}}

\newcommand\Cdot{\boldsymbol{\cdot}}

\newcommand{\GB}[0]{\texttt{GB}}
\newcommand{\CIFAR}[0]{\texttt{C10}} 
\newcommand{\TinyImageNet}[0]{\texttt{TI}}
\newcommand{\CatVsDog}[0]{\texttt{C2-A}}
\newcommand{\VehVsAni}[0]{\texttt{C2-B}}

\newcommand{\MNIST}[0]{\texttt{E\&O}}

\newcommand{\MLPA}[0]{\texttt{SHLP}}
\newcommand{\MLPB}[0]{\texttt{MHLP}}
\newcommand{\ResNet}[0]{\texttt{ResNet34}}

\newcommand{\MLPmix}[0]{\texttt{MLP-mixer}}
\newcommand{\ResNetHundredOne}[0]{\texttt{ResNet101}}
\newcommand{\SwinL}[0]{\texttt{Swin-L}}

\newcommand{\pdf}[3]{f_{#1}^{#2} \left( {#3}  \right)}

\newcommand{\cdf}[3]{F_{#1}^{#2} \left( {#3}  \right)}

\newcommand{\Dmu}[0]{\Delta_{\mu}}
\newcommand{\Dsigma}[0]{\Delta_{\sigma^2}}
\newcommand{\eB}[0]{\epsilon_{\BatchSize}}

\DeclareMathOperator\erf{erf}

\newcommand{\LR}[1]{ \left( #1 \right)}

\newcommand{\StdKaiming}[0]{\sigma_w}

\tcbuselibrary{breakable} 

\theoremstyle{plain}
\newtheorem{theorem}{Theorem}[section]
\newtheorem{proposition}[theorem]{Proposition}
\newtheorem{lemma}[theorem]{Lemma}

\theoremstyle{definition}
\newtheorem{definition}[theorem]{Definition}

\theoremstyle{remark}

\newtcolorbox{mybox}[2][]
{
  colframe = #2!25,
  colback  = #2!25!white!25,
  left=1mm,
  top=1mm,
  #1,
  breakable
}

\newenvironment{theorembox}
   {\begin{mybox}{gray}\begin{theorem}}
   {\end{theorem}\end{mybox}}

\newenvironment{defbox}
   {\begin{mybox}{gray}\begin{definition}}
   {\end{definition}\end{mybox}}

\newenvironment{lemmabox}
   {\begin{mybox}{gray}\begin{lemma}}
   {\end{lemma}\end{mybox}}

\usepackage[round]{natbib}

\begin{document}

%
\runningtitle{Where You Place the Norm Matters}

%
\runningauthor{Francazi, Pinto, Lucchi \& Baity-Jesi}

\twocolumn[

\aistatstitle{{Where You Place the Norm Matters: \\
From Prejudiced to Neutral Initializations}}

\aistatsauthor{E. Francazi \And F. Pinto}

\aistatsaddress{
Physics Department\\
EPFL\\\texttt{emanuele.francazi@epfl.ch}  \And  Department of Statistics\\
Sapienza, University of Rome\\\texttt{pinto.1871045@studenti.uniroma1.it} } 

\aistatsauthor{A. Lucchi \And  M. Baity-Jesi }
\aistatsaddress{
Department of Mathematics and Computer Science\\
University of Basel\\\texttt{aurelien.lucchi@unibas.ch} \And  
SIAM Department\\
Eawag (ETH)\\\texttt{marco.baityjesi@eawag.ch} }

]

\begin{abstract}
Normalization layers were introduced to stabilize and accelerate training, yet their influence is critical already at initialization, where they shape signal propagation and output statistics before parameters adapt to data. In practice, both \emph{which} normalization to use and \emph{where} to place it are often chosen heuristically, despite the fact that these decisions can qualitatively alter a model’s behavior. We provide a theoretical characterization of how normalization \emph{choice} and \emph{placement} (Pre-Norm vs.\ Post-Norm) determine the distribution of class predictions at initialization, ranging from unbiased (Neutral) to highly concentrated (Prejudiced) regimes. We show that these architectural decisions induce systematic shifts in the initial prediction regime, thereby modulating subsequent learning dynamics. By linking normalization design directly to prediction statistics at initialization, our results offer principled guidance for more controlled and interpretable network design, including clarifying how widely used choices such as BatchNorm vs.\ LayerNorm and Pre-Norm vs.\ Post-Norm shape behavior from the outset of training.
\end{abstract}

\begin{figure*}[t]
    \centering
    \includegraphics[width=0.9\textwidth]{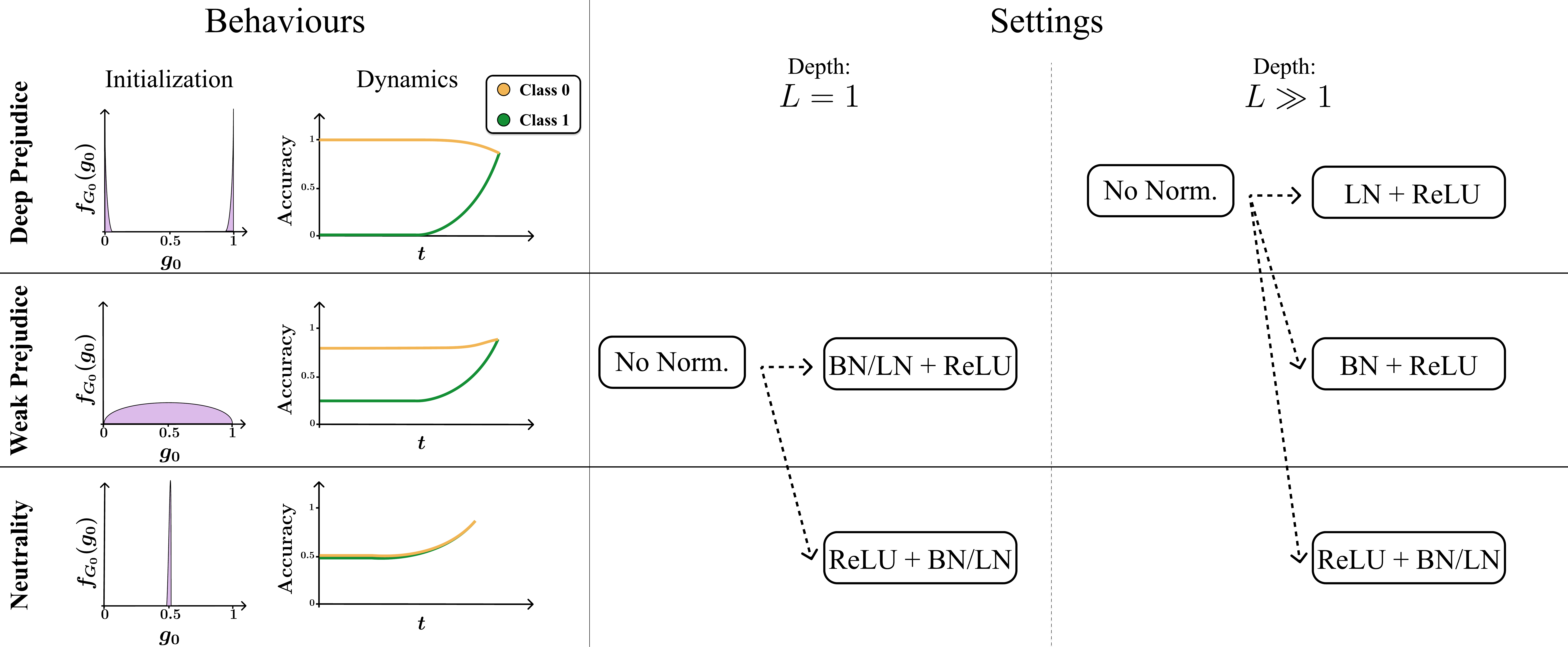}
\caption{Diagram summarizing the effects of normalization type and placement on IGB. The left part of the diagram introduces three distinct predictive behaviors at initialization—ranging from neutral to deeply prejudiced. Each regime is illustrated through its characteristic initialization state, represented by the distribution \(\pdf{\RClassFraction{0}}{}{\ClassFraction{0}}\), and its typical class-wise training dynamics (assuming class 0 as the initially favored class). On the right, all analyzed normalization configurations are grouped according to the behavioral regime they induce, comparing them to the baseline architecture without normalization (“No Norm.”, as analyzed in \cite{pmlr-v235-francazi24a}). Variants with LN and BN—examined in our work—are connected to the baseline by dashed arrows.}
    \label{fig:Thms_summary}
\end{figure*}

\section{Introduction}
Deep neural networks (DNNs) have achieved remarkable success across domains such as vision~\citep{zoph2018learning}, language~\citep{chen2017cnn}, and decision-making~\citep{silver2017mastering}, driven by architectural innovations like convolutions~\citep{lecun1989handwritten}, residual connections~\citep{he2016deep}, and attention mechanisms~\citep{vaswani2017attention}.
\newline
Among these, normalization layers represent a particularly impactful development. They were introduced to stabilize signal propagation and improve trainability, and this stabilizing role is especially consequential at initialization, when the network's outputs, and therefore its predictions,
are entirely determined by architectural choices and random weights. Despite their widespread empirical success, however, normalization layers are still often adopted through heuristic design patterns rather than grounded in a clear theoretical understanding of their implications. Two decisions are routinely made by experimentation: \emph{which} normalization to use (e.g., BatchNorm versus alternatives such as LayerNorm or RMSNorm) and \emph{where} to place it within a block (before or after the activation). These choices are ubiquitous in practice. For example, original Transformer architectures commonly applied LayerNorm in a post-activation/post-sublayer fashion, whereas some newer large language model architectures (e.g., LLaMA) adopt RMSNorm in a pre-activation/pre-sublayer configuration. Such shifts have largely been driven by empirical tuning, even though they can induce qualitatively different behaviors at initialization.
\newline
Understanding initialization behavior is particularly crucial in light of recent findings showing that architectural design choices can profoundly affect the behavior of untrained neural networks. Specifically, these choices have been shown to shape initial logit statistics, leading to qualitatively different prediction states \citep{pmlr-v235-francazi24a}. These initial states can be intuitively categorized into two types: \textbf{neutrality}, characterized by an unbiased initial condition where predictions are roughly equally distributed among classes; and \textbf{prejudice}, where predictions initially concentrate predominantly on one or a subset of classes. These initial conditions have significant practical implications, as they strongly affect subsequent training dynamics, including convergence speed, training stability, and fairness across classes~\citep{bassi2025normplacement}. In particular, when hyperparameter tuning relies on early training trajectories, as is common under finite compute budgets or in few-shot regimes, initialization-induced biases can steer selection toward configurations that perform well early, yet predominantly on a subset of classes. A deeper understanding of these effects is therefore essential, not only for advancing theory but also for making architectural design choices more controlled and interpretable.
\newline
In this work, we tackle the broad question: \textbf{How do normalization layers influence the initial prediction bias in DNNs?}
To build intuition, Fig.~\ref{fig:Thms_summary} visually summarizes the range of initialization behaviors explored in this work, and how they relate to different normalization configurations. It previews the core theoretical and empirical findings developed throughout the paper. \newline
Our theoretical analysis characterizes how normalization layers affect the initial state by altering the distribution \(\pdf{\RClassFraction{c}}{}{\ClassFraction{c}}\) of the fraction of data-points predicted as class \( c \) by the untrained model. \footnote{In the plots, we use $c=0$ as the reference class.}
We then examine the role of normalization placement relative to activation functions. Our results indicate that placing normalization layers after activations consistently promotes a neutral initialization state, mitigating initial prejudices. Conversely, the widely used heuristic of placing normalization before activations frequently results in prejudiced initialization regimes.
\newline
Our empirical experiments support these theoretical predictions. Simulations clearly demonstrate the substantial influence of these initial prediction regimes on training dynamics, particularly regarding convergence speed and fairness among classes. Additionally, experiments on widely used architectures (such as MLP-Mixer, ResNet) suggest that re-evaluating conventional normalization placement in light of initialization behavior can lead to meaningful gains in realistic scenarios.
\newline
By bridging architectural design with rigorous theoretical insights into initial prediction biases, our study advances theoretical understanding and provides practical design guidelines. In particular, it clarifies how everyday normalization decisions, such as those distinguishing Transformer-style and LLaMA-style designs, map to distinct and predictable initialization regimes. Because these regimes shape early learning dynamics, our framework provides a principled alternative to purely heuristic trial-and-error design.

\section{Background and Framework}\label{Sec:Backg&Fram}
In this section, we introduce the main notation and summarize the key concepts from prior work that are relevant to our analysis. For a more detailed discussion, we refer the reader to App.~\ref{sec:Notation} and App.~\ref{sec:Settings}.

\begin{figure*}[t!]
    \centering
\includegraphics[width=0.99\textwidth]{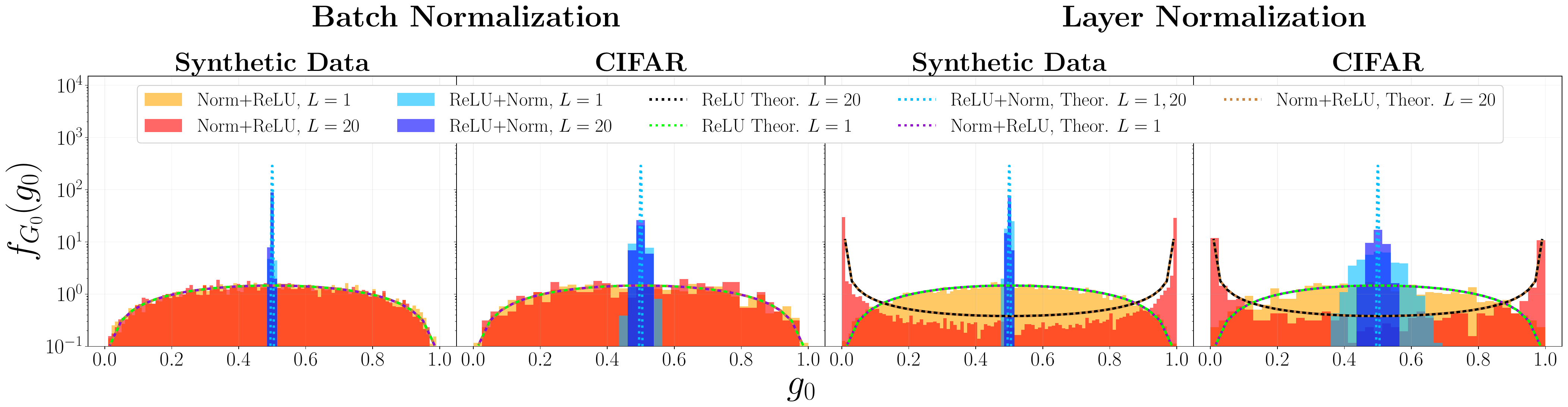}
    \caption{
Distribution of the initial guessing bias \(\pdf{\RClassFraction{0}}{}{\ClassFraction{0}}\) under different normalization settings and depths. \textbf{Left subplot:} BN; \textbf{Right subplot:} LN. For each normalization scheme, we report the distribution of \(\RClassFraction{0}\) when normalization is applied before or after the activation function. Within each subplot, we compare results from single hidden layer and deep MLP architectures. Each plot displays the empirical distribution for networks initialized on random unstructured inputs (left), the corresponding theoretical prediction, and the empirical distribution on structured real data (CIFAR10) (right). Post-activation normalization (ReLU + Norm.) consistently concentrates the distribution around 0.5, whereas pre-activation normalization (Norm. + ReLU) preserves or amplifies the prejudice depending on architecture depth and normalization type.
}
    \label{fig:static_Norm}
\end{figure*}

\paragraph{Initial State: Trainability and Fairness.}
The study of the initial state of DNNs has gained increasing attention in recent years, driven by a growing recognition of its impact on the subsequent training dynamics~\citep{he2015delving, mishkin2015all, boulila2024effective}. A first line of research has focused on how initialization choices affect model trainability, with particular emphasis on the propagation of signals and gradients through depth. In this context, mean field theory has provided a powerful framework to characterize initialization regimes, linking the correlation behaviour of forward signal propagation to gradient stability at initialization~\citep{schoenholz2016deep, xiao2018dynamical, noci2022signal}. A central insight is the existence of two phases—\emph{ordered} and \emph{chaotic}—distinguished by the correlation between output signals for different inputs as they propagate through layers. Only at the critical boundary between these phases, known as the \emph{edge of chaos} (EOC), gradients are stable ensuring trainability.\newline
A complementary line of research has more recently examined how architectural design shapes the network’s predictions at initialization. In particular, certain design patterns can induce a systematic predictive bias toward specific classes—a phenomenon termed \emph{Initial Guessing Bias} (IGB) \citep{pmlr-v235-francazi24a}. While previous mean field approaches analyze how a fixed input propagates across layers, IGB examines how a randomly initialized DNNs transform the entire dataset distribution as it propagates through the network.
Bias intuitively arises because the hidden layer representations of inputs drift away from the center of the activation space, causing the decision boundary—initialized near the origin—to no longer intersect the region occupied by these representations. As a result, the decision boundary initially assigns most or all datapoints to the same side, yielding prejudiced class predictions. We provide a more detailed intuitive discussion of this phenomenon in App.~\ref{sec:hint}.
Given an architecture \(\Arch\) and a pre-processed dataset \(\PreprData\), a natural and informative measure of this effect is the \emph{variance ratio} \(\VarRatio\), introduced by~\cite{pmlr-v235-francazi24a} as a proxy for the level of IGB. It compares the typical value of centers of the node output distributions to the variability of each node across data inputs. Large values of \(\VarRatio\) indicate that the output representations are consistently displaced from the origin, causing the decision boundary—initially centered—to miss the data cloud and produce prejudiced predictions. In contrast, small values of \(\VarRatio\) correspond to output representations that remain near the origin, allowing the decision boundary to intersect the data cloud and yield more balanced predictions. For a formal definition of \(  \VarRatio \) see App.~\ref{app:IGB_overview}
\newline
Notably,~\cite{bassi2025normplacement} established a connection between IGB and the gradient stability analysis of mean field theory \citep{schoenholz2016deep}, showing how prejudiced initial outputs relate to conditions for stable training. Interestingly, in certain architectures without normalization layers, IGB and gradient stability can coexist at the EOC: despite an initial prejudiced state, the prejudice is transient and gets reabsorbed during the early stages of training. However, this favorable coexistence breaks down in architectures with Batch Normalization (BN). As shown in~\cite{yang2019mean}, in BN networks, tuning the initialization parameters alone does not suffice to reach the EOC, suggesting that alternative architectural adjustments are needed to ensure stable learning while avoiding the risk of persistent prejudice in the predictions.
\paragraph{Normalization.}
\cite{bassi2025normplacement} formally established a connection between prejudiced initial states and increased correlation in signal propagation across network layers at initialization. This correlation is empirically demonstrated to be directly linked to convergence time~\cite{lubana2021beyond}, highlighting the ability of normalization techniques to reduce initial correlations and improve convergence speed.~\cite{lubana2021beyond} also pointed out significant differences between normalization methods (\textit{e.g.}, BN vs.\ Layer Normalization (LN)) in deep networks.\newline
A closely related set of design choices arises in Transformer-style residual blocks, where practitioners distinguish \emph{Post-Norm} (normalization applied after the sublayer and residual addition, as in the original Transformer) from \emph{Pre-Norm} (normalization applied to the sublayer input inside the residual path). Prior work has analyzed this Pre-Norm/Post-Norm choice through the lens of optimization stability: Post-Norm training can be sensitive at the start of optimization and often relies on careful learning-rate warm-up, whereas Pre-Norm improves gradient flow and enables more stable training in deep settings~\citep{xiong2020layer,nguyen2019transformers,liu2020understanding}. In parallel, architectural variants such as RMSNorm remove mean-centering to reduce overhead while retaining similar normalization effects in practice, and have been studied as efficient alternatives in pre-normalized Transformers~\citep{zhang2019root,jiang2023pre}.
\newline
While these analyses shows that normalization placement can strongly affect signal/gradient propagation and training stability (especially in Pre-/Post-Norm residual blocks), they do not determine how \emph{normalization type and placement jointly} reshape the distribution of initial model outputs and class predictions. The IGB framework \citep{pmlr-v235-francazi24a} describes initialization-time predictive state explicitly at the \emph{per-class} level (Neutral vs Prejudiced), while remaining linked to learning stability~\citep{bassi2025normplacement}. However, the existing IGB theory is developed primarily for architectures \emph{without} normalization layers, and therefore does not characterize how normalization mechanisms modify these predictive regimes at initialization.
\newline
In this paper, we explicitly investigate how BN and LN shape predictive behavior prior to training.\footnote{In our framework, other normalizations such as Instance Norm or Group Norm are trivial extensions of the discussion on LN.} Our analysis reveals that normalization introduces meaningful and systematic alterations to prediction statistics, thereby influencing learning dynamics, with effects that manifest clearly in class-specific performance differences rather than only in global metrics. Importantly, even in the full-batch limit BN does \emph{not} reduce to LN: under IGB, hidden nodes within a layer are not identically distributed at a fixed initialization, so averaging across samples at a node (BN) is not equivalent to averaging across nodes for a sample (LN), yielding distinct prediction statistics at initialization (see Fig.~\ref{fig:static_Norm}). Moreover, because the layer-wise mean estimator in LN vanishes at initialization, LN is formally equivalent to RMSNorm in this regime, so our LN results directly extend to RMSNorm as well.

\begin{figure*}[t!]
    \centering
    \includegraphics[width=0.7\textwidth]{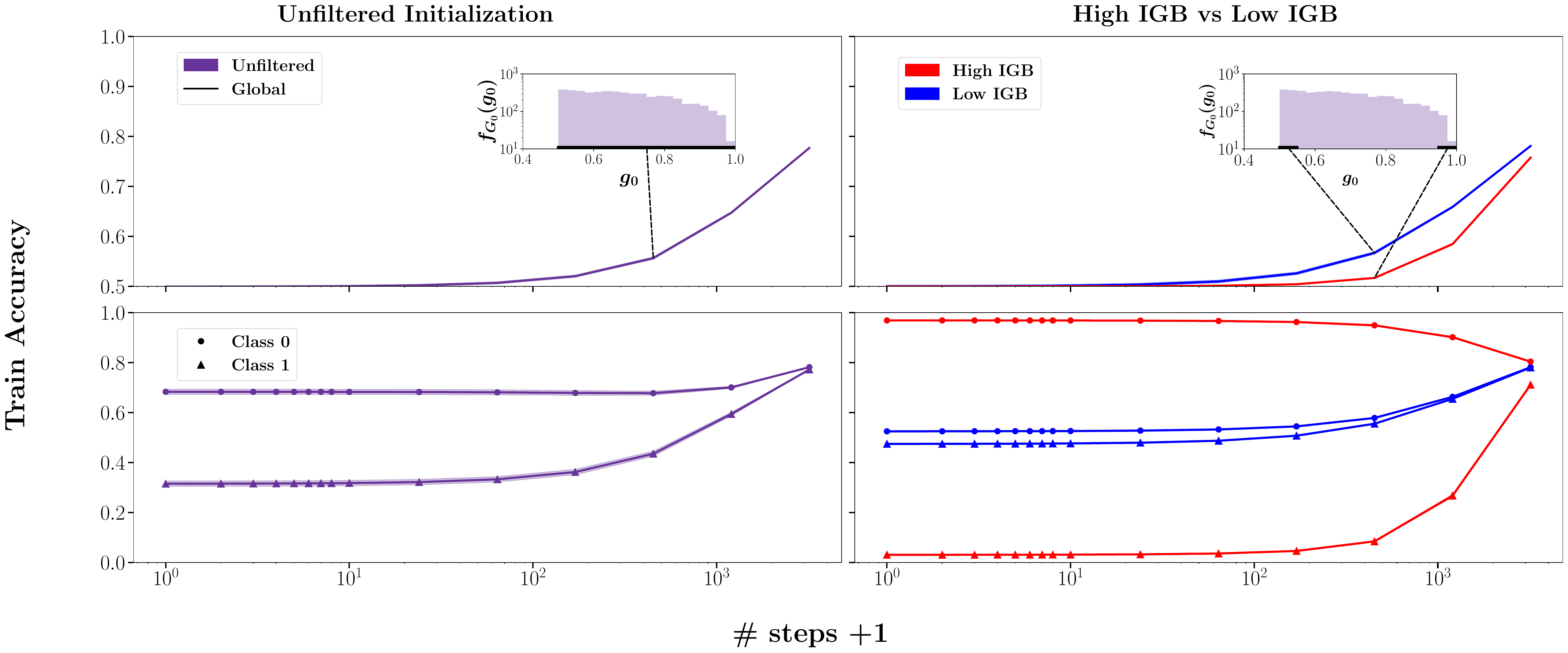}
\caption{Training accuracy dynamics measured for the whole dataset (global, top) and for single-class performance (bottom). Curves are averaged over multiple simulations. \textbf{Left}: unfiltered initializations, randomly sampled. \textbf{Right}: simulations filtered based on \(\RClassFraction{0}\) to isolate neutral and prejudiced initializations. Curves are averaged over multiple runs; in each run, class 0 corresponds to the initially dominant class to avoid averaging out class-specific effects. Filtering enables comparison of the distinct convergence behaviors associated with each initialization regime. Model: \MLPA{}, Data: \GB~(see App.~\ref{app:Exp} for more details).}
    \label{fig:training_dynamics}
\end{figure*}

\paragraph{Setting and Main Notation.}
Our theory focuses on the most fundamental building block in deep learning, the Multi-Layer Perceptron (MLP) \citep{rosenblatt1958perceptron}. MLP components are prevalent in successful modern architectures, including convolutional networks  and transformers \citep{krizhevsky2012imagenet, vaswani2017attention}.
We consider MLPs with Rectified Linear Unit (ReLU) activation functions processing a dataset composed of \(N\) inputs \(\{ \RVecInputValue{a}{} \}_{a=1}^{N}\) through \(L\) hidden layers. 
When normalization is included, we analyze both pre-activation (Pre-Norm) and post-activation (Post-Norm) configurations; we denote these respectively as \emph{Norm.\ + Activation} (e.g., Norm.\ + ReLU) and \emph{Activation + Norm.} (e.g., ReLU + Norm.).
We focus on ReLU activations \citep{nair2010rectified, krizhevsky2012imagenet} as they constitute one of the most widespread choice in modern architectures and represent a prototype of activation that lead to prejudiced states \citep{pmlr-v235-francazi24a}.
For the sake of clarity, we primarily focus on binary classification (so \(\NumberClasses =2\) classes). Our findings can be straightforwardly extended to multiclass classification, through the procedure outlined in~\cite{pmlr-v235-francazi24a}.
\newline
For our theory, we adopt the following notation and statistical assumptions:
\begin{itemize}
\item Datapoints are sampled as independent Gaussian random variables: \(\RCompInputValue{a}{;i} \sim \NormalDistr{0}{1}\). Here \(a\) indexes the sample and \(i\) the input coordinate (e.g., a pixel).
    \item Weights are independently initialized as Gaussian random variables: \(W_{ij}^{(l)} \sim \mathcal{N}\left(0, \frac{\sigma_w^2}{n^{(l-1)}}\right)\), with biases set to zero. Here, \(n^{(l)}\) denotes the number of nodes in the \(l\)-th hidden layer, and \(\sigma_w^2\) is a layer-independent constant, consistent with standard initialization practices.
\end{itemize}
The choice of random, unstructured Gaussian inputs aligns with common practice in theoretical studies ~\citep{pennington2018spectrum, koehler2021uniform, loureiro2021learning, mignacco2020dynamical}, simplifying analyses and removing confounding effects introduced by structured or correlated data. Importantly, this approach allows us to isolate and rigorously quantify class prejudice inherent to network architecture and initialization choices, free from data-induced effects. 
\newline
Similarly, initializing bias parameters to zero is standard in both theoretical and practical settings ~\citep{glorot2010understanding, he2015delving}, fostering symmetry and simplifying interpretation. Extending our analysis to nonzero biases is straightforward, as shown in \cite{bassi2025normplacement}.\newline
Finally, although our main analysis adopts the popular Gaussian (Kaiming) initialization \citep{he2015delving}, our theoretical insights naturally extend to other initialization schemes, provided they preserve independence and scale appropriately with layer size~\citep{pmlr-v235-francazi24a}. \newline
Given a DNN with a set of weights \(\WeightSet{}\), a key quantity in our analysis is the fraction of datapoints guessed as class \(c\), denoted by \(\RClassFraction{c}(\WeightSet{})\). 
Because the weights are randomly initialized, \(\RClassFraction{c}(\WeightSet{})\) is itself a random variable across the ensemble of initializations. We denote by \(\pdf{\RClassFraction{0}}{}{\ClassFraction{0}}\) its probability density function (p.d.f.) over this ensemble.\footnote{We denote by \(\pdf{X}{}{x}\) the probability density function (p.d.f.) of a random variable \(X\).} \newline
While we often focus on class 0 for clarity, the analysis is symmetric under label permutation. Accordingly, the distribution \(\pdf{\RClassFraction{0}}{}{\ClassFraction{0}}\) is symmetric across random initializations: in a prejudiced state, the model strongly favors one class, but which class is favored varies randomly from one initialization to another. 
\newline
To avoid averaging out important class-wise effects in our empirical analysis, especially when reporting per-class performance across multiple runs, we adopt a consistent labeling convention: in MLP experiments, we relabel class 0 to denote the most-predicted class at initialization. This ensures clarity and comparability of per-class statistics, particularly in regimes of deep prejudice.

\section{Not All Initializations are Equal: Impact of Neutrality vs. Prejudice}\label{Sec:InitAndDyn}

\subsection{Comparative Dynamics of Neutral \textit{vs}. Prejudiced Initializations}
Recent studies \citep{pmlr-v235-francazi24a} have demonstrated that architectural choices significantly shape the initial predictions of untrained DNNs. In particular, MLPs with ReLU exhibit a pronounced \emph{prejudice} in their initial predictions, a bias that amplifies with increasing depth. Specifically, deeper MLP architectures tend to assign predictions predominantly to a single class at initialization, resulting in highly prejudiced initial states. This phenomenon is quantitatively captured by the probability density function \( \pdf{\RClassFraction{0}}{}{\ClassFraction{0}} \), which broadens and shifts toward the extremes as depth increases.\newline
To explore the practical implications of these differences, we compare training dynamics initiated from distinct regions of the $G_0$ distribution. We select randomly initialized networks, dividing them into two groups:
\begin{itemize}
    \item \textbf{Neutral Initializations}, \textit{i.e.}, \textbf{Low IGB} (\(\RClassFraction{0} \approx 0.5\)): networks start with predictions uniformly distributed across classes; this corresponds to a sharply peaked distribution \(\pdf{\RClassFraction{0}}{}{\ClassFraction{0}}\) centered at 0.5 (Fig.~\ref{fig:Thms_summary}--bottom-left), indicating minimal prejudice in predictions.
    \item \textbf{Prejudiced Initializations} (\(\RClassFraction{0} > 0.5\)): networks exhibit biased predictions at initialization. \footnote{In MLP dynamics we relabel each initialization so that class~0 is the majority prediction at $t=0$ (see end of Sec.~\ref{Sec:Backg&Fram}); thus $G_0=\max_c G_c$ and, for $N_C=2$, $G_0\in[0.5,1]$. Static experiments keep fixed labels.} We distinguish between two subtypes: \emph{weak prejudice}, where \(\pdf{\RClassFraction{0}}{}{\ClassFraction{0}}\) is broadly spread over \([0.5, 1]\) (Fig.~\ref{fig:Thms_summary}--center-left); and \emph{deep prejudice}, \textit{i.e.}, \textbf{High IGB} (\(\RClassFraction{0} \approx 1\)), where it is sharply peaked near 1, indicating extreme initial favoring of one class (Fig.~\ref{fig:Thms_summary}--top-left).
\end{itemize}
The experiments in Fig.~\ref{fig:training_dynamics}, conducted on synthetic data (\GB{}) using  a ReLU MLP without normalization (\MLPA{}) (setting details in App.~\ref{sec:reprod}), showcase notable differences in training dynamics between these prejudiced and neutral initial states. 
Our experiments reveal that these two filtered groups of initializations display nontrivial differences in convergence time (Fig.~\ref{fig:training_dynamics}, upper right). In addition to this, we observe significant differences in the behavior of per-class performance metrics (Fig.~\ref{fig:training_dynamics}, bottom right). Networks initialized in a neutral state exhibit a more symmetric evolution in class-wise performance, allowing global metrics to more faithfully represent the learning behavior across all classes. By contrast, prejudiced initializations lead to a marked divergence in per-class accuracy that persists throughout the onset of training. As a result, the overall performance metric becomes dominated by the initially favored class and no longer serves as a reliable indicator of model behavior on the full dataset. Interestingly, this persistent performance gap resembles the behavior typically observed in imbalanced classification problems \citep{francazi2023theoretical}—yet here, the dataset is balanced, and the asymmetry originates purely from initialization picking.
\newline
These differences hold important practical implications, particularly for hyperparameter optimization procedures, which typically rely on initial performance trends. Our findings demonstrate that prejudiced initializations can result in hyperparameter choices that inadvertently optimize performance exclusively for the dominant class at initialization, neglecting balanced class performance and thereby impairing fairness and overall training efficiency.
App.~\ref{sec:hint} provides an illustration of how the decision boundary evolves under neutral versus prejudiced initializations, together with an intuitive argument of how these differences shape the ensuing training dynamics.

\begin{figure*}[t!]
    \centering
    \includegraphics[width=0.45\textwidth]{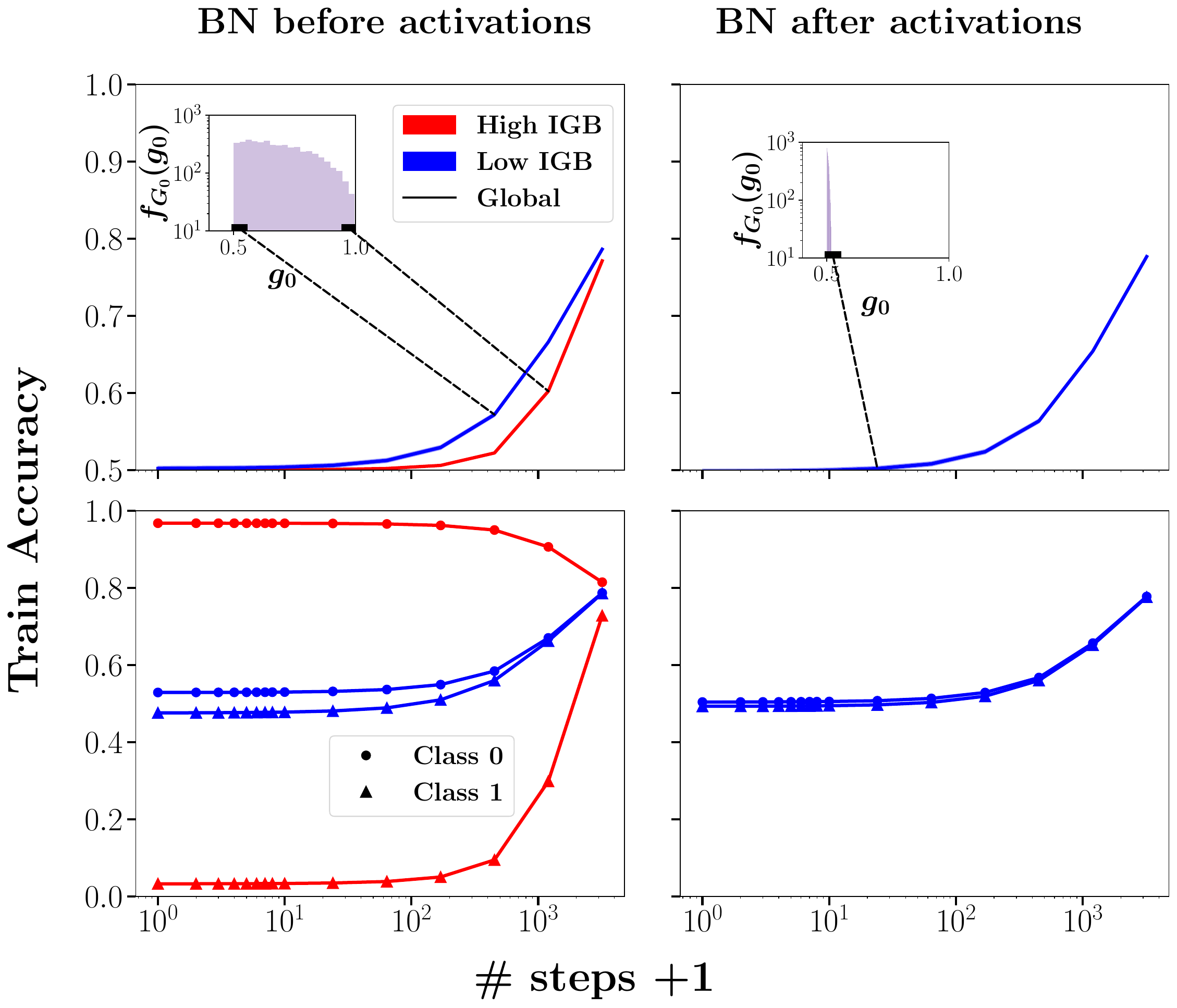}
    \includegraphics[width=0.45\textwidth]{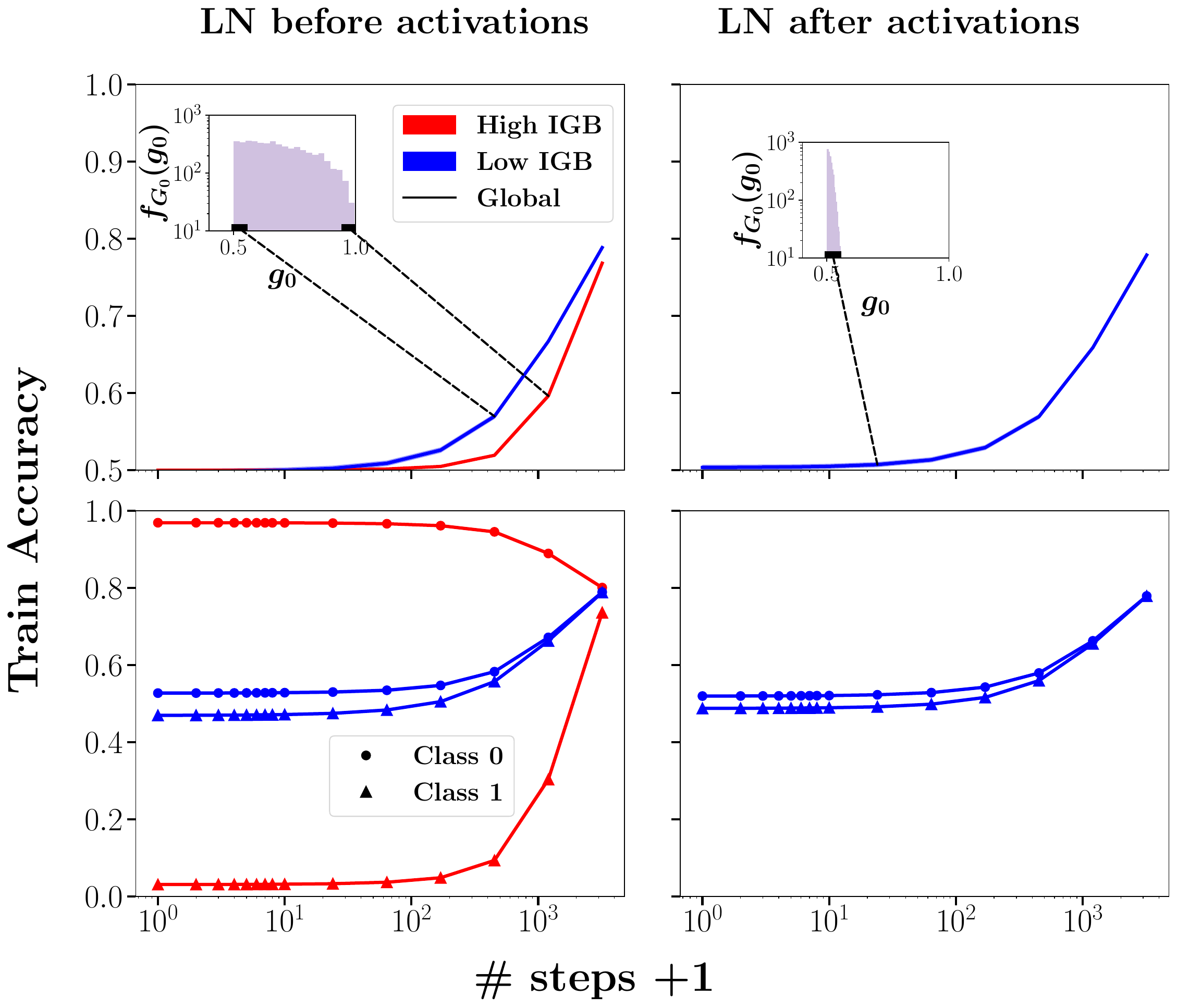}
    \caption{
Training accuracy dynamics measured for the whole dataset (global, top) and for single-class performance (bottom). From the left to the right: BN before ReLU,  BN after ReLU, LN before ReLU,  LN after ReLU. Curves are averaged over multiple simulations. Simulations are filtered based on \(\RClassFraction{0}\) to isolate neutral and prejudiced initializations. In each run, class 0 corresponds to the initially dominant class to avoid averaging out class-specific effects. Filtering enables comparison of the distinct convergence behaviors associated with each initialization regime. Model: \MLPA{}, Data: \GB~(see App.~\ref{app:Exp} for more details).}
    \label{fig:1HL_Dyn}
\end{figure*}

\section{How Normalization Layers Influence the Initial Guessing Bias}

Sec.~\ref{Sec:InitAndDyn} discussed how the initial prediction bias---quantified by the statistic \(\RClassFraction{0}\)---plays a key role in shaping the training dynamics of neural networks. Different values of \(\RClassFraction{0}\), sampled from the ensemble of random initializations, induce qualitatively different learning processes. Thus, the distribution of \(\RClassFraction{0}\) across initializations is a critical factor that determines the expected learning trajectory of a model. 
\cite{pmlr-v235-francazi24a} have shown that activation functions and network depth significantly influence the distribution of \(\RClassFraction{0}\). 
In this section, we investigate how normalization layers affect \( \pdf{\RClassFraction{0}}{}{\ClassFraction{0}} \), highlighting the significant impact of their position relative to activation functions.
\newline
Fig.~\ref{fig:static_Norm} visually illustrates these outcomes, highlighting that post-activation normalization consistently yields initializations that fall within a neutral state, while pre-activation normalization leads to a distribution spread over the whole support, including both neutral and prejudiced initializations.
\newline
These differences translate into the dynamics.
Building on the setup from Fig.~\ref{fig:training_dynamics}, Fig.~\ref{fig:1HL_Dyn} incorporates BN or LN either before or after ReLU in the \MLPA{}. The resulting dynamics, along with the distributions \(\pdf{\RClassFraction{0}}{}{\ClassFraction{0}}\) shown in the inset plots, resemble the trends observed in Fig.~\ref{fig:training_dynamics} when normalization is applied \textbf{before} activation, for both BN and LN. However, moving the normalization \textbf{after} the activation dramatically alters the distribution: only the neutral region remains populated, effectively eliminating access to prejudiced states through random sampling. As a consequence, unfiltered (random) initializations exhibit the behavior characteristic of neutral regimes and depart significantly from the dynamics shown in Fig.~\ref{fig:training_dynamics}.

\subsection{Proving the Impact of Normalizations on Networks of Arbitrary Depth}

Now, we provide theorems showing the impact of BN and LN on the distribution of \(\RClassFraction{0}\) at initialization. We will see that, BN and LN are equivalent for single-hidden layer models, but their effect becomes different as depth increases. 
We also prove that placing the norms before or after the activation has qualitatively different effects in terms of IGB.
The four theorems are empirically corroborated in Fig. \ref{fig:static_Norm} through the theoretical curves corresponding to each case.
\newline
Our proofs are for ReLU networks because the effect of normalizations is generally a reduction of IGB, so it does not make sense to analyse activations such as $\tanh$, which do not produce IGB \citep{pmlr-v235-francazi24a}.
Extension to other activations causing IGB is straightforward and we do not expect qualitative differences.

\subsubsection{Batch Normalization}
We provide two theorems for BN, depending on its placement with respect to the ReLU activation.

\begin{theorembox}[Informal: BatchNorm + ReLU]
\label{thm:bn-deep-before}
Consider a deep MLP with ReLU activation, within the general setting described in Sec.~\ref{Sec:Backg&Fram}. Placing BN \textbf{before} the activation leads to a weakly prejudiced initialization, that remains stable with increasing depth.
\end{theorembox}
The formal formulation of Th. \ref{thm:bn-deep-before} is provided in Th. \ref{thm:BN_B_A}, alongside its proof.

\begin{theorembox}[Informal: ReLU + BatchNorm]
\label{thm:bn-deep-after}
Consider a deep MLP with ReLU activation, within the general setting described in Sec.~\ref{Sec:Backg&Fram}. Placing BN \textbf{after} the activation promotes a neutral initialization state.
\end{theorembox}
The formal formulation of Th. \ref{thm:bn-deep-after} is provided in Th. \ref{thm:BN_A_A}, alongside its proof.
\newline
These results show that Batch Normalization induces two qualitatively different behaviors depending on placement. When applied \textit{after} the activation, BN effectively suppresses the initial prejudice and drives the model toward neutrality, reducing the spread of the guessing distribution \(\pdf{\RClassFraction{0}}{}{\ClassFraction{0}}\) around 0.5. 
\newline
In contrast, placing BN \textit{before} the activation retains the initial prejudice seen in shallow unnormalized networks. However, crucially, this prejudice does not grow with depth: the distribution \(\pdf{\RClassFraction{0}}{}{\ClassFraction{0}}\) remains broad but does not become increasingly concentrated at the extremes. Thus, BN-before acts as a stabilizing mechanism, preventing the depth-induced amplification of IGB that would otherwise occur in the absence of normalization \citep{pmlr-v235-francazi24a}, though without actively eliminating the prejudice.

\subsubsection{Layer Normalization}

Also for LN, we provide two theorems, depending on its placement with respect to the ReLU activation.

\begin{theorembox}[Informal: LayerNorm + ReLU]
\label{thm:ln-deep-before}
Consider a deep MLP with ReLU activation, within the general setting described in Sec.~\ref{Sec:Backg&Fram}. Placing LN \textbf{before} the activation leads to an increasingly prejudiced initialization as depth increases.
\end{theorembox}
The formal formulation of Th. \ref{thm:ln-deep-before} is provided in Th. \ref{thm:LN_before}, alongside its proof.

\begin{theorembox}[Informal: ReLU + LayerNorm]
\label{thm:ln-deep-after}
Consider a deep MLP with ReLU activation, within the general setting described in Sec.~\ref{Sec:Backg&Fram}. Placing LN \textbf{after} the activation promotes a neutral initialization state.
\end{theorembox}
The formal formulation of Th. \ref{thm:ln-deep-after} is provided in Th. \ref{thm:LN_A_A}, alongside its proof.
\newline
The behavior of Layer Normalization exhibits a more polarized dependence on placement. When LN is placed \textit{after} the activation, it mirrors the prejudice-reducing effect of ReLU + BN: the initialization becomes neutral, with the distribution \(\pdf{\RClassFraction{0}}{}{\ClassFraction{0}}\) sharply centered around 0.5, indicating balanced class predictions.
However, placing LN \textit{before} the activation leads to a significantly different behavior. In this case, the network retains the depth-amplified prejudice observed in architectures without normalization: as depth increases, the distribution \(\pdf{\RClassFraction{0}}{}{\ClassFraction{0}}\) becomes increasingly concentrated near the extremes (0 and 1), reflecting highly prejudiced predictions. Unlike BN-before, which dampens this growth, LN-before leaves the depth-induced amplification of prejudice unchanged—mirroring the behavior observed in networks without normalization \citep{pmlr-v235-francazi24a}.
\newline
\subsubsection{LayerNorm vs RMSNorm.} 
Under LN + ReLU, the distribution of an individual hidden layer node is no longer centered at $0$: each node becomes centered around a random shift. These shifts vary across nodes and are themselves drawn from a distribution that is symmetric and centered at $0$ across the layer. Consequently, although single-node distributions are not symmetric around $0$, the \emph{layer-level} distribution obtained by aggregating over all nodes remains symmetric.\newline
Crucially, in LN the mean is estimated over the entire layer (i.e., across nodes for a given input), rather than across samples at a fixed node (as in BN). At initialization, this layer-wise mean estimator is null. Therefore, even in the LayerNorm + ReLU configuration—where individual nodes exhibit non-zero means—LayerNorm is \emph{formally equivalent to RMSNorm at initialization}. The practical implication is that RMSNorm avoids explicit mean estimation while preserving the same initialization-time stabilization effect, providing a principled explanation for why RMSNorm-style choices are natural, even in Pre-Norm architectures.
\subsubsection{Norm Methods: Overview of Effects}
Together, these results reveal critical differences in how normalization layers interact with architectural depth and activation placement. Both BN and LN placed \textit{after} activations consistently suppress prejudice and maintain a balanced initialization across depths. However, when placed \textit{before} activations, their behaviors diverge sharply: BN + ReLU stabilizes the level of prejudice, whereas LN + ReLU allows it to keep growing with depth. 
\newline
This divergence is not merely a quantitative effect but reflects a qualitative difference in the mechanisms at play. Under IGB, neurons within the same layer are not identically distributed: averaging across samples (as in BN) and averaging across neurons (as in LN) lead to fundamentally different normalizing factors. As a result, BN + ReLU and LN + ReLU cannot be regarded as interchangeable design choices, despite both belonging to the same family of normalization methods.
\newline
These theoretical results are confirmed by the simulations in Fig.~\ref{fig:static_Norm}: the theoretical curves closely match the empirical distributions across synthetic inputs and real data. The figure illustrates how placement and normalization type shape \(\pdf{\RClassFraction{0}}{}{\ClassFraction{0}}\) across depths: BN-before flattens out the depth-induced amplification, LN-before maintains the extremal, prejudiced states—similarly to the unnormalized case—and both BN-after and LN-after maintain narrow, centered distributions indicative of neutrality.

\begin{figure*}[t!]
    \centering
    \includegraphics[width=0.8\textwidth]{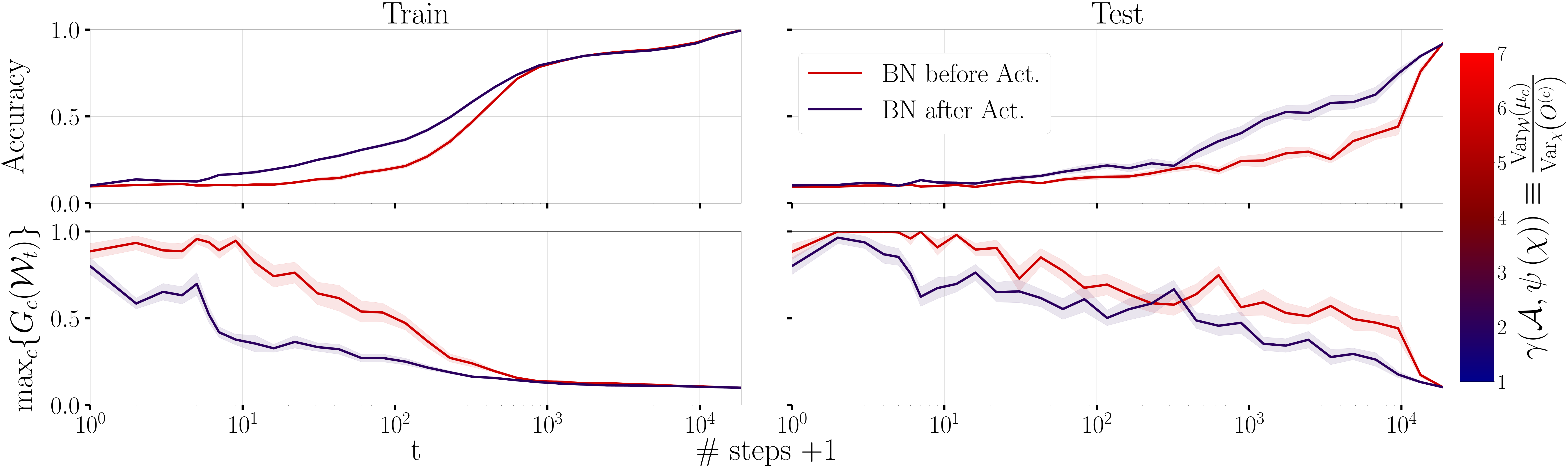}
    \caption{Comparison of ResNet-9 training on CIFAR-10 with the standard configuration (BN before activation) and an alternative design (BN after activation).  The level of IGB in each setting is quantified using the variance ratio \(\VarRatio\). \textbf{Top panels:} accuracy evaluated over the entire dataset. \textbf{Bottom panels:} evolution of the initial guessing bias, quantified at each training step by \(\max_{c} \RClassFraction{c}(\WeightSet{}_t)\), the maximum fraction of datapoints assigned to a single class. As predicted by our analysis, the post-activation configuration yields a lower \(\VarRatio\), indicating a more balanced initial state. See App.~\ref{sec:reprod} for further experimental details.}

    \label{fig:ResNet_exp}
\end{figure*}

\section{Complementary Experiments}\label{sec:data_rob}

\subsection{Observations on real data.}
Our empirical observations extend beyond MLPs and synthetic data. In Fig.~\ref{fig:static_Norm}, we report the distribution \(\pdf{\RClassFraction{0}}{}{\ClassFraction{0}}\) computed on CIFAR-10 in lieu of random Gaussian inputs. The resulting profiles closely track those obtained with random data, indicating a regime in which architectural factors chiefly govern the distribution \(\pdf{\RClassFraction{0}}{}{\ClassFraction{0}}\) while dataset structure plays a secondary role. \emph{In this regime, the theoretical prediction from random-input analysis is not only qualitatively accurate but also descriptively faithful for real data.} These similarities, visible in the static distributions at initialization, are corroborated by further experiments on the dynamics shown in App.~\ref{sec:exp_mlp}, which mirror the random-data case.
\newline
Note however that our predictions are not expected to quantitatively describe any real dataset.
As argued in \citep{pmlr-v235-francazi24a}, strong correlations in the data  are expected to amplify IGB. To this end, we examine MNIST, where pixels hold similar values through the whole dataset. As seen in Figs.~\ref{fig:mnist_1hl_train} and~\ref{fig:mnist_1hl_test}, MNIST departs more from the random-input predictions than CIFAR-10 (the qualitative trends remain, but there are quantitative deviations), confirming the argument of \citep{pmlr-v235-francazi24a}. 
Crucially, the \emph{relative} impact of normalization placement remains consistent with our analysis: placing normalization \emph{before} (rather than \emph{after}) the activation increases the level of prejudice. The same qualitative pattern holds across different architectures and datasets (see App.~\ref{app:exp_Other_arch}). Taken together, our results show regimes in which architecture-induced effects alone yield descriptive predictions for structured data, and others where data structure perturbs \(\pdf{\RClassFraction{0}}{}{\ClassFraction{0}}\), leaving the architecture-only theory at a qualitative level. Delineating these regimes and extending the theory to explicitly incorporate data-induced factors remains a rich direction for future work.

\subsection{Exploring impact of normalization placement.}
Given the difference between \textit{Norm.+ReLU} and \textit{ReLU+Norm.}, we examine widely adopted architectures—ResNets with BN and MLP-Mixers with LN—comparing the standard placement (\emph{before} the activation) with the alternative (\emph{after} the activation) suggested by our analysis (see App.~\ref{app:exp_Other_arch}). Without filtering initializations, we directly evaluate the training dynamics to assess the practical implications of these choices.
\newline
As shown in Fig.~\ref{fig:ResNet_exp}, and consistent with our theoretical and empirical insights, we observe clear differences in convergence and class-wise behaviour between the two placements. Placing normalization \emph{after} the activation yields faster convergence and more balanced dynamics, especially in the early phase. This is evident not only in performance (top panels) but also in the bias trajectory, \(\max_{c}\RClassFraction{c}(\WeightSet{}_t)\) (bottom), which decays earlier under post-activation normalization. As model selection and hyperparameter tuning rely on early validation windows, persistent onset bias can skew the chosen configuration toward the initially over-predicted classes rather than the dataset as a whole. Notably, even when training curves later overlap, test performance continues to reflect a separation consistent with the earlier convergence-speed gap, suggesting that initialization prejudice can persist and influence generalization in a non-trivial way. These findings indicate that standard normalization placement heuristics may deserve closer scrutiny, and that revisiting them in light of initialization behavior can yield tangible improvements.

\section{Conclusion}
This work provides a theoretical characterization of how normalization design shapes the predictive behavior of DNNs at initialization, yielding distinct regimes of initial guessing bias (IGB) ranging from neutral to prejudiced. A key practical implication is that \emph{placement} (Pre-Norm vs.\ Post-Norm) is not an implementation detail: Post-Norm shifts DNNs toward neutral regimes, while Pre-Norm preserves prejudice. This initialization gap in turn impacts convergence and class-wise dynamics during the onset of training. Importantly, because this bias can persist during the onset stage, global accuracy may be misleading for hyperparameter selection, as it can reflect the behavior of the initially dominant class more than that of the dataset as a whole.
\newline
Our results also clarify why \emph{normalization type} can induce different regimes in deep settings. Under IGB, neurons within a layer are not identically distributed, so estimators computed at a single node (BN: averaging across samples) are not equivalent to estimators computed at the layer level (LN: averaging across nodes), especially in Pre-Norm configurations. Finally, while the single-node mean estimator can be non-zero in Pre-Norm, the layer-wise mean estimator is null at initialization; consequently, LayerNorm is formally equivalent to RMSNorm at initialization, explaining why RMSNorm can avoid mean estimation without changing the initialization-time stabilization effect in that regime.

\clearpage

\section*{Acknowledgements}
This work was supported by the Swiss National Foundation, SNF grant \# 196902.


\bibliographystyle{plainnat}
\bibliography{IGB_Norm}

\clearpage

\section*{Checklist}

\begin{enumerate}

  \item For all models and algorithms presented, check if you include:
  \begin{enumerate}
    \item A clear description of the mathematical setting, assumptions, algorithm, and/or model. [Yes]\\
    \emph{Justification:} The paper presents the mathematical setting and intuition in the main text and enumerates assumptions alongside the formal statements in the supplemental material.

    \item An analysis of the properties and complexity (time, space, sample size) of any algorithm. [Not Applicable]\\
    \emph{Justification:} The work provides a theoretical analysis of initialization bias rather than proposing a new algorithm that would necessitate a runtime or memory complexity study.

    \item (Optional) Anonymized source code, with specification of all dependencies, including external libraries. [Yes]\\
    \emph{Justification:} An anonymized repository with all code and instructions to reproduce figures and results is provided; environment and dependency details are documented in App.~\ref{sec:reprod}.
  \end{enumerate}

  \item For any theoretical claim, check if you include:
  \begin{enumerate}
    \item Statements of the full set of assumptions of all theoretical results. [Yes]\\
    \emph{Justification:} Each theoretical result is accompanied by a complete list of assumptions in the appendix/supplement.

    \item Complete proofs of all theoretical results. [Yes]\\
    \emph{Justification:} Full, formal proofs for all stated results are included in the supplemental material, with intuition provided in the main text.

    \item Clear explanations of any assumptions. [Yes]\\
    \emph{Justification:} Assumptions (e.g., unstructured/random data) are motivated in the main paper, and their scope and limitations are discussed, including robustness considerations in Section~\ref{sec:data_rob}.
  \end{enumerate}

  \item For all figures and tables that present empirical results, check if you include:
  \begin{enumerate}
    \item The code, data, and instructions needed to reproduce the main experimental results (either in the supplemental material or as a URL). [Yes]\\
    \emph{Justification:} Reproduction materials (code, scripts, and step-by-step instructions) are supplied via an anonymized repository; details are summarized in App.~\ref{sec:reprod}.

    \item All the training details (e.g., data splits, hyperparameters, how they were chosen). [Yes]\\
    \emph{Justification:} Training/test splits, hyperparameters, optimizer settings, and selection criteria are comprehensively documented in App.~\ref{sec:reprod}.

    \item A clear definition of the specific measure or statistics and error bars (e.g., with respect to the random seed after running experiments multiple times). [Yes]\\
    \emph{Justification:} Results are averaged over multiple random seeds; error bars report the standard error of the mean as stated in App.~\ref{app:Exp}.

    \item A description of the computing infrastructure used. (e.g., type of GPUs, internal cluster, or cloud provider). [No]\\
    \emph{Justification:} Detailed compute specifications (hardware type, execution time) are not reported; experiments were designed to be lightweight and feasible without reliance on large-scale GPU infrastructure, ensuring reproducibility.
  \end{enumerate}

  \item If you are using existing assets (e.g., code, data, models) or curating/releasing new assets, check if you include:
  \begin{enumerate}
    \item Citations of the creator If your work uses existing assets. [Yes]\\
    \emph{Justification:} Public datasets used (e.g., CIFAR-10, MNIST) are properly cited in the manuscript.

    \item The license information of the assets, if applicable. [Yes]\\
    \emph{Justification:} Licenses and terms of use for all datasets are respected and explicitly mentioned as appropriate.

    \item New assets either in the supplemental material or as a URL, if applicable. [Not Applicable]\\
    \emph{Justification:} No new datasets or models are introduced beyond the reproducibility code package.

    \item Information about consent from data providers/curators. [Not Applicable]\\
    \emph{Justification:} No new data were collected; only standard publicly available datasets are used.

    \item Discussion of sensible content if applicable, e.g., personally identifiable information or offensive content. [Not Applicable]\\
    \emph{Justification:} The work relies on common benchmarks and synthetic data and does not involve PII or sensitive content.
  \end{enumerate}

  \item If you used crowdsourcing or conducted research with human subjects, check if you include:
  \begin{enumerate}
    \item The full text of instructions given to participants and screenshots. [Not Applicable]\\
    \emph{Justification:} The study does not involve crowdsourcing or human-subject experiments.

    \item Descriptions of potential participant risks, with links to Institutional Review Board (IRB) approvals if applicable. [Not Applicable]\\
    \emph{Justification:} No human participants were involved, so IRB approval is not required.

    \item The estimated hourly wage paid to participants and the total amount spent on participant compensation. [Not Applicable]\\
    \emph{Justification:} No participants were recruited and no compensation was provided.
  \end{enumerate}

\end{enumerate}

\clearpage
\appendix
\thispagestyle{empty}

\onecolumn

\appendix

\counterwithin{figure}{section}
\onecolumn

\vbox{
{\hrule height 2pt \vskip 0.15in \vskip -\parskip}
\centering
{\LARGE\bf Appendix\par}
{\vskip 0.2in \vskip -\parskip \hrule height 0.5pt \vskip 0.09in}
}
\vspace{-5mm}

\newcommand\invisiblepart[1]{%
\refstepcounter{part}%
\addcontentsline{toc}{part}{\protect\numberline{\thepart}#1}%
}

\invisiblepart{Appendix}
\setcounter{tocdepth}{2}
\localtableofcontents

\clearpage

\section{Notation}\label{sec:Notation}
In this section, we thoroughly present the setting and notation used in our analysis. Our theory examines Multi-Layer Perceptrons (MLPs), where the propagation of an input signal, the datapoint "$a$", denoted as $\RVecInputValue{;a}{}$, through the network is governed by the following set of equations:
\begin{align}
\RCompHidNodeBAF{i;a}{l+1} &= \sum_{j} \Weights{l+1}{ij} \RCompDeepInputValue{j}{;a}{l} \, ,\label{eq:h_prop_def} \\
\RVecDeepInputValue{}{;a}{l} &= \Layerperator{\left\{ \RVecHidNodeBAF{;a}{l} \right\}_{a=1}^{\DatasetSize} } \, \\
\RCompHidNodeBAF{c;a}{L+1} &\equiv \ROutNode{c;a}{} \,
, \label{eq:DNN_prop}
\end{align}
where $l \in \{ 0, \dots, L \}$ indicate the layer, $a \in \{ 1, \dots, \DatasetSize \}$ the sample index, $i,j \in \{ 1, \dots, \LayerNumNodes{l} \}$ the node index. The operator $\Layerperator{\cdot}$ is given by the composition of two operations, the activation function and the normalization. In our theoretical analysis, we will consider two different normalization schemes (BN and LN) and, for each of them, two different settings defined by the placement of the normalization with respect to the activations. We will therefore consider in total 4 different settings:
\begin{itemize}
    \item  activation + BN : $\Layerperator{\cdot} = \BNOperatorNoArg \, \circ \, \AFOperator{\cdot}$
    \item BN + activation: $\Layerperator{\cdot} = \AFOperatorNoArg \, \circ \, \BNOperator{\cdot}$
    \item activation + LN: $\Layerperator{\cdot} = \LNOperatorNoArg \, \circ \, \AFOperator{\cdot}$
    \item LN + activation: $\Layerperator{\cdot} = \AFOperatorNoArg \, \circ \, \LNOperator{\cdot}$
\end{itemize}

$\AFOperator{\cdot} : \mathbb{R} \rightarrow \mathbb{R}$ represents the activation function, applied elementwise across the layer.

We now define the normalization operators used throughout our analysis. For a fixed layer $l$, node index $i$, and sample index $a$, the Batch Normalization (BN) and Layer Normalization (LN) operators are defined as follows:

\paragraph{Batch Normalization (BN).} The BN operator normalizes across the minibatch:
\begin{align}
\BNOperator{\RCompHidNodeBAF{i;a}{l}} 
= \alpha_i^l \cdot 
\frac{
    \RCompHidNodeBAF{i;a}{l} - \BNMean{i}{l}
}{
    \sqrt{ \BNVar{i}{l} + \epsilon }
} + \beta_i^l
\,, \label{eq:bn_def}
\end{align}
where the statistics are computed as:
\begin{align*}
\BNMean{i}{l} &= \frac{1}{\BatchSize} \sum_{b \in B(a)} \RCompHidNodeBAF{i;b}{l}, \\
\BNVar{i}{l}  &= \frac{1}{\BatchSize} \sum_{b \in B(a)} \left( \RCompHidNodeBAF{i;b}{l} - \BNMean{i}{l} \right)^2.
\end{align*}
where $B(a)$ indicates the batch of where the datapoint $a$ belongs.
The parameters $\alpha_i^l$ and $\beta_i^l$ are trainable scaling and shifting factors, and $\epsilon$ is a small constant added for numerical stability.

\paragraph{Layer Normalization (LN).} The LN operator normalizes across the layer dimensions for each sample independently:
\begin{align}
\LNOperator{\RCompHidNodeBAF{i;a}{l}} 
= \alpha_i^l \cdot 
\frac{
    \RCompHidNodeBAF{i;a}{l} - \LNMean{a}{l}
}{
    \sqrt{ \LNVar{a}{l} + \epsilon }
} + \beta_i^l
\,, \label{eq:ln_def}
\end{align}
where the mean and variance are computed as:
\begin{align*}
\LNMean{a}{l} &= \frac{1}{\LayerNumNodes{l}} \sum_{i=1}^{\LayerNumNodes{l}} \RCompHidNodeBAF{i;a}{l}, \\
\LNVar{a}{l}  &= \frac{1}{\LayerNumNodes{l}} \sum_{i=1}^{\LayerNumNodes{l}} \left( \RCompHidNodeBAF{i;a}{l} - \LNMean{a}{l} \right)^2.
\end{align*}

\begin{figure}[h]
    \centering
    \includegraphics[width=0.95\columnwidth]{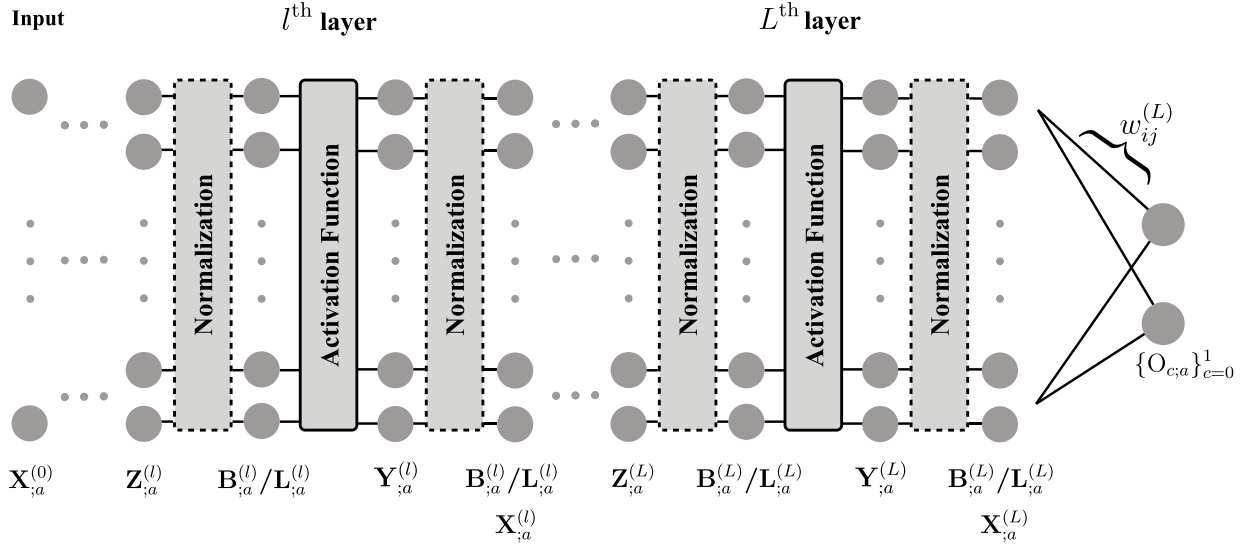}
    \caption{Illustration of a generic feedforward neural network used for binary classification, along with the main notation for node-level variables. Normalization layers (representing either Batch Normalization or Layer Normalization) are shown with dashed contour to indicate that they may be positioned either before or after the activation function, depending on the configuration.}
\label{fig:NN_scheme}
\end{figure}

The principal notation is summarized in Fig.~\ref{fig:NN_scheme}.
\newline
We also use the following notation:
\begin{itemize}
\item $\{ \cdot \}_{i=0}^{M-1}$: set of $M$ elements. If some of the indices of the variables are fixed (\emph{\textit{i.e.}} equal for every element of the set), the set indices (indices that vary across different elements of the set) are reported explicitly on the right. If the index of the set elements is not explicitly reported, it means the absence of fixed indices for the set variables (\textit{\textit{i.e.}} all indices are set indices).
\item $\RVecHidNodeABN{l}{;a}$: activation vector at layer $l$ for input sample $a$, given by $\left( \RCompHidNodeABN{l}{1;a} , \dots, \RCompHidNodeABN{l}{\LayerNumNodes{l};a}\right)$, after passing through the BN; $\RCompHidNodeABN{l}{i;a}$ indicate the component corresponding to node $i$.
\item $\CE{}{x}$: Indicate the expectation value of the argument, $x$. If the average involves only one source of randomness this is explicitly indicated, \emph{\textit{e.g.}} $\CE{\Dataset}{x}$ indicates an average over the dataset distribution, while $\CE{\WeightSet{}}{x}$ an average over the distribution of network weights. 
\item $\erf\left( \cdot \right) $:  Error function. $\erf\left(\ArgHidNodeAAF \right) \equiv \frac{2}{\sqrt{\pi}} \int_0^{x} e^{-t^2} dt$
\item $\cdf{X}{}{x}$:Given a \textit{r.v.} $X$, we denote its cumulative distribution function (\textit{c.d.f.}) as $\cdf{X}{}{x}$, i.e., $\cdf{X}{}{x} = \Prob{}{X<x}$. Considering the \textit{r.v.} $X(\Dataset, \WeightSet{})$, which is a function of two independent sets of random variables $\Dataset$ and $\WeightSet{}$, when one of these sources of randomness is fixed, we explicitly indicate the active source in the notation. For example, $\cdf{X}{(\Dataset)}{x} = \Prob{}{X<x \mid \WeightSet{}}$.

\item $\RMultiClassFraction{c}{}$: fraction of dataset elements classified as belonging to  class $c$. The argument $M$ indicates the total number of output nodes  for the variable definition, \emph{i.e.} the number of classes considered. For binary problems we omit this argument ($\RClassFraction{0}$) as there is only one non-trivial possibility, \emph{i.e.} $M=2$ .
\item $\RMultiClassRankedFraction{c}{}$: the set $\left\{ \RMultiClassFraction{c}{} \right\}_{c=0}^{M-1}$ contain the same elements of $\left\{ \RMultiClassRankedFraction{c}{} \right\}_{c=0}^{M-1}$, but these are ranked by magnitude, such that  $\RMultiClassRankedFraction{0}{}$ is the greatest output value between the $M$ elements of the set, $\RMultiClassRankedFraction{1}{}$ the second one and so on.

\item $\RVecHidNodeALN{l}{;a}$: activation vector at layer $l$ for input sample $a$, given by $\left( \RCompHidNodeALN{l}{1;a} , \dots, \RCompHidNodeALN{l}{\LayerNumNodes{l};a}\right)$, after passing through LN; $\RCompHidNodeALN{l}{i}$ indicate the component corresponding to node $i$.
\item $\NumberClasses \equiv \LayerNumNodes{L+1}$: number of output nodes, \emph{\textit{i.e.}} the number of classes.
\item $\LayerNumNodes{l}$: number of nodes in the $l^{\text{th}}$ layer; $\LayerNumNodes{0}\equiv d$ indicates the dimension of the input data (number of input layer nodes) while $\LayerNumNodes{L+1}$ the number of classes (number of output layer nodes).

\item $\NormalDens{x}{\mu}{\sigma^2}$: Given a Gaussian \textit{r.v.} $X$, we indicate with $\NormalDens{x}{\mu}{\sigma^2}$ the \textit{p.d.f.} computed at $X=x$, \emph{\textit{i.e.}}
$\NormalDens{x}{\mu}{\sigma^2} \equiv \pdf{X}{}{x} = \frac{e^{-\frac{1}{2 \sigma^2}(x-\mu)^2}}{\sqrt{2 \pi \sigma^2}}$

\item $\ROutNode{c,}{M}$: output layer node; $c$ is the node index; the index M, instead, indicate the total number of nodes considered. For binary problems we omit the subscript index to keep the notation lighter, \emph{\textit{i.e.}} $\ROutNode{c}{}$ .
\item $\ROutNode{\tilde 0,}{M}$: the set $\left\{ \ROutNode{\tilde c,}{M} \right\}_{ c=0}^{M-1}$ contain the same elements of $\left\{ \ROutNode{\tilde c,}{M} \right\}_{\tilde c=0}^{M-1}$, but these are ranked by magnitude, such that  $\ROutNode{\tilde 0,}{M}$ is the greatest output value between the $M$, $\ROutNode{\tilde 1}{M}$ the second one, and so on.
\item $\Prob{}{A}$: Denotes the probability associated with event $A$. $\Prob{}{A \mid B}$ indicates the probability of event $A$ given event $B$.

\item $\pdf{X}{}{x'}$: Given a random variable $X$, $\pdf{X}{}{x'}$ denotes the probability density function (\textit{p.d.f.}) evaluated at $X=x'$. Formally, $\pdf{X}{}{x'} = \frac{d}{dx} \cdf{X}{}{x}\Bigr|_{\substack{x=x'}}$. For variables with multiple sources of randomness, if some of these sources are either fixed or marginalized, we will specify the active sources of randomness as subscripts. For example, given $X(\Dataset, \WeightSet{})$, a function of two independent sets of random variables $\Dataset$ and $\WeightSet{}$, $\pdf{X}{(\Dataset)}{x'}$ equals $\cdf{X}{(\Dataset)}{x}\Bigr|_{\substack{x=x'}}$.

\item $\CVar{}{\cdot}$: Indicate the variance of the argument. Since we have \textit{r.v.}s with multiple sources of randomness where necessary we will specify in the subscript the source of randomness used to compute the expectation. For example $\CVar{\Dataset}{\cdot} \equiv \Einput{\cdot - \Einput{\cdot}}^2$. For the sake of compactness, we will employ sometimes the shorthand notation $\CVar{\Dataset}{\cdot} = \sigma_{\cdot}^2$.

\item $\WeightSet{}_{t}$: shorthand notation for the set of network weights, $\{ \Weights{l}{ij} \}$ at time $t$; $\WeightSet{} \equiv \WeightSet{}_{0}$. We use, instead the notation $\WeightSet{l} \subseteq \WeightSet{}$ to indicate the subset of weights relative to a specific layer, \emph{\textit{i.e.}} $\WeightSet{l} \equiv \{ \Weights{l}{ij} \}_{\substack{j \in [0, \dots, \LayerNumNodes{l}]\\i \in [0, \dots, \LayerNumNodes{l+1}]}}$. $\WeightSet{<l}, \; \WeightSet{>l}, \dots $ are defined analogously.

\item $\Weights{l}{ij}$: element $ij$ of the matrix $\WeightsMat{l}$, connecting two consecutive layers ($l \in [0, \dots, L]$). Given the matrix $\WeightsMat{l}$ we use a ‘placeholder’ index, $\Cdot$, to return column and row vectors from the weight matrices. In particular $\Weights{l}{j \Cdot}$ denotes row $j$ of the weight matrix $\WeightsMat{l}$; similarly, $\Weights{l}{\Cdot j}$ denotes column j.
\item $\RVecInputValue{;a}{} \in \mathbb{R}^{d}$: $a^{\text{th}}$ input vector; when the index $a$ is omitted we mean a generic vector, $\RVecInputValue{}{}$, drawn from the population distribution.
\item $\RVecHidNodeAAF{a}{l}$: activation vector at layer $l$ for input sample $a$, given by $\left( \RCompHidNodeAAF{1;a}{l}, \dots, \RCompHidNodeAAF{\LayerNumNodes{l};a}{l} \right)$. Each component $\RCompHidNodeAAF{i;a}{l}$ corresponds to the post-activation value of node $i$.

\item $\RVecHidNodeBAF{;a}{l}$: pre-activation vector at layer $l$ for input sample $a$, obtained after the linear transformation of the activations from layer $l-1$ and before applying any activation or normalization. Explicitly, $\RVecHidNodeBAF{;a}{l} = \left( \RCompHidNodeBAF{1;a}{l}, \dots, \RCompHidNodeBAF{\LayerNumNodes{l};a}{l} \right)$.

\item $\Dirac{x}$: Dirac delta function.

\item $\Heavyside{x}$: Heaviside step function.

\item $\RMean{X} \equiv \Einput{X}$: expected value (mean) of a generic random variable $X$ over the dataset distribution, conditioned on a fixed realization of the weights $\WeightSet{}$. For example, $\RMean{\ROutNode{c}{}}$ denotes the average value of the output node $c$ over the dataset. As shorthand, we write $\RMean{c} \equiv \Einput{\ROutNode{c}{}}$.

\item $\RStd{X} \equiv \sqrt{\Einput{(X - \RMean{X})^2}}$: standard deviation of a generic random variable $X$ over dataset randomness, given a fixed realization of the weights $\WeightSet{}$. This quantifies variability in $X$ across different input samples from the dataset.

\item $\Dataset = (\RVecInputValue{;a}{}, \LabelValue_a)_{a=1}^{\DatasetSize}$: dataset composed by $\DatasetSize$ pairs of input vectors-labels.

\end{itemize}

\paragraph{Abbreviations}
\begin{itemize}
    \item BN: Batch Normalization
    \item \textit{c.d.f.}: cumulative distribution function
    \item CNN: Convolutional neural network
    \item DNN: Deep neural network
    \item EOC: edge of chaos
    \item IGB: Initial guessing bias
    \item LN: Layer Normalization
    \item MLP: Multi-layer perceptron
     \item \textit{p.d.f.}: probability density function
     \item \textit{r.v.}: random variable
     \item \textit{w.h.p.}: with high probability
\end{itemize}

\section{Theoretical Setting and additional Related Work}
\label{sec:Settings}

Multi-Layer Perceptrons (MLPs) are fundamental building blocks for various contemporary architectures \cite{he2016deep, vaswani2017attention, dosovitskiy2020image}, including recent models such as MLP-mixer \cite{tolstikhin2021mlp}, achieving notable results on complex tasks. Understanding the initialization behavior of MLPs is essential due to their broad applicability and evolving importance. We specifically examine MLPs integrated with normalization techniques—Batch Normalization (BN) and Layer Normalization (LN)—and their effects on output distributions and Initial Guessing Bias (IGB).

Normalization methods, particularly BN and LN, have become essential components in modern neural network architectures. Originally introduced to address internal covariate shifts and stabilize training processes \citep{ioffe2015batch, ba2016layer}, these techniques normalize activations by adjusting their means and variances, significantly impacting prediction statistics. Their empirical success has spurred widespread adoption across various domains, such as computer vision and natural language processing \citep{bjorck2018understanding, xu2019understanding}. While normalization's influence on training dynamics, optimization and function representation has been extensively explored \citep{santurkar2018does, yang2019mean, xu2019understanding}, comparatively fewer studies have focused on its effects on prediction behavior at initialization \citep{daneshmand2020batch, kohler2019exponential, ni2024nonlinearity, lyu2022understanding, balestriero2022batch}.

We consider an MLP with $L$ hidden layers, applied to a dataset of input vectors $\{ \RVecInputValue{;a}{} \}_{a=1}^{\DatasetSize}$. Each input is processed through the network to produce two output values, $\{ \ROutNode{c;a}{} \}_{c=0,1}$, with the predicted class determined by the larger of the two. We focus on unstructured Gaussian inputs, where each component is independently sampled as $\RCompInputValue{i}{} \sim \NormalDistr{0}{1}$.
\newline
Weights are initialized using \textbf{Kaiming Initialization}, with $\Weights{l}{ij} \sim \NormalDistr{0}{\frac{\StdKaiming^2}{\LayerNumNodes{l}}}$, where $\LayerNumNodes{l}$ is the number of neurons in the $l^\text{th}$ hidden layer. In our experiments we adopt the standard setting $\StdKaiming^2 = 2$. While our analysis focuses on this concrete case for clarity, the framework extends to more general classes of random input distributions and initialization schemes, as discussed in~\citep{pmlr-v235-francazi24a}. For complete notation, see App.~\ref{sec:Notation}.

Our analysis focuses on the infinite-width limit, employing tools such as the Central Limit Theorem (CLT) and its generalizations \citep{pmlr-v235-francazi24a}, assuming Gaussian-distributed inputs and using Kaiming Normal initialization \cite{he2015delving}. These assumptions provide a symmetric and unbiased environment, isolating network-intrinsic behaviors and facilitating the study of IGB and normalization effects. While Gaussian inputs simplify theoretical analysis, our findings generalize to other input distributions satisfying CLT conditions. Similarly, although we primarily utilize ReLU activation functions for explicit calculations, our theoretical results hold broadly across generic activation functions.

The normalization layers' positions are flexible in our study, considering BN and LN applied either before or after the activation functions. BN normalizes across batches, while LN normalizes within individual layers. This variability helps us systematically explore their influence on network initialization and the resulting biases. Furthermore, while biases are set to zero to maintain symmetry, extensions to non-zero biases and structured input data scenarios remain feasible, following methodologies similar to those in \citep{pmlr-v235-francazi24a, francazi2023theoretical}.

Beyond these directions, other works have examined predictive behavior at initialization from complementary perspectives. \citet{joudaki2023bridging} introduced an entropy-based characterization of predictive imbalance. While this approach highlights empirical, indirect signatures of class bias, the IGB framework provides a direct theoretical formalization of the per-class prediction distribution. These views are complementary: entropy-based insights can be embedded within the IGB formalism to yield explicit theoretical predictions. \citet{joudaki2024emergence} analyzed how changes in LayerNorm placement affect signal correlations across weight configurations. This classical mean-field measure can be directly connected to IGB bias measures, as shown in \citet{bassi2025normplacement}. Finally, density-of-states methods have been proposed at finite width, where the energy is defined via misclassified examples, thereby granting access to predictive behaviour across possible configurations~\citep{mele2024density}.

\section{Overview of Initial Guessing Bias (IGB)}\label{app:IGB_overview}

Initial Guessing Bias (IGB) characterizes the phenomenon where untrained neural networks exhibit an inherent bias in their initial class predictions, even in symmetric settings with identically distributed classes and the absence of non-null bias terms. Specifically, given an architecture parameterized by weights \(\WeightSet{}\), the fraction of points initially guessed as class \(c\) by the network is denoted by \(\RClassFraction{c}(\WeightSet{})\). IGB manifests as an imbalance in this fraction across classes.

For clarity, we focus on binary classification problems, noting that the analysis naturally extends to multi-class scenarios. Without loss of generality, class 0 is used as representative, leveraging the statistical symmetry between classes.

In the limit of infinitely many datapoints (\(\DatasetSize \rightarrow \infty\)), by the Law of Large Numbers, we have:
\begin{align}
\lim_{\DatasetSize \rightarrow \infty} \RClassFraction{0}\left( \WeightSet{} \right) &= \Prob{}{\ROutNode{0}{}> \ROutNode{1}{} \mid \WeightSet{} } \\
&= \int_0^{\infty} \pdf{\RDiff{\ROutNode{0}{}}}{(\Dataset)}{ \Diff{\OutNode{}{}} } \; d\Diff{\OutNode{}{}}\, , \quad \text{with} \quad \RDiff{\ROutNode{0}{}} \equiv \ROutNode{0}{} - \ROutNode{1}{}. \notag
\end{align}

The IGB framework allows the derivation of the distribution \(\pdf{\ROutNode{c}{}}{(\Dataset)}{ \OutNode{}{}}\). Specifically, we recall the following result from \cite{pmlr-v235-francazi24a}:

\begin{theorembox}[Informal]
\label{thm:out_dist}
Consider a Gaussian distributed dataset processed through an MLP with \(L\) hidden layers and weights initialized according to Kaiming normal initialization (or schemes with similar scaling properties). In the limit of infinite width, the distribution of an output node \(\ROutNode{c}{}\), at initialization, converges to:
\begin{align}
\pdf{\ROutNode{c}{}}{(\Dataset)}{ \OutNode{}{}} \xrightarrow{| \WeightSet{} | \rightarrow \infty} \NormalDens{\OutNode{}{}}{\RMean{c}}{\CVar{\Dataset}{\ROutNode{c}{}}},
\end{align}
where \(\RMean{c} = \Einput{\ROutNode{c}{}} \equiv\mathbb{E}_{\RVecInputValue{}{}}[\ROutNode{c}{}]\) and \(\CVar{\Dataset}{\ROutNode{c}{}}\equiv\text{Var}_{\RVecInputValue{}{}}(\ROutNode{c}{}).\) Moreover, the center \(\RMean{c}\) is itself a random variable varying across initializations, converging to:
\begin{align}
\pdf{\RMean{c}}{}{\Mean{}} \xrightarrow{| \WeightSet{} | \rightarrow \infty} \NormalDens{\Mean{}}{0}{\CVar{\WeightSet{}}{\RMean{c}}}.
\end{align}
\end{theorembox}

In other words, the IGB framework reveals that initial guessing biases can be characterized comparing two types of fluctuations: those in \(\pdf{\RMean{c}}{}{\Mean{}}\), representing variability across different output nodes, and those in \(\pdf{\ROutNode{c}{}}{(\Dataset)}{ \OutNode{}{}}\), representing variability across datapoints for a fixed initialization. Intuitively:
\begin{itemize}
    \item \textbf{Absence of IGB:} If initialization fluctuations (\(\pdf{\RMean{c}}{}{\Mean{}}\)) are significantly smaller than data-driven fluctuations (\(\pdf{\ROutNode{c}{}}{(\Dataset)}{ \OutNode{}{}}\)), outputs for both classes largely overlap, resulting in nearly equal fractions of assigned classes (\(\RClassFraction{0}\approx 1/2\)).
    \item \textbf{Strong IGB:} Conversely, if initialization fluctuations dominate, output distributions for each class become well-separated, leading to an overwhelming fraction assigned to a single class (\(\RClassFraction{0}\approx 1\)).
\end{itemize}

A natural translation of this intuition into a quantifiable metric is provided by the following definition:

\begin{defbox}[Variance Ratio]
\label{def:VarRatio}
Given an architecture \(\Arch\) and a dataset \(\PreprData\), we define the Variance Ratio \(\VarRatio\) as:
\begin{equation}
\VarRatio \equiv \frac{\CVar{\WeightSet{}}{\RMean{c}}}{\CVar{\Dataset}{\ROutNode{c}{}}},
\end{equation}
where:
\begin{itemize}
    \item \(\CVar{\WeightSet{}}{\RMean{c}}\) is the variance (over random initializations \(\WeightSet{}\)) of the output node mean over the dataset randomness \(\RMean{c} = \Einput{\ROutNode{c}{}}\),
    \item \(\CVar{\Dataset}{\ROutNode{c}{}}\) is the variance (over data inputs \(\RVecInputValue{}{}\)) of the output \(\ROutNode{c}{}\), for a fixed initialization.
\end{itemize}
\end{defbox}

We further generalize this definition by introducing \(\LayerVarRatio{l}\), computed analogously at any layer \(l\) of the network.

\subsection{Geometric Interpretation of IGB Effects: Intuition Behind Differing Dynamics}\label{sec:hint}

\begin{figure*}[t!]
    \centering
    \includegraphics[width=0.9\textwidth]{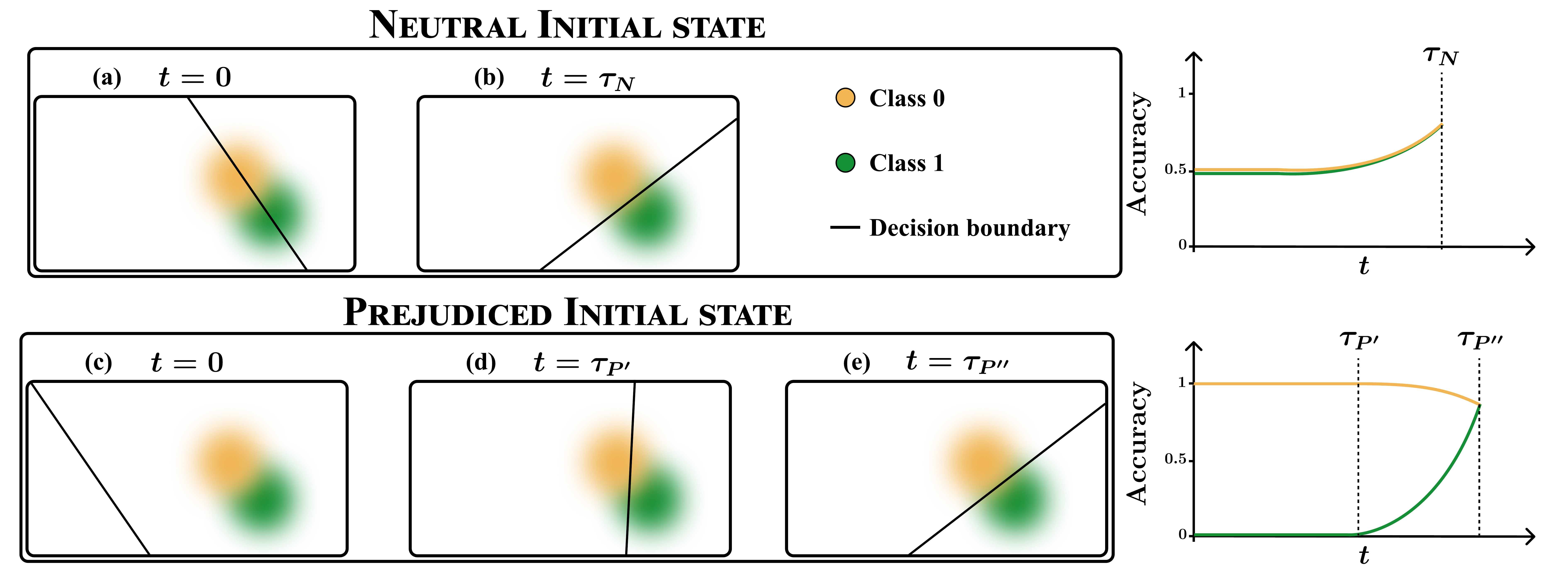}
    \caption{Low-dimensional representation of decision boundary dynamics for prejudiced vs. neutral initializations. In the neutral case, the boundary already intersects the relevant region (a), and training involves only directional adjustment (b). In the prejudiced case, instead, the distribution of the hidden representation occupy a region far from the decision boundary (c). In this case one can argue that the training proceeds in two steps: translating the boundary to the high-density region of datapoint representations (d), and adjusting the orientation to optimize class separation (e). These regimes reflect fundamentally different underlying dynamics.}
    \label{fig:decision_boundary_dynamics}
\end{figure*}

We provide an intuitive argument for the interpretation of the distinct training dynamics observed under neutral and prejudiced initializations, as shown in Fig.~\ref{fig:training_dynamics}. These differences stem from IGB, which reflects the spatial relationship between the decision boundary and the distribution of input representations \citep{pmlr-v235-francazi24a}.
\newline
In the prejudiced state, IGB is caused by a drift in the hidden representation of the last layer. As the data representations move away from the center of the activation space—\textit{i.e.}, the vector space formed by the layer's output activations—the decision boundary—typically initialized near the origin—fails to bisect the signal. Consequently, all data points fall predominantly on one side of the hyperplane, causing the network to assign the vast majority of inputs to a single class (Fig.~\ref{fig:decision_boundary_dynamics} (c)).
Conversely, in the neutral state, the distribution of the hidden representation of the last layer remains approximately centered. The decision boundary intersects this central region, partitioning the datapoints more evenly between classes (Fig.~\ref{fig:decision_boundary_dynamics} (a)). This leads to a more balanced class-wise prediction behavior at initialization.
\newline
These initial conditions can naturally give rise to structurally different learning dynamics, which appear to operate over distinct regimes and characteristic timescales. In prejudiced initializations (Fig.~\ref{fig:decision_boundary_dynamics} (c,d,e)), training unfolds in two steps. During the initial regime ($0 < t < \tau_{P'}$), the decision boundary approaches the region where data representations are concentrated. This is followed by a second phase ($\tau_{P'} < t < \tau_{P''}$), where the boundary is gradually rotated and refined to achieve effective class separation.
In contrast, neutral initializations involve only a single training phase (Fig.~\ref{fig:decision_boundary_dynamics} (a,b)). Since the decision boundary already intersects the central region of the data representations, training proceeds via local directional adjustments to align the boundary with the true class separation. This single-phase regime is governed by a unified timescale $\tau_N$.
\newline
To illustrate these differences visually, Fig.~\ref{fig:decision_boundary_dynamics} presents a low-dimensional schematic of the decision boundary evolution. In the neutral case, the boundary initially intersects the region of interest (a) and requires only directional tuning (b). In the prejudiced case, the boundary must first be translated toward the signal region (d) before undergoing orientation adjustments to optimize separation (e).
The right side of Fig.~\ref{fig:decision_boundary_dynamics} translates the geometric interpretation discussed on the left into the temporal structure of training dynamics, summarizing how the different phases and timescales associated with each initialization regime can give rise to the distinct convergence behaviors observed in Fig.~\ref{fig:training_dynamics}.

\clearpage

\section{Conditions for Distribution convergence}

In this section, we briefly review the key results on the convergence in distribution of random variables, which form the foundation of our theoretical analysis. These results have been extensively discussed in earlier works; for more detailed explanations and proofs, we refer the reader to \cite{ petrov2012sums, pmlr-v235-francazi24a}. Below, we outline the necessary conditions for convergence and characterize the resulting asymptotic distributions.

\subsection{Convergence in Distribution}
Our analysis focuses on the convergence of random variable combinations to their asymptotic distributions. The cornerstone of this discussion is the \textit{Central Limit Theorem (CLT)}, which describes the conditions under which sums of random variables converge to a Gaussian distribution.

\subsubsection{Central Limit Theorem}
The \textit{Central Limit Theorem (CLT)} applies to sequences of independent and identically distributed (\textit{i.i.d.}) random variables. Given a sequence of \textit{i.i.d.} random variables $x_1, x_2, \dots$ with finite mean $\mu$ and variance $\sigma^2$, the sample mean $\bar{x}_n$ of the first $n$ variables converges to a normal distribution:
\[
Z = \lim_{n \rightarrow \infty} \frac{\bar{x}_n - \mu}{\sigma/\sqrt{n}} \sim \mathcal{N}(0, 1).
\]
Corrections to this asymptotic form for finite $n$ are discussed in \cite{keller2001rate}.

\subsubsection{Extensions of the CLT}
The CLT can be extended to sequences of independent but non-identically distributed random variables. Under conditions like the Lyapunov or Lindeberg conditions, it remains possible to demonstrate convergence to a Gaussian distribution. One such condition, the \textit{Lindeberg condition}, for a sequence of independent random variables $X_1, X_2, \dots$, ensures that:
\[
\lim_{n \rightarrow \infty} \sum_{i=1}^n \frac{1}{s_n^2} \int_{|x| \geq \epsilon s_n} (x - \mu_i)^2 dF_{X_i}(x) = 0,
\]
where $s_n^2 = \sum_{i=1}^n \mathbb{E}[(X_i - \mu_i)^2]$ and $F_{X_i}(x)$ is the cumulative distribution function of $X_i$.

\begin{theorembox}[Lindeberg's Theorem]
\label{th:Lindeberg}
For a sequence of independent random variables $\{ X_i \}_{i=1}^n$, if the Lindeberg condition is satisfied for all $\epsilon > 0$, then the normalized sum $Z_n = \frac{S_n - \sum_i \mu_i}{s_n}$ converges to a standard normal distribution:
\[
\pdf{Z_n}{}{z} \xrightarrow{n\rightarrow\infty} \mathcal{N}(z; 0, 1).
\]
\end{theorembox}

For additional details, see Ref.~\cite{petrov2012sums}.

\subsection{Sufficient Conditions for Convergence}
We will also rely on an alternative set of necessary and sufficient conditions that guarantee convergence to a normal distribution. These conditions are outlined below, and the full proof can be found in Refs.~\cite{petrov2012sums, pmlr-v235-francazi24a}.

\begin{theorembox}[Distribution Convergence]
\label{thm:DistrConv}
Consider a set of independent zero-mean random variables $\{ X_i \}_{i=1}^n$. If, for every $\epsilon > 0$, the following conditions are satisfied:
\begin{align}
& \text{Concentration:} \quad \sum_{i=1}^{n} \Prob{}{ |X_i| \geq \epsilon } \xrightarrow{n\rightarrow\infty} 0 \quad \forall \epsilon \in \mathbb{R}^+, \label{eq:concentration_cond} \\
& \text{Mean Normalization:} \quad \sum_{i} \left( \int_{|x| < \epsilon} x \pdf{X_i}{}{x} dx \right) \xrightarrow{n\rightarrow\infty} 0, \label{eq:MeanNorm_cond}\\
& \text{Variance Normalization:} \\ &\quad s_n^2 = \sum_i \left( \int_{|x| < \epsilon} x^2 \pdf{X_i}{}{x} dx - \left( \int_{|x| < \epsilon} x \pdf{X_i}{}{x} dx \right)^2 \right) \xrightarrow{n\rightarrow\infty} \sigma^2, \label{eq:VarNorm_cond}
\end{align}
then the distribution of the sum $\sum_i X_i$ converges to $\mathcal{N}(0, \sigma^2)$.
\end{theorembox}

\subsection{Sufficient Conditions for Theorem~\ref{thm:DistrConv}}
In many cases, it is sufficient to check simpler conditions for convergence. Specifically, the following conditions are sufficient for convergence to a Gaussian distribution (the full proof can be found in Ref.~\cite{pmlr-v235-francazi24a}.):

\begin{theorembox}[Sufficient Conditions for Distribution Convergence]
\label{thm:DistrConv_SC}
Let $\{ X_i \}_{i=1}^n$ be a sequence of independent, zero-mean random variables. If they satisfy the following conditions:
\begin{align}
& \text{Variance Convergence:} \quad \sum_{i=1}^n \mathbb{E}[X_i^2] \xrightarrow{n \rightarrow \infty} n \tilde{\sigma}^2, \label{eq:Var_conv_cond} \\
& \text{Fast Decreasing Tails:} \quad \lim_{x \rightarrow \pm \infty} \pdf{X_i}{}{x} = \mathcal{O}\left( \frac{1}{x^4} \right) \quad \forall i, \label{eq:fast_tail_dec_cond}
\end{align}
then the normalized sum of the random variables converges in distribution to $\mathcal{N}(0, \sigma^2)$.
\end{theorembox}

\newpage

\section{Batch Normalization}\label{app:BN}

\subsection{BN + ReLU}\label{app:BN_B_A}

In this section, we show that, when BN is applied before the activation function, the output distribution of each layer remains consistent throughout the network, regardless of depth. Specifically, after applying BN, the output distribution of a deep network with many hidden layers is equivalent to that of a network with just a single hidden layer without any normalization. 
We establish this result using an induction argument, demonstrating that the distribution of nodes at each hidden layer, at initialization, follows the same distribution, independent of the network depth.
For clarity, we focus on neural networks with weights initialized using the Kaiming Normal scheme, but the analysis extends to other initialization methods (\textit{e.g.}, Xavier Uniform) as long as the variance scaling is preserved. Furthermore, these results extend to more complex activation architectures, including when incorporating pooling layers~\cite{pmlr-v235-francazi24a}.

The following is a formal formulation of Th. \ref{thm:bn-deep-before}.

\begin{theorembox}[BN + ReLU]
\label{thm:BN_B_A}
Consider the setting described in Sec.~\ref{sec:Settings} with an MLP where BN is applied before the activation function. Assuming that the activations $\left\{ \RCompHidNodeAAF{i}{l}  \right\}$ satisfy the conditions of Thm.~\ref{thm:DistrConv_SC}, in the limit of infinite input size and  infinite width, the distribution of the output node $\ROutNode{c}{}$ at initialization converges to:
\begin{align}
\label{eq:PO_asym}
\pdf{\ROutNode{c}{}}{(\Dataset)}{\OutNode{}{}} \xrightarrow{| \WeightSet{} | \rightarrow \infty} \NormalDens{\OutNode{}{}}{\RMean{c}}{\StdKaiming^2 \CVar{\Dataset}{\RCompHidNodeAAF{0}{1}}{}},
\end{align}
where $\CVar{\Dataset}{\RCompHidNodeAAF{0}{1}}{}$ is the variance of a generic node in the first layer activations, over the dataset randomness. Moreover, while the variance of the distribution, $\CVar{\Dataset}{\ROutNode{c}{}}$, converges with high probability (w.h.p.) to a deterministic value, the mean $\RMean{c}$ of the distribution converges in distribution to:
\begin{align}
\label{eq:P_cen_asym}
\pdf{\RMean{c}}{}{\Mean{}} \xrightarrow{| \WeightSet{} | \rightarrow \infty} \NormalDens{\Mean{}}{0}{\StdKaiming^2 \Einput{\RCompHidNodeAAF{0}{1}}^2},
\end{align}
where $\Einput{\RCompHidNodeAAF{0}{1}}$ is the mean over the dataset randomness of the activated nodes.

Here, $| \WeightSet{} | \rightarrow \infty$ refers to the limit where the number of neurons $\LayerNumNodes{l}$ in each hidden layer $l \in \{0, \dots, L \}$ tends to infinity.
\end{theorembox}

\paragraph{Presence and Absence of IGB:} Thm.~\ref{thm:BN_B_A} gives rise to two distinct scenarios:

\begin{itemize}
    \item \textbf{Absence of IGB}: If the mean $\Einput{\RCompHidNodeAAF{1}{0}}$ converges to zero as the dataset size increases, the distribution of $\RMean{c}$ converges to a Dirac delta at zero:
    \begin{align}\label{eq:NoIGB_Thm_out_cond}
        \lim_{\DatasetSize \rightarrow \infty} \pdf{\RMean{c}}{}{\Mean{}}  = \delta(\Mean{}),
    \end{align}
    indicating the absence of Initial Guessing Bias (IGB).

\item \textbf{Presence of IGB}: If $\Einput{\RCompHidNodeAAF{0}{1}}$ converges to a finite value, the network exhibits IGB, and:
\begin{align}
\label{eq:IGB_Thm_out_cond}
    \lim_{\DatasetSize \rightarrow \infty} \pdf{\RMean{c}}{}{\Mean{}}  = \NormalDens{\Mean{}}{0}{\StdKaiming^2 \Einput{\RCompHidNodeAAF{0}{1}}^2}.
\end{align}
In this case, both the output node distribution \eqref{eq:PO_asym} and the center distribution \eqref{eq:P_cen_asym} depend only on the first hidden layer statistics, meaning the distributional properties remain unchanged as the network depth increases; therefore, the bias does not amplify with network depth.

\end{itemize}

 In this section, we focus on the effect of activation functions, but we highlight that the analysis can be extended to deal with more complex architectures (\textit{e.g.} when including pooling layers).

\subsubsection{Full Batch Case}

We begin by proving Theorem~\ref{thm:BN_B_A} in the full-batch setting. In this case, all datapoints are used to compute the shift and scale estimators for the BN, as defined in Eq.~\eqref{eq:bn_def}. Consequently, the random variables $\RVecHidNodeABN{1}{;a}$ no longer depend on the sample index $a$, since all datapoints are part of the same batch. For clarity, we omit the sample index from the notation.\newline
We prove the result by induction, showing that the distribution $\pdf{\RCompHidNodeABN{l}{i}}{(\Dataset)}{\ArgHidNodeABN}$ remains invariant across layers $l$. First, we show that:
\begin{equation}
    \pdf{\RCompHidNodeABN{1}{i}}{(\Dataset)}{\ArgHidNodeABN} = \NormalDens{\ArgHidNodeABN}{0}{1}.
\end{equation}
Starting with multivariate Gaussian-distributed inputs, \textit{i.e.}, $\RCompInputValue{i}{} \sim \NormalDistr{\mu}{\sigma^2}, \, \forall i \in [1, \dots, \LayerNumNodes{0}]$, the nodes will remain normally distributed after passing through the dense layer, since the input undergoes a linear transformation:
\begin{equation}
     \RCompHidNodeBAF{i}{1} = \sum_{j=1}^{\LayerNumNodes{0}} \Weights{1}{ij} \RCompInputValue{j}{}, 
 \end{equation}
where $\RCompHidNodeBAF{i}{1}$ represents the $i$-th node after the Dense Layer, before passing through the activation function in the first hidden layer. 
After the linear transformation the input is processed by BN (see recurrent Eq.~\eqref{eq:DNN_prop} for the configuration BN \textbf{before} activation).
The output of Batch Normalizationm for each node $i$, $\RCompHidNodeABN{1}{i}$, follows a standard normal distribution by construction. Specifically,
\begin{align}  \label{distr after batchnorm}
    \RCompHidNodeABN{1}{i} = \frac{\RCompHidNodeBAF{i}{1} - \CE{\Dataset}{\RCompHidNodeBAF{i}{1}}}{\sigma_{\RCompHidNodeBAF{i}{1}}} \sim \NormalDistr{0}{1}.
\end{align}

Next, we establish the inductive step: assuming $\pdf{\RCompHidNodeABN{l}{i}}{(\Dataset)}{\ArgHidNodeABN} =  \NormalDens{\ArgHidNodeABN}{0}{1}$ for all $i$, we aim to show that $\pdf{\RCompHidNodeABN{l+1}{i}}{(\Dataset)}{\ArgHidNodeABN} =  \NormalDens{\ArgHidNodeABN}{0}{1}$. This will be done via the following sequence of transformations (see Fig.~\ref{fig:NN_scheme}):
\[
\pdf{\RCompHidNodeABN{l}{i}}{(\Dataset)}{\ArgHidNodeABN} \rightarrow \pdf{\RCompHidNodeAAF{i}{l}}{(\Dataset)}{\ArgHidNodeAAF} \rightarrow \pdf{\RCompHidNodeBAF{i}{l+1}}{(\Dataset)}{\ArgHidNodeBAF} \rightarrow \pdf{\RCompHidNodeABN{l+1}{i}}{(\Dataset)}{\ArgHidNodeABN}.
\]

The nodes after BN are identically distributed by hypothesis. Since the activation function applies element-wise to each node, the activated nodes $\{ \RCompHidNodeAAF{i}{l} \}_{i=1}^{\LayerNumNodes{l}}$ are also identically distributed. Passing through the dense layer combines the activated nodes linearly:
\begin{equation}\label{h^l+1 distribution}
    \RCompHidNodeBAF{i}{l+1} = \sum_{j=1}^{N_l} \Weights{l}{ij} \RCompHidNodeAAF{j}{l}.
\end{equation}
The expectation over the dataset randomness is:
\begin{equation}\label{mean_h}
    \Einput{\RCompHidNodeBAF{i}{l+1}} = \sum_{j=1}^{N_l} \Weights{l}{ij}\Einput{\RCompHidNodeAAF{j}{l}} \stackrel{\text{IID}}{=} \Einput{\RCompHidNodeAAF{0}{1}} \sum_{j=1}^{N_l} \Weights{l}{ij} \equiv \Einput{\RCompHidNodeAAF{0}{1}} S_w^{(l)},
\end{equation}


where the independence of $\Weights{l}{ij}$ and $\RCompHidNodeAAF{j}{l}$ is used to extract the mean value $\Einput{\RCompHidNodeAAF{0}{l}}$, which is identical for all $j$. Here, $S_w^{(l)} = \sum_{j=1}^{N_l} \Weights{l}{ij}$ represents the sum over the weights.

In the last passage, we introduced $S_w^{(l)}$ and used the fact that $ \pdf{\RCompHidNodeAAF{i}{l}}{(\Dataset)}{\ArgHidNodeAAF} = \pdf{\RCompHidNodeAAF{i}{1}}{(\Dataset)}{\ArgHidNodeAAF}$, meaning the distributions of the activated nodes do not change with depth.

The variance, conditional on the weights, is given by:
\begin{align}\label{var_h}
    \CVar{\Dataset}{\RCompHidNodeBAF{i}{l+1}} &=  
    \CVar{\Dataset}{\sum_{j=1}^{N_l} \Weights{l}{ij}\RCompHidNodeAAF{j}{l}}{}
    = \sum_{j=1}^{N_l} \left(\Weights{l}{ij}\right)^2   
    \CVar{\Dataset}{\RCompHidNodeAAF{j}{l}}{} \notag \\
    &\stackrel{\text{ID}}{=} \CVar{\Dataset}{\RCompHidNodeAAF{0}{l}}{} S_{w^2}^{(l)} \stackrel{\LayerNumNodes{l} \to \infty}{=} \CVar{\Dataset}{\RCompHidNodeAAF{0}{l}}{} \sigma^2_w,
\end{align}
where $S_{w^2}^{(l)} = \sum_{j=1}^{N_l} \left(\Weights{l}{ij}\right)^2$. As $\LayerNumNodes{l} \to \infty$, the distribution of $S_{w^2}^{(l)}$ concentrates around a deterministic value \cite{pmlr-v235-francazi24a}.

Finally, by the Central Limit Theorem (CLT), the distribution of $\RCompHidNodeBAF{i}{l+1}$ is:
\begin{equation}\label{eq:fZ^l+1BN_FB}
   \pdf{\RCompHidNodeBAF{i}{l+1}}{(\Dataset)}{\ArgHidNodeBAF}  = \NormalDens{\ArgHidNodeBAF}{\Einput{\RCompHidNodeAAF{0}{1}} S_w^{(l)}}{\CVar{\Dataset}{\RCompHidNodeAAF{0}{1}} \sigma_w^2}.
\end{equation}

\textbf{Remark:} The distribution in Eq.~\ref{eq:fZ^l+1BN_FB} is identical to the output distribution of a network with just one ReLU and Dense layer, i.e. a ReLU single hidden layer perceptron without normalization \citep{pmlr-v235-francazi24a}.

\subsubsection{Mini Batch case}

In practice, we often do not have direct access to the underlying dataset statistics; instead, these are estimated over a random subset $\Batch$ of $\BatchSize$ dataset samples (batch) .
In particular, we define
\begin{itemize}
    \item $ \BNMean{i}{l}= \frac{1}{\BatchSize}\sum_{b \in \Batch(a)} \RCompHidNodeBAF{i;b}{l}$ is the estimator of $\mathbb{E}_\chi\left[\RCompHidNodeBAF{i}{l} \right] = \RMean{\RCompHidNodeBAF{i}{l}}$,
    \item $ \BNStd{i}{l} = \sqrt{ \frac{1}{\BatchSize}\sum_{b \in \Batch(a)} \left( \RCompHidNodeBAF{i;b}{l} - \BNMean{i}{l} \right)^2}$ is the   estimator of the standard deviation $\RStd{\RCompHidNodeBAF{i}{l}}$,
    \item $\RCompHidNodeABN{l}{i;a} = \frac{\RCompHidNodeBAF{i;a}{l} - \BNMean{i}{l}}{\BNStd{i}{l}}$ is the BN transformation.
\end{itemize}

When these estimators are computed over a finite number of samples, $B$, we can identify the following differences compared to the full batch case ( \textit{\textit{i.e.}} the case when $\BatchSize \to \infty$):
\begin{itemize}
    \item In the full batch case, the estimators converge to the mean and standard deviation of the distribution ($\RMean{\RCompHidNodeBAF{i}{l}}$ and $\RStd{\RCompHidNodeBAF{i}{l}}$), which are constants, while in the mini-batch case, the estimators $\BNMean{i}{l}$ and $\BNStd{i}{l}$ are functions of the set $\{ \RCompHidNodeBAF{i;b}{l} \}_{b \in \Batch(a)}$-where $\Batch(a)$ indicates the batch whose $a$ is element of-making the estimators themselves random variables.
    \item It is also noted that $\BNMean{i}{l}$ and $\RProxyStd{\RCompHidNodeBAF{i;a}{1}}$ are therefore functions of $\RCompHidNodeBAF{i;a}{l}$ and are thus correlated with it.
\end{itemize}

However, $\BNMean{i}{l}$  and $\BNStd{i}{l}$  are not the only possible choices of estimators capable of solving the problem of internal covariate shift. 

We make a small change to the estimators to facilitate the analysis, where we simply discard the datapoint $a$ from the average and standard deviation estimator computed for each $\RCusCompHidNodeABN{l}{i;a}$. We will see that the resulting estimator has the same expectation and only the variance is slightly increased. These modified estimators are used to center the variables $\RCusCompHidNodeABN{l}{i;a}$ at 0 and scale them so that they have a variance of $\mathcal{O}(1)$, similar to what is done with the classic batch norm. We will examine the differences from the canonical estimators, demonstrating that as the batch size increases, the disparity between the canonical and modified versions becomes negligible.

Specifically, we define the new transformation
\begin{align}
  \RCusCompHidNodeABN{l}{i;a} &= \frac{\RCompHidNodeBAF{i;a}{l} - \BNCusMean{i}{l}}{\BNCusStd{i}{l}}\,, \label{eq:bi_def}\\
    \BNCusMean{i}{l} & = \frac{1}{\BatchSize -1 } \sum_{\substack{b \in \Batch(a) \\ b \neq a}} \RCompHidNodeBAF{i;b}{l}\,,\label{eq:muCus_def}\\
    \BNCusStd{i}{l} & = \sqrt{\frac{1}{\BatchSize -1 } \sum_{\substack{b \in B(a) \\ b \neq a}} \left(\RCompHidNodeBAF{i;b}{l}-\BNCusMean{i}{l} \right)^2}\,.
\end{align}
$\BNCusMean{i}{l}$ and $\BNCusStd{i}{l}$ are alternative estimators. It is straightforward to verify that:
\begin{align}
    \CE{\Dataset}{\BNCusMean{i}{l}} &= \CE{\Dataset}{\BNMean{i}{l}} = \RMean{\RCompHidNodeBAF{i}{l}}\,, \\
     \CE{\Dataset}{\BNCusVar{i}{l}} &=  \RStd{\RCompHidNodeBAF{i}{1}}^2 \left( \frac{\BatchSize-2}{\BatchSize-1} \right) \,.\label{eq:Exp_sigma}
\end{align}

\paragraph{Derivation of Eq.~\eqref{eq:Exp_sigma}}
\begin{align}
\label{eq:sig_til_comp}
\CE{\Dataset}{\BNCusVar{i}{l}} &=
  \CE{\Dataset}{ \frac{1}{\BatchSize -1 } \sum_{\substack{b \in B(a) \\ b \neq a}} \left( \RCompHidNodeBAF{i;b}{l}-\BNCusMean{i}{l} \right)^2 } \\
  &= \frac{1}{\BatchSize -1 } \sum_{\substack{b \in B(a) \\ b \neq a}} \left( \underbrace{ \CE{\Dataset}{\LR{\RCompHidNodeBAF{i;b}{l}}^2}}_\text{\textbf{A}} + \underbrace{\CE{\Dataset}{\LR{\BNCusMean{i}{l}}^2}}_\text{\textbf{B}} -2 \underbrace{\CE{\Dataset}{ \RCompHidNodeBAF{i;b}{l} \,\,\BNCusMean{i}{l}}}_\text{\textbf{C}} \right)
\end{align}
\begin{align}
    \text{\textbf{A}} &=\CE{\Dataset}{\LR{\RCompHidNodeBAF{i;b}{l}}^2} = \RStd{\RCompHidNodeBAF{i}{l}}^2 + \RMean{\RCompHidNodeBAF{i}{l}}^2\,; \label{eq:A}\\
   \notag \text{\textbf{B}} &= \CE{\Dataset}{\LR{\BNCusMean{i}{l}}^2} = \frac{1}{(\BatchSize -1)^2} \sum_{\substack{m \in B(a) \\ m \neq a}} \sum_{\substack{n \in B(a) \\ n \neq a}} \CE{\Dataset}{\RCompHidNodeBAF{i;m}{l} \RCompHidNodeBAF{i;n}{l}} \\
   \notag &= \frac{1}{(\BatchSize -1)^2} \LR{ \sum_{\substack{m,n \in B(a) \\ m \neq a \\ m=n}} \CE{\Dataset}{\LR{\RCompHidNodeBAF{i;m}{l}}^2} + \sum_{\substack{m,n \in B(a) \\ m \neq a \\ m \neq n}} \CE{\Dataset}{\RCompHidNodeBAF{i;m}{l}} \CE{\Dataset}{\RCompHidNodeBAF{i;n}{l}} } \\
     &= \frac{1}{(\BatchSize -1)^2} \left( (\BatchSize -1) \LR{ \RStd{\RCompHidNodeBAF{j}{l}}^2 + \RMean{\RCompHidNodeBAF{j}{l}}^2 } + (\BatchSize -1)(\BatchSize -2) \RMean{\RCompHidNodeBAF{i}{l}}^2 \right) = \frac{\RStd{\RCompHidNodeBAF{j}{l}}^2}{(\BatchSize -1)} + \RMean{\RCompHidNodeBAF{j}{l}}^2\,;  \label{eq:B} \\
    \notag \text{\textbf{C}} &= \frac{1}{(\BatchSize -1)} \CE{\Dataset}{ \RCompHidNodeBAF{i;b}{l} \sum_{\substack{m \in B(a) \\ m \neq a}} \RCompHidNodeBAF{i;m}{l}} = \frac{1}{(\BatchSize -1)}  \LR{\CE{\Dataset}{\LR{\RCompHidNodeBAF{i;b}{l}}^2} +  \sum_{\substack{m \in B(a) \\ m \neq a,b}} \CE{\Dataset}{\RCompHidNodeBAF{i;b}{l}}\CE{\Dataset}{\RCompHidNodeBAF{i;m}{l}}} \\ 
    &= \frac{1}{(\BatchSize -1)} \LR{ \LR{\RStd{\RCompHidNodeBAF{j}{l}}^2 + \RMean{\RCompHidNodeBAF{j}{l}}^2} + \LR{(\BatchSize -2) \RMean{\RCompHidNodeBAF{j}{l}}^2}} = \frac{\sigma^2}{(\BatchSize -1)} + \RMean{\RCompHidNodeBAF{j}{l}}^2\,.
    \label{eq:C}
\end{align}

Substituting Eqs.~\eqref{eq:A}, \eqref{eq:B} and \eqref{eq:C} into Eq.~\eqref{eq:sig_til_comp} we get:
\begin{align}
   \CE{\Dataset}{\BNCusVar{i}{l}} = \LR{\RStd{\RCompHidNodeBAF{j}{l}}^2 + \RMean{\RCompHidNodeBAF{j}{l}}^2} + \left( \frac{\RStd{\RCompHidNodeBAF{j}{l}}^2 }{(\BatchSize -1)} + \RMean{\RCompHidNodeBAF{j}{l}}^2 \right) -2 \left(  \frac{\RStd{\RCompHidNodeBAF{j}{l}}^2}{(\BatchSize -1)} + \RMean{\RCompHidNodeBAF{j}{l}}^2 \right) = \RStd{\RCompHidNodeBAF{j}{l}}^2 \frac{(\BatchSize -2)}{(\BatchSize -1)}\,.
\end{align}

Furthermore, it is possible to express the introduced estimators as functions of the random variables $\BNMean{i}{l}$ and $\BNStd{i}{l}$ showing that the difference between the two choices tends to 0 (in probability) as the batch size $\BatchSize$ increases. For example:
\begin{align}
    \BNMean{i}{l}= \BNCusMean{i}{l} \left( \frac{\BatchSize -1}{\BatchSize}  \right) + \frac{\RCompHidNodeBAF{i;a}{l}}{\BatchSize} \Rightarrow \Dmu \equiv \BNMean{i}{l} - \BNCusMean{i}{l} = \frac{1}{\BatchSize} \left(\RCompHidNodeBAF{i;a}{l} - \BNCusMean{i}{l}  \right) \equiv \eB\,.
\end{align}
Analyzing the random variable $\eB$, we can show that the difference between the two estimators converges to 0 in probability, in particular:
\begin{align}
    \CE{\Dataset}{\eB} &= \frac{1}{\BatchSize} \LR{ \CE{\Dataset}{\RCompHidNodeBAF{i;a}{l}} -  \CE{\Dataset}{\BNCusMean{i}{l}}} = \frac{1}{\BatchSize} ( \RMean{\RCompHidNodeBAF{i}{l}} - \RMean{\RCompHidNodeBAF{i}{l}}) = 0\,, \\
    \CVar{}{\eB} &= \frac{1}{\BatchSize^2} \left( \RStd{\RCompHidNodeBAF{i}{l}}^2 + \frac{\BatchSize -1}{\BatchSize^2} \RStd{\RCompHidNodeBAF{i}{l}}^2 \right) = \mathcal{O} \left( \frac{1}{\BatchSize^2} \right)\,.
\end{align}
Similarly, the same can be shown for $\Dsigma \equiv \RProxyStd{\RCompHidNodeBAF{i}{l}}^2 - \RCusProxyStd{\RCompHidNodeBAF{i}{l}}^2$. \newline
We can also derive an expression for the distributions $\pdf{\BNCusStd{i}{l}}{}{y}$, $\pdf{\BNCusMean{i}{l}}{}{h}$, and $\pdf{\RCompHidNodeBAF{i}{l}}{}{w}$.
\paragraph{\texorpdfstring{$\boldsymbol{\pdf{\RCompHidNodeBAF{i}{l}}{}{w}}$}{Pw(xi)}:} Let us assume that the variables we are transforming with the normalization are normally distributed, \textit{i.e.}:
\begin{equation}
    \pdf{\RCompHidNodeBAF{i}{l}}{}{w} = \frac{1}{\sqrt{2 \pi \sigma^2}} \exp{\LR{-\frac{\LR{w-\RMean{\RCompHidNodeBAF{i}{l}}}^2}{2 \RStd{\RCompHidNodeBAF{}{l}}^2}}}\,,
\end{equation}
with \( \RStd{\RCompHidNodeBAF{i}{l}}^2 = \RStd{\RCompHidNodeBAF{}{l}}^2 \,\,\, \forall i\).

We will derive the other relevant distributions, included the p.d.f. of $\RCusCompHidNodeABN{l}{i}$ starting from this assumption. We will then use this result to complete the proof in the IGB setting, \textit{i.e.} assuming random data as input.

\paragraph{Derivation of \texorpdfstring{$\boldsymbol{\pdf{\BNCusMean{i}{l}}{}{h}}$ :}{Pw(f0)}} $\RCusProxyMean{}$ is a sum of $(\BatchSize -1)$ i.i.d. Gaussian variables [Eq.~\eqref{eq:muCus_def}]; thus, it is also distributed normally. Specifically:
\begin{align}
    \pdf{\BNCusMean{i}{l}}{}{h} = \sqrt{\frac{(\BatchSize -1)}{2 \pi \RStd{\RCompHidNodeBAF{}{l}}^2}} \exp\LR{- \frac{- y^2}{2} \frac{(\BatchSize -1)}{\RStd{\RCompHidNodeBAF{}{l}}^2}}\,.
\end{align}

\paragraph{Derivation of \texorpdfstring{$\boldsymbol{\pdf{\BNCusStd{i}{l}}{}{y} :}$}{Pw(f0)}}
To derive the p.d.f., we start from the sample variance calculated on $(\BatchSize -1)$ elements, which is distributed according to a chi-squared distribution with $(\BatchSize -2)$ degrees of freedom. Specifically, defining
\begin{align}
    S = \BNCusVar{i}{l} \frac{(\BatchSize -1)}{\RStd{\RCompHidNodeBAF{}{l}}^2} = \frac{1}{\RStd{\RCompHidNodeBAF{}{l}}^2} \sum^{\BatchSize-1}_{\substack{j \in B \\ j \neq i}} (\RCompHidNodeBAF{j}{l}-\BNCusMean{i}{l})^2\,,
\end{align}
we have $S \sim \chi (\BatchSize - 2)$. Since $S$ is proportional to $\RCusProxyStd{\RCompHidNodeBAF{i}{l}}^2$, with a simple change of variables, we obtain
\begin{align}
    \pdf{\BNCusVar{i}{l}}{}{y'} = \frac{1}{2^{\frac{(\BatchSize -2)}{2}} \Gamma \left( \frac{\BatchSize -2}{2} \right)} \left( \frac{\BatchSize -1}{\RStd{\RCompHidNodeBAF{}{l}}^2} \right)^{\frac{\BatchSize}{2} -1} (y')^{\frac{\BatchSize}{2} - 2} \,\, \exp\LR{-\frac{y'}{2} \frac{(\BatchSize -1)}{\RStd{\RCompHidNodeBAF{}{l}}^2}}\,.
\end{align}
Finally, with a further change of variables, we obtain
\begin{align}
    \pdf{\BNCusStd{i}{l}}{}{y} = \frac{1}{2^{\frac{\BatchSize}{2} -2} \Gamma \left( \frac{\BatchSize -2}{2} \right)} \left( \frac{\BatchSize -1}{\RStd{\RCompHidNodeBAF{}{l}}^2} \right)^{\frac{\BatchSize}{2} -1} y^{\BatchSize - 3} \, \, \, \exp\LR{-\frac{y^2}{2} \frac{(\BatchSize -1)}{\RStd{\RCompHidNodeBAF{}{l}}^2}}\,.
\end{align}
It is noteworthy that the three random variables $\RCompHidNodeBAF{i;a}{l}$, $\BNCusMean{i}{l}$, and $\BNCusStd{i}{l}$ are independent due to the independence between the sample mean and sample variance of Gaussian variables (See, \textit{e.g.}, Theorem 5.3.1 in \cite{casella2024statistical}), and because $\BNCusMean{i}{l}$ and $\BNCusStd{i}{l}$ are functions of a set of variables independent from $\RCompHidNodeBAF{i;a}{l}$.
This independence among the three random variables significantly simplifies the analysis. For example, we can simply calculate $\CE{\Dataset}{\RCusCompHidNodeABN{l}{i}}$ and $\CVar{\Dataset}{\RCusCompHidNodeABN{l}{i}}$ as follows:
\begin{align}
    \CE{\Dataset}{\RCusCompHidNodeABN{l}{i}} &= \int_{0}^{\infty} dy \int_{-\infty}^{\infty} dx \int_{-\infty}^{\infty} dw \; \frac{w-x}{y} \pdf{\BNCusStd{i}{l}}{}{y} \pdf{\BNCusMean{i}{l}}{}{x} \pdf{\RCompHidNodeBAF{i}{l}}{}{w} \notag \\
    &=  \int_{0}^{\infty} dy \; \frac{\pdf{\BNCusStd{i}{l}}{}{y}}{y} \int_{-\infty}^{\infty} dx \; \pdf{\BNCusMean{i}{l}}{}{x} \int_{-\infty}^{\infty} dw \; (w-x)   \pdf{\RCompHidNodeBAF{i}{l}}{}{w} \notag\\
&=  \int_{0}^{\infty} dy \; \frac{\pdf{\BNCusStd{i}{l}}{}{y}}{y}   \int_{-\infty}^{\infty} dx \; \pdf{\BNCusMean{i}{l}}{}{x} (w-\RMean{\RCompHidNodeBAF{i}{l}}) \notag \\
& =  \int_{0}^{\infty} dy \; \frac{\pdf{\BNCusStd{i}{l}}{}{y}}{y} (\RMean{\RCompHidNodeBAF{i}{l}} - \RMean{\RCompHidNodeBAF{i}{l}}) = 0\,,  \\
\CVar{\Dataset}{\RCusCompHidNodeABN{l}{i}} &= \CE{\Dataset}{\LR{\RCusCompHidNodeABN{l}{i}}^2} = \int_{0}^{\infty} dy \int_{-\infty}^{\infty} dx \int_{-\infty}^{\infty} dw \; \frac{(w-x)^2}{y^2} \pdf{\BNCusStd{i}{l}}{}{y} \pdf{\BNCusMean{i}{l}}{}{x} \pdf{\RCompHidNodeBAF{i}{l}}{}{w}  \notag \\
&= \int_{0}^{\infty} dy \; \frac{\pdf{\BNCusStd{i}{l}}{}{y} }{y^2} \int_{-\infty}^{\infty} dx \; \pdf{\BNCusMean{i}{l}}{}{x} \int_{-\infty}^{\infty} dw \;    \pdf{\RCompHidNodeBAF{i}{l}}{}{w} (w^2 + x^2 -2 xw)  \notag \\
&= \int_{0}^{\infty} dy \;   \frac{1}{2^{\frac{\BatchSize}{2} -2 } \Gamma \left( \frac{\BatchSize -2 }{2}\right)} \left( \frac{\BatchSize -1}{\RStd{\RCompHidNodeBAF{}{l}}^2} \right)^{\frac{\BatchSize}{2} -1} y^{\BatchSize - 5} \exp\LR{-\frac{y^2}{2} \frac{(\BatchSize - 1)}{\RStd{\RCompHidNodeBAF{}{l}}^2}} \cdot  \notag \\
\cdot & \left( \RStd{\RCompHidNodeBAF{}{l}}^2 + \RMean{\RCompHidNodeBAF{i}{l}}^2 + \frac{\RStd{\RCompHidNodeBAF{}{l}}^2}{\BatchSize -1} + \RMean{\RCompHidNodeBAF{i}{l}}^2 -2 \RMean{\RCompHidNodeBAF{i}{l}}^2 \right) \notag \\
&\stackrel{\boldsymbol{a}}{=} \RStd{\RCompHidNodeBAF{}{l}}^2 \frac{\BatchSize}{\BatchSize -1 }  \frac{1}{2^{\frac{\BatchSize}{2} -2 } \Gamma \left( \frac{\BatchSize -2 }{2}\right)} \left( \frac{\BatchSize -1}{\RStd{\RCompHidNodeBAF{}{l}}^2} \right)^{\frac{\BatchSize}{2} -1} \frac{\Gamma \left( \frac{\BatchSize -4 }{2}\right)}{2} \left( \frac{2 \RStd{\RCompHidNodeBAF{}{l}}^2}{(\BatchSize -1)}\right)^{\frac{\BatchSize -4}{2}} \notag \\
&= \frac{\BatchSize}{\BatchSize -4}\,. \label{eq:Var_b_i}
\end{align}

where in step $\boldsymbol{a}$ we used the identity
\begin{align}
     \int_{0}^{\infty} x^n e^{-a x^b} dx = \frac{\Gamma \left( \frac{n+1}{b} \right)}{b a^{\frac{(n+1)}{b}}}\,.
\end{align}

The use of the estimators $\BNCusMean{i}{l}$ and $\BNCusStd{i}{l}$ greatly simplifies the analysis and allows us, for example, to derive the p.d.f. of the random variable $\RCusCompHidNodeABN{l}{i}$, distribution $\pdf{\RCusCompHidNodeABN{l}{i}}{(\Dataset)}{z}$, defined by the transformation in Eq.~\eqref{eq:bi_def}. Specifically:

\begin{align}\label{eq:b_int_expr}
    \pdf{\RCusCompHidNodeABN{l}{i}}{(\Dataset)}{\ArgHidNodeCusABN{}} &= \int_{0}^{\infty} dy \int_{-\infty}^{\infty} dx  \int_{-\infty}^{\infty} dw \Dirac{\ArgHidNodeCusABN{} - \frac{(w -x)}{y}} \pdf{\BNCusStd{i}{l}}{}{y} \pdf{\BNCusMean{i}{l}}{}{x} \pdf{\RCompHidNodeBAF{i}{l}}{}{w}  \notag \\
    &= C \int_{0}^{\infty} dy \int_{-\infty}^{\infty} dx  \int_{-\infty}^{\infty} dw \; \Dirac{\ArgHidNodeCusABN{} - \frac{(w -x)}{y}} \cdot \notag \\
    & \exp\LR{-\frac{(w-\RMean{\RCompHidNodeBAF{i}{l}})^2}{2 \RStd{\RCompHidNodeBAF{i}{l}}^2}} \exp\LR{-\frac{\ArgHidNodeAAF-\RMean{\RCompHidNodeBAF{i}{l}})^2}{2 \RStd{\RCompHidNodeBAF{i}{l}}^2} (\BatchSize -1)} y^{\BatchSize -3} \exp\LR{- \frac{y^2}{2} \frac{(\BatchSize -1)}{\RStd{\RCompHidNodeBAF{i}{l}}^2}} \notag \\
    &=  C \int_{0}^{\infty} dy \; y^{\BatchSize -3} e^{- \frac{ y^2}{2} \frac{(\BatchSize -1)}{\RStd{\RCompHidNodeBAF{i}{l}}^2}} y \int_{-\infty}^{\infty} dx  \int_{-\infty}^{\infty} dw \; \Dirac{\ArgHidNodeCusABN{}y - (w -x)} \cdot \notag \\
   & \exp\LR{-\frac{(w-\RMean{\RCompHidNodeBAF{i}{l}})^2}{2 \RStd{\RCompHidNodeBAF{i}{l}}^2}} \exp\LR{-\frac{(\ArgHidNodeAAF-\RMean{\RCompHidNodeBAF{i}{l}})^2}{2 \RStd{\RCompHidNodeBAF{i}{l}}^2} (\BatchSize -1)}\notag \\
    &= C \int_{0}^{\infty} dy \; y^{\BatchSize -2} \exp\LR{- \frac{- y^2}{2} \frac{(\BatchSize -1)}{\RStd{\RCompHidNodeBAF{i}{l}}^2}} \int_{-\infty}^{\infty} dx  \int_{-\infty}^{\infty} dw \; \Dirac{w - (\ArgHidNodeCusABN{}y +x)} \cdot \notag \\
    & \exp\LR{-\frac{(w-\RMean{\RCompHidNodeBAF{i}{l}})^2}{2 \RStd{\RCompHidNodeBAF{i}{l}}^2}} \exp\LR{-\frac{\ArgHidNodeAAF-\RMean{\RCompHidNodeBAF{i}{l}})^2}{2 \RStd{\RCompHidNodeBAF{i}{l}}^2} (\BatchSize -1)} \notag \\
    &= C \int_{0}^{\infty} dy \; y^{\BatchSize -2} \exp\LR{- \frac{ y^2}{2} \frac{(\BatchSize -1)}{\RStd{\RCompHidNodeBAF{i}{l}}^2}} \cdot \notag \\
   & \int_{-\infty}^{\infty} dx \; \exp\LR{-\frac{(\ArgHidNodeCusABN{}y + x -\RMean{\RCompHidNodeBAF{i}{l}})^2}{2 \RStd{\RCompHidNodeBAF{i}{l}}^2}} \exp\LR{-\frac{\ArgHidNodeAAF-\RMean{\RCompHidNodeBAF{i}{l}})^2}{2 \RStd{\RCompHidNodeBAF{i}{l}}^2} (\BatchSize -1)} \notag \\
    &= C \int_{0}^{\infty} dy \; y^{\BatchSize -2} \exp\LR{- \frac{ y^2}{2} \frac{(\BatchSize -1)}{\RStd{\RCompHidNodeBAF{i}{l}}^2}} \cdot \notag \\
    & \int_{-\infty}^{\infty} dx \;  \exp\LR{- \frac{1}{2 \RStd{\RCompHidNodeBAF{i}{l}}^2} (\ArgHidNodeCusABN{}^2y^2 + x^2 + 2 xy\ArgHidNodeCusABN{} + (\BatchSize -1) x^2)}  \notag \\
    &=  C \int_{0}^{\infty} dy \; y^{\BatchSize -2} \exp\LR{- \frac{ y^2}{2} \frac{(\BatchSize -1)}{\RStd{\RCompHidNodeBAF{i}{l}}^2}} \cdot \notag \\ 
    & \int_{-\infty}^{\infty} dx \;  \exp\LR{- \frac{1}{2 \RStd{\RCompHidNodeBAF{i}{l}}^2} \left( \ArgHidNodeCusABN{}^2y^2  \left( 1 - \frac{1}{\BatchSize}\right) \right)} \exp\LR{-\frac{1}{2 \RStd{\RCompHidNodeBAF{i}{l}}^2} \left(\sqrt{\BatchSize} x + \frac{y\ArgHidNodeCusABN{}}{\sqrt{\BatchSize}}\right)^2} \notag \\
    &= C' \int_{0}^{\infty} dy \; y^{\BatchSize -2} \exp\LR{- \frac{ y^2}{2} \frac{(\BatchSize -1)}{\RStd{\RCompHidNodeBAF{i}{l}}^2} \left( 1 + \frac{\ArgHidNodeCusABN{}^2}{\BatchSize}\right)} \notag \\
    &= C' \left( \frac{(\BatchSize -1)}{2 \RStd{\RCompHidNodeBAF{i}{l}}^2}\right)^{-\frac{( \BatchSize -1)}{2}} \left( 1 + \frac{\ArgHidNodeCusABN{}^2}{\BatchSize}\right)^{-\frac{( \BatchSize -1)}{2}} \Gamma \left( \frac{\BatchSize -1 }{2}\right) \frac{1}{2} \notag \\
    &= \sqrt{\frac{1}{\BatchSize \pi}}  \frac{\Gamma \left( \frac{\BatchSize -1}{2}\right)}{\Gamma \left( \frac{\BatchSize -2}{2}\right)} \left( 1 + \frac{\ArgHidNodeCusABN{}^2}{\BatchSize}\right)^{-\frac{(\BatchSize -1)}{2}} \, ,
\end{align}
where
\begin{align}
    C &= \frac{\sqrt{\BatchSize -1}}{(2 \pi \RStd{\RCompHidNodeBAF{i}{l}}^2)} \frac{1}{2^{\frac{\BatchSize}{2} - 2} \Gamma \left( \frac{\BatchSize -2}{2} \right)} \left( \frac{\BatchSize -1}{\RStd{\RCompHidNodeBAF{i}{l}}^2} \right)^{\frac{\BatchSize}{2} -1} \notag \\
    C' &= C \sqrt{\frac{2 \pi \RStd{\RCompHidNodeBAF{i}{l}}^2}{\BatchSize}} \, .
\end{align}

The distribution outlined in Eq.~\eqref{eq:b_int_expr} is expected to converge to the Gaussian profile observed in the full batch scenario. To rigorously demonstrate this assertion, we explicitly expand the binomial term in Eq.~\eqref{eq:b_int_expr} in the limit as the batch size $\BatchSize$ approaches infinity. This expansion allows us to directly compare each power of $\ArgHidNodeCusABN{}$ in the binomial series with the corresponding term in the Taylor series expansion of the Gaussian profile (centered around zero) that emerges in the full-batch case. By examining each power individually, we show that, in the full batch limit, every term in the expanded form of Eq.~\eqref{eq:b_int_expr} precisely matches its counterpart in the Gaussian Taylor series. This one-to-one correspondence confirms the convergence of the expression to a standard Gaussian as $\BatchSize \to \infty$.

To demonstrate that the expression in Eq.~\eqref{eq:b_int_expr} converges to a Gaussian as $\BatchSize \to \infty$, we expand the term
\[
\left( 1 + \frac{\ArgHidNodeCusABN{}^2}{\BatchSize} \right)^{-\frac{\BatchSize - 1}{2}}
\]
using the binomial series expansion. For any $\alpha \in \mathbb{R}$, the binomial expansion for $(1 + x)^{\alpha}$ is
\[
(1 + x)^{\alpha} = \sum_{k=0}^{\infty} \binom{\alpha}{k} x^k,
\]
where $\binom{\alpha}{k} = \frac{\alpha (\alpha - 1) \cdots (\alpha - k + 1)}{k!}$. In our case, we set $x = \frac{\ArgHidNodeCusABN{}^2}{\BatchSize}$ and $\alpha = -\frac{\BatchSize - 1}{2}$, giving
\[
\left(1 + \frac{\ArgHidNodeCusABN{}^2}{\BatchSize}\right)^{-\frac{\BatchSize - 1}{2}} = \sum_{k=0}^{\infty} \binom{-\frac{\BatchSize - 1}{2}}{k} \left(\frac{\ArgHidNodeCusABN{}^2}{\BatchSize}\right)^k.
\]
The general term in this expansion is
\[
\binom{-\frac{\BatchSize - 1}{2}}{k} \left(\frac{\ArgHidNodeCusABN{}^2}{\BatchSize}\right)^k.
\]
Using the definition of the binomial coefficient for negative exponents, we have
\[
\binom{-\frac{\BatchSize - 1}{2}}{k} = \frac{(-1)^k \left(\frac{\BatchSize - 1}{2}\right) \left(\frac{\BatchSize - 1}{2} + 1\right) \cdots \left(\frac{\BatchSize - 1}{2} + k - 1\right)}{k!}.
\]
For large $\BatchSize$, each term in the product $\left(\frac{\BatchSize - 1}{2} + j\right) \approx \frac{\BatchSize}{2}$, so we approximate
\[
\binom{-\frac{\BatchSize - 1}{2}}{k} \approx \frac{(-1)^k}{k!} \left(\frac{\BatchSize}{2}\right)^k.
\]
Substituting this approximation into the general term, we find
\[
\binom{-\frac{\BatchSize - 1}{2}}{k} \left(\frac{\ArgHidNodeCusABN{}^2}{\BatchSize}\right)^k \approx \frac{(-1)^k}{k!} \frac{\left(\frac{\BatchSize}{2}\right)^k \left(\frac{\ArgHidNodeCusABN{}^2}{\BatchSize}\right)^k}{\BatchSize^k} = \frac{(-1)^k}{k!} \frac{\ArgHidNodeCusABN{}^{2k}}{2^k}.
\]
Thus, in the limit $\BatchSize \to \infty$, we recover the series expansion of the Gaussian function:
\[
\exp\LR{-\frac{\ArgHidNodeCusABN{}^2}{2}} = \sum_{k=0}^{\infty} \frac{(-1)^k}{k!} \left(\frac{\ArgHidNodeCusABN{}^2}{2}\right)^k.
\]
This demonstrates that each term in the expansion of Eq.~\eqref{eq:b_int_expr} matches the corresponding term in the Gaussian Taylor series in the full batch limit $\BatchSize \to \infty$.

Finally, let us consider the prefactor in Eq.~\eqref{eq:b_int_expr},
\[
\sqrt{\frac{1}{\BatchSize \pi}} \frac{\Gamma \left( \frac{\BatchSize -1}{2} \right)}{\Gamma \left( \frac{\BatchSize -2}{2} \right)}.
\]
Using Stirling's approximation for large $\BatchSize$, we have
\[
\Gamma\left(\frac{\BatchSize - 1}{2}\right) \approx \sqrt{2 \pi} \left(\frac{\BatchSize - 1}{2}\right)^{\frac{\BatchSize - 1}{2} - \frac{1}{2}} e^{-\frac{\BatchSize - 1}{2}},
\]
and similarly,
\[
\Gamma\left(\frac{\BatchSize - 2}{2}\right) \approx \sqrt{2 \pi} \left(\frac{\BatchSize - 2}{2}\right)^{\frac{\BatchSize - 2}{2} - \frac{1}{2}} e^{-\frac{\BatchSize - 2}{2}}.
\]
Thus,
\[
\frac{\Gamma\left(\frac{\BatchSize - 1}{2}\right)}{\Gamma\left(\frac{\BatchSize - 2}{2}\right)} \approx \sqrt{\frac{\BatchSize}{2}}.
\]
Combining this result with the factor $\sqrt{\frac{1}{\BatchSize \pi}}$ gives
\[
\sqrt{\frac{1}{\BatchSize \pi}} \cdot \sqrt{\frac{\BatchSize}{2}} = \frac{1}{\sqrt{2 \pi}},
\]
which is precisely the normalization factor of a standard Gaussian distribution. 

Therefore, in the limit $\BatchSize \to \infty$, Eq.~\eqref{eq:b_int_expr} converges to
\[
\frac{1}{\sqrt{2 \pi}} \exp\LR{-\frac{\ArgHidNodeCusABN{}^2}{2}},
\]
confirming that the expression approaches a standard Gaussian distribution as $\BatchSize \to \infty$.

\begin{figure}[]
    \centering
\includegraphics[width=.45\textwidth]{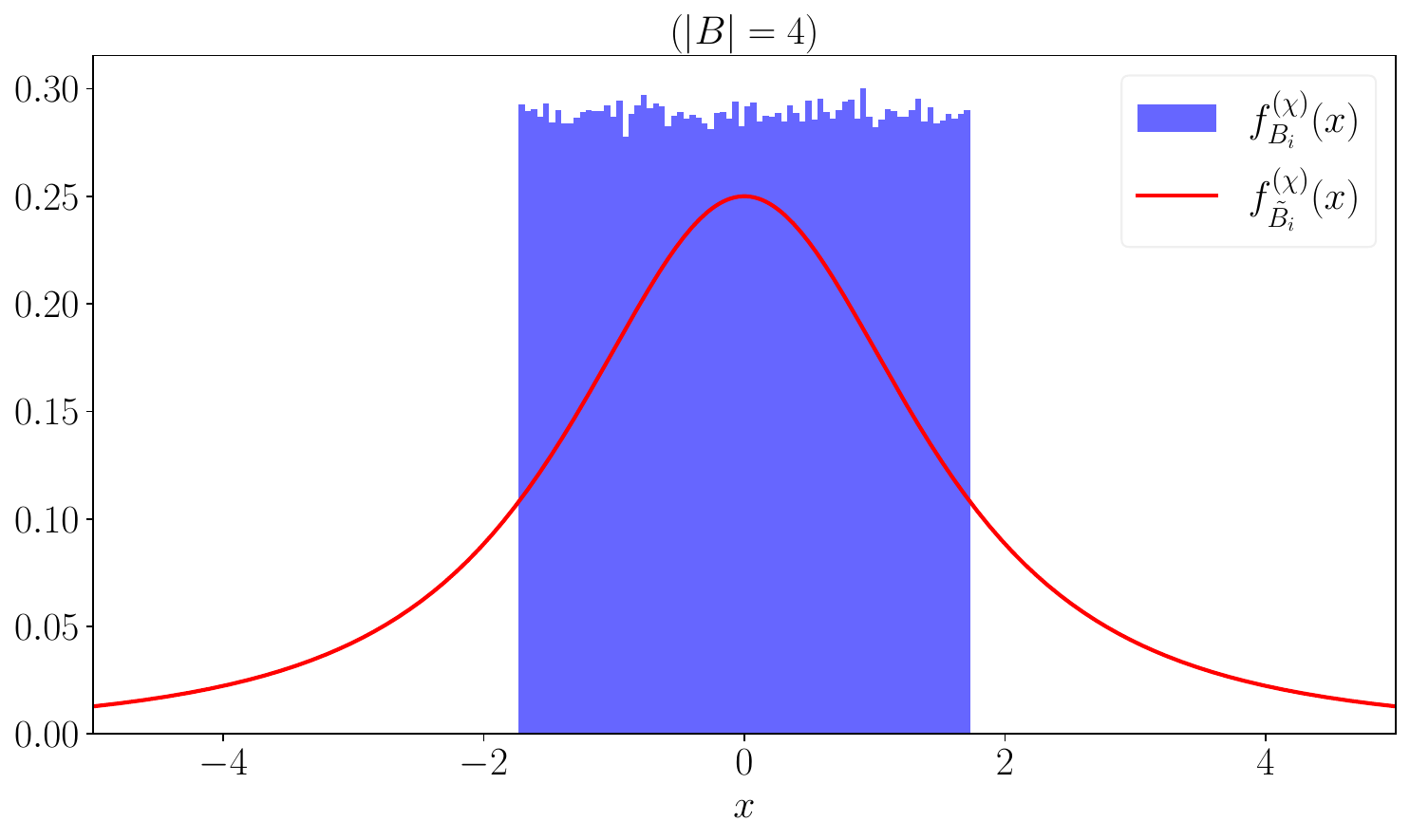} 
\includegraphics[width=.45\textwidth]{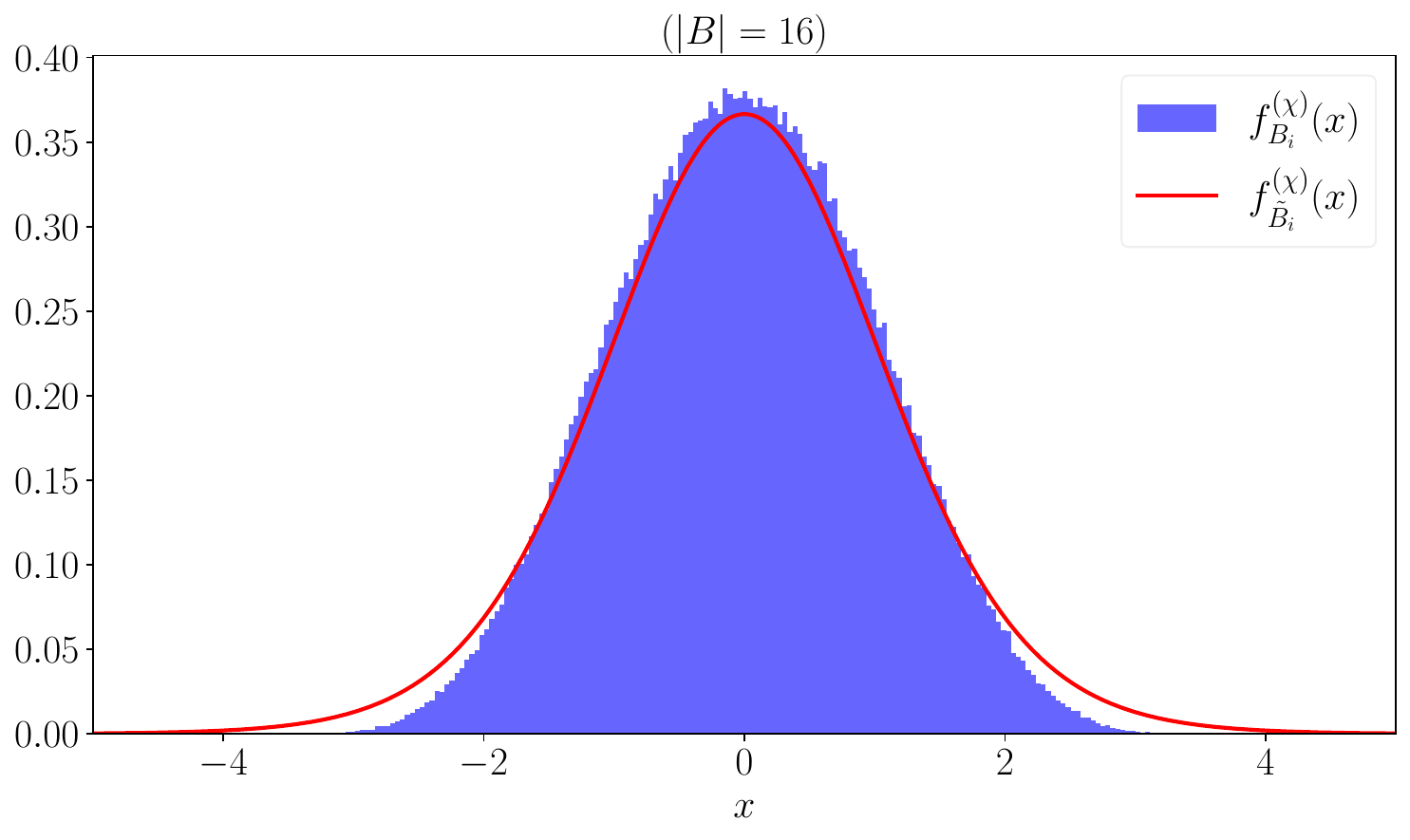} 
\includegraphics[width=.45\textwidth]{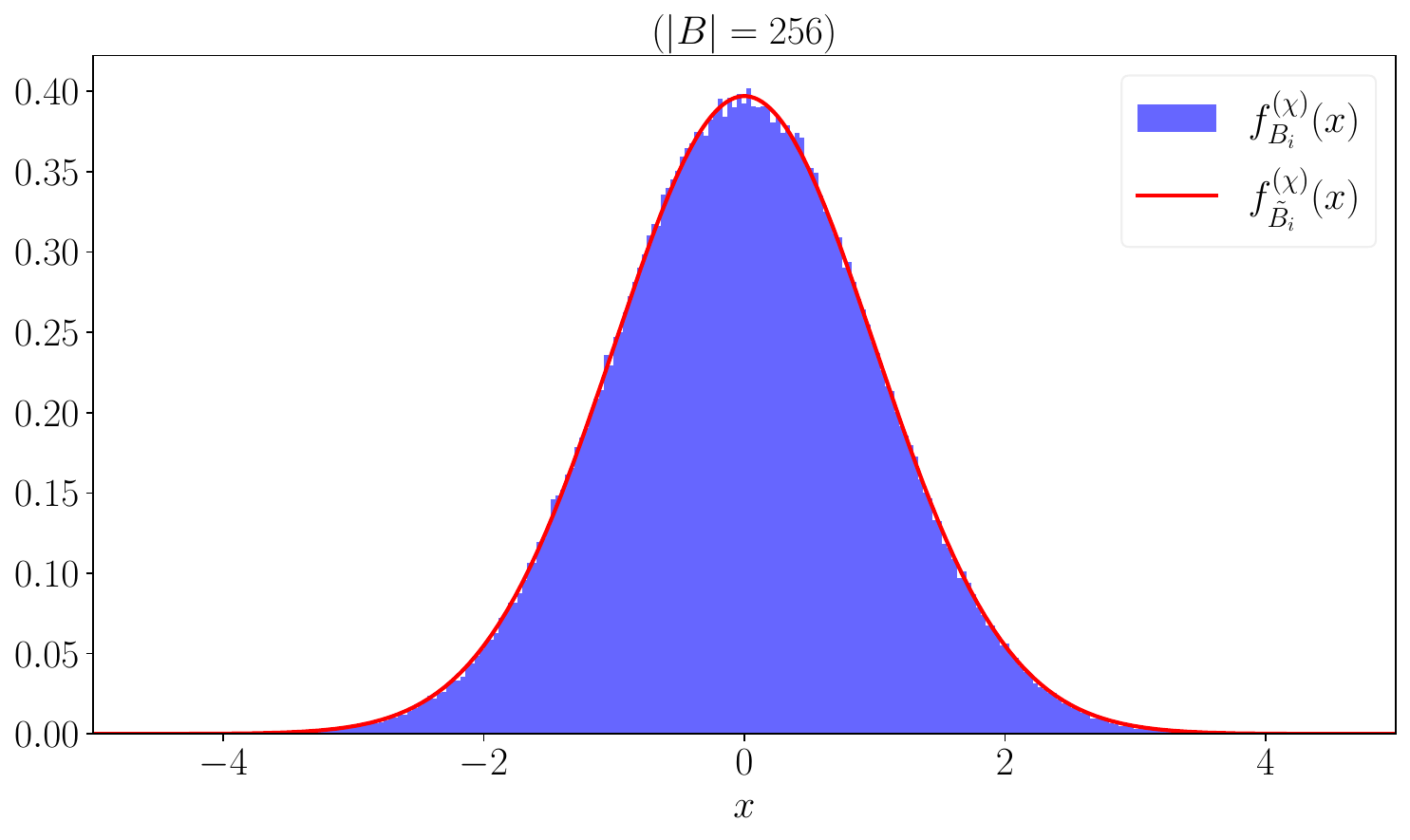}
    \caption{Comparison, for different batch sizes, between the empirical p.d.f. of $\RCompHidNodeABN{l}{i}$ (following the canonical batch norm transformation) and the theoretical p.d.f. of $\RCusCompHidNodeABN{l}{i}$ [Eq.\eqref{eq:bi_def}]. }
        \label{fig:btilde_comp}%
\end{figure}

\begin{figure}[]
    \centering 
\includegraphics[width=.45\textwidth]{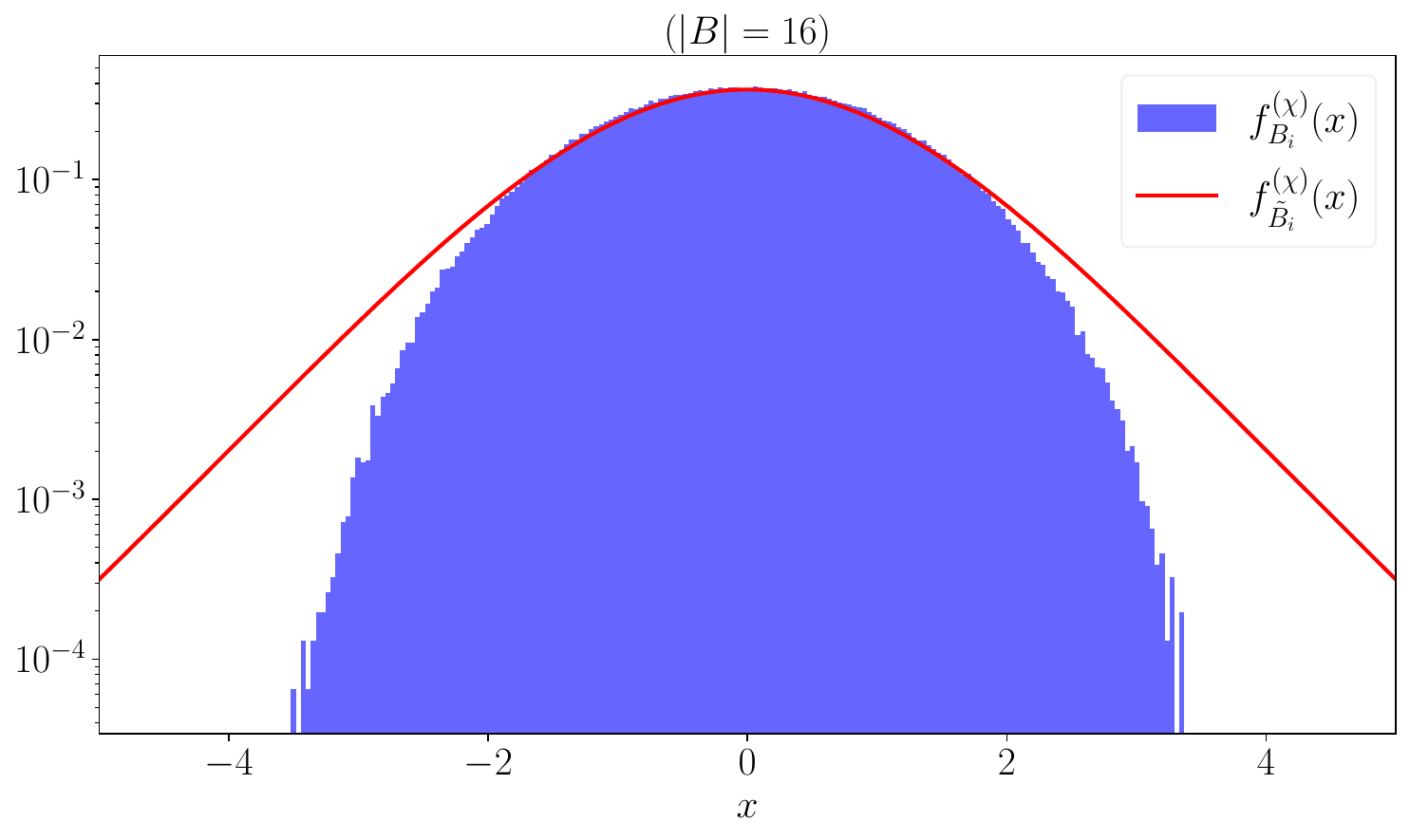} 
\includegraphics[width=.45\textwidth]{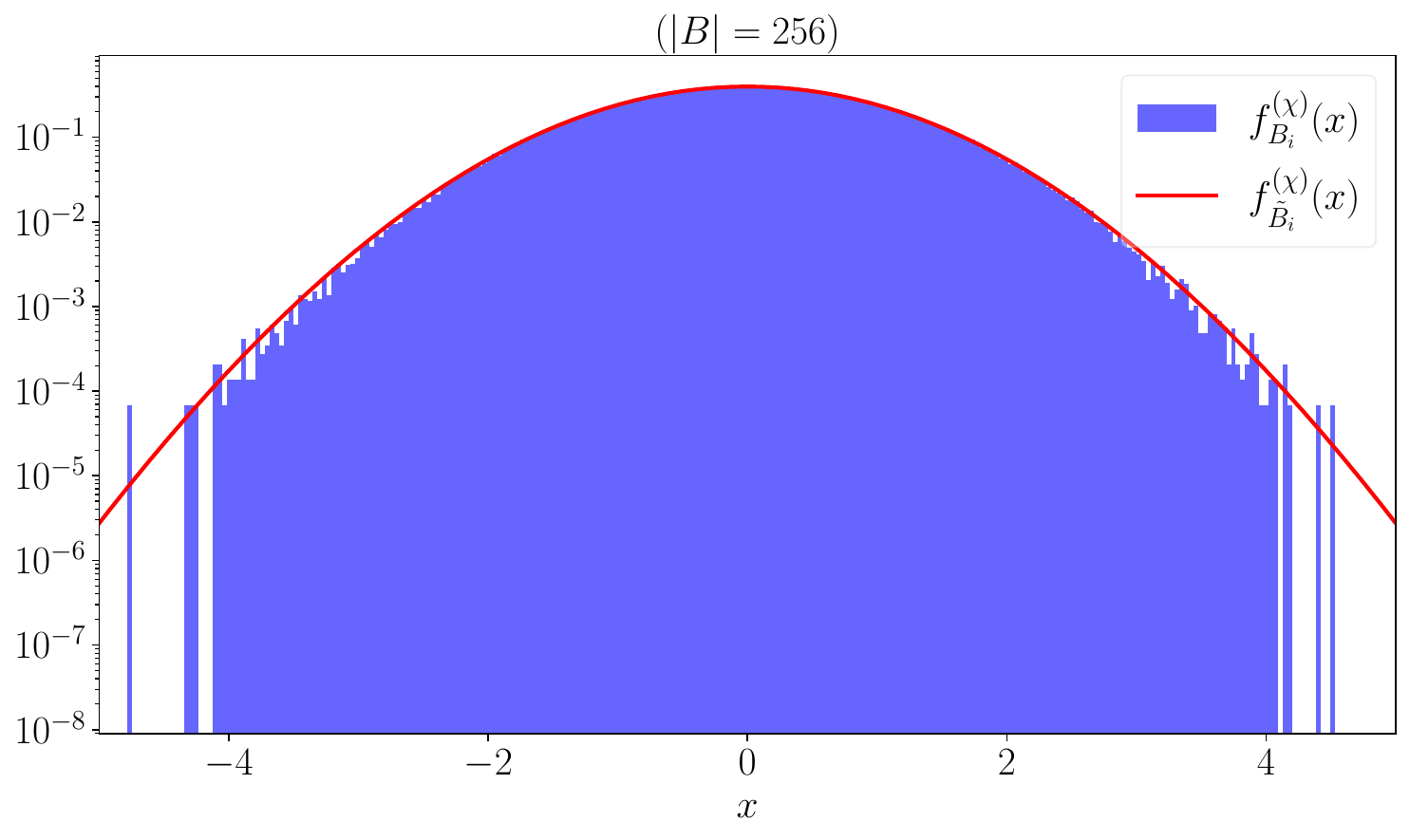}
    \caption{Same comparison of Fig.~\ref{fig:btilde_comp} in log scale, to better visualize the tails density.}
    \label{fig:btilde_comp_log}
\end{figure}

Figure~\ref{fig:btilde_comp} illustrates a comparison between the empirical distribution of Gaussian random variables after processing through a batch normalization layer and the theoretical p.d.f. as derived from Eq.\eqref{eq:b_int_expr}. As expected, increasing the batch size progressively minimizes the differences between these distributions until they become virtually identical. Figure~\ref{fig:btilde_comp_log} shows how the disparity between the distributions increases from the center toward the tails. It is critical to recognize that the tails of the distribution cannot be fully explored empirically due to the dataset's finite size. Thus, with a limited dataset, as the batch size increases, a point is reached where the differences between the distributions are no longer observable within the empirically accessible range. This phenomenon is depicted in Figure~\ref{fig:btilde_comp_log} (right), where the empirically accessible region covers the interval $[-5, 5]$. In this interval, the distributions are perfectly aligned.

With the p.d.f. $\pdf{\RCusCompHidNodeABN{l}{i}}{(\Dataset)}{\ArgHidNodeCusABN{}}$ established, the following process is undertaken:
\begin{enumerate}[I.]
\item Begin with datapoints that are normally distributed.
\item Input these datapoints (organized into mini-batches) into a fully connected layer (using Gaussian weights).
\item Process these through the batch normalization layer.
\item Process through the activation function (ReLU).
\item (For deep architectures, iteratively repeat steps II through IV.)
\item Repeat step II to link the last hidden layer to the output layer.
\item Obtain the $\pdf{\RClassFraction{0}}{}{\ClassFraction{0}}$.
\end{enumerate}

We will follow now the above propagation process step by step deriving the output nodes distribution. We will show that the distributions of the output nodes do not change with depth, \textit{i.e.} IGB does not amplify with depth.
Let us start considering a random vector as input, whose components $\{\RCompInputValue{i}{}  \}_{i=1}^{\LayerNumNodes{0}} $ are distribuited according to  
\begin{equation}
    \RCompInputValue{i}{} \sim \NormalDistr{\mu_i}{\sigma_i^2} \, .
\end{equation}
After passing throught the fully connected layer we will have a set of random variables $\{ \RCompHidNodeBAF{i}{1} \}_{i=1}^{\LayerNumNodes{1}} $, with 

\begin{equation}\label{eq:h1_def}
    \RCompHidNodeBAF{i}{1} = \sum_{j} \Weights{1}{ij} \RCompInputValue{i}{} \, , 
\end{equation}
where 
\begin{equation}
    \Weights{l}{ij} \sim \NormalDistr{0}{\frac{\StdKaiming^2}{\LayerNumNodes{l-1}}} \forall \,l, i, k
\end{equation}
are random weights fixed with the initialization; this means that for a given initialization they will form a set of random coefficients. Therefore, from Eq.~\eqref{eq:h1_def}, $\RCompHidNodeBAF{i}{1}$ is a linear combination of independent Gaussian random variables, meaning that  will be itself normally distributed. In particular:
\begin{equation}\label{eq:h1_dist}
    \RCompHidNodeBAF{i}{1} \sim \NormalDistr{\sum_j \Weights{1}{ij} \mu_j}{\sum_j (\Weights{1}{ij} \sigma_j)^2}\,.
\end{equation}
Note that, for our analysis we don't need to compute explicitly the mean and variance of the distribution in Eq.~\eqref{eq:h1_dist}.
In fact, as a next step, we have to process the set of pre-activated random variables $\{ \RCompHidNodeBAF{i}{1} \}_{i=1}^{\LayerNumNodes{1}} $ through the BN layer.   We derived a proxy for the distribution (Eq.~\eqref{eq:b_int_expr}) for the processed variables. This distribution only assumes i.i.d. Gaussian variables for the batches in input in our layer. Remarkably the p.d.f. from Eq.~\eqref{eq:b_int_expr} does not depend on the mean and variance of the input random variables but only on the batch size $\BatchSize$. This means that, if we are in the regime where Eq.~\eqref{eq:b_int_expr} well approximates the distribution of the hidden layer node after the BN ($\BatchSize \gg 1$), $\RCompHidNodeABN{1}{i}$, then 
\begin{equation}
     \pdf{\RCompHidNodeABN{1}{i}}{(\Dataset)}{\ArgHidNodeABN{}} \approx \pdf{\RCusCompHidNodeABN{1}{i}}{(\Dataset)}{\ArgHidNodeABN{}} = \sqrt{\frac{1}{\BatchSize \pi}}  \frac{\Gamma \left( \frac{\BatchSize -1}{2}\right)}{\Gamma \left( \frac{\BatchSize -2}{2}\right)} \left( 1 + \frac{\ArgHidNodeABN{}^2}{\BatchSize}\right)^{-\frac{(\BatchSize -1)}{2}} \, .
\end{equation}

Therefore, after passing through the ReLU, using $\RCusCompHidNodeABN{1}{i}$ and the corresponding distribution as proxy for $\RCompHidNodeABN{l}{i}$, we will have
\begin{equation}\label{eq:g1_distr}
    \pdf{\RCompHidNodeAAF{i}{1}}{(\Dataset)}{\ArgHidNodeAAF} = \frac{1}{2} \Dirac{\ArgHidNodeAAF} + \Theta(\ArgHidNodeAAF) \sqrt{\frac{1}{\BatchSize \pi}}  \frac{\Gamma \left( \frac{\BatchSize -1}{2}\right)}{\Gamma \left( \frac{\BatchSize -2}{2}\right)} \left( 1 + \frac{\ArgHidNodeAAF^2}{\BatchSize}\right)^{-\frac{(\BatchSize -1)}{2}} \,.
\end{equation}

Note that $\{ \RCompHidNodeAAF{i}{1} \}_{i=1}^{\LayerNumNodes{1}}$ are identically distributed; from Eq.~\eqref{eq:g1_distr} we can compute the average over the dataset randomness that will be the same for each node; in particular:
\begin{align}
    \Einput{\RCompHidNodeAAF{i}{1}} =& \sqrt{\frac{1}{\BatchSize \pi}}  \frac{\Gamma \left( \frac{\BatchSize -1}{2}\right)}{\Gamma \left( \frac{\BatchSize -2}{2}\right)}  \int_{0}^{\infty} dz \, z \, \left( 1 + \frac{z^2}{\BatchSize}\right)^{-\frac{(\BatchSize -1)}{2}}  \notag \\
    =& \sqrt{\frac{1}{\BatchSize \pi}}  \frac{\Gamma \left( \frac{\BatchSize -1}{2}\right)}{\Gamma \left( \frac{\BatchSize -2}{2}\right)} \frac{\BatchSize}{2}  \int_{0}^{\infty} d \left( 1 + \frac{z^2}{\BatchSize}\right) \left( 1 + \frac{z^2}{\BatchSize}\right)^{-\frac{(\BatchSize -1)}{2}}  \notag \\
    =& \sqrt{\frac{1}{\BatchSize \pi}}  \frac{\Gamma \left( \frac{\BatchSize -1}{2}\right)}{\Gamma \left( \frac{\BatchSize -2}{2}\right)} \frac{\BatchSize}{2} \left.  \frac{2}{-\BatchSize +3} \left( 1 + \frac{z^2}{\BatchSize}\right)^{-\frac{(\BatchSize -3)}{2}}  \right|_{0}^{\infty} \notag \\
    =& \sqrt{\frac{1}{\BatchSize \pi}}  \frac{\Gamma \left( \frac{\BatchSize -1}{2}\right)}{\Gamma \left( \frac{\BatchSize -2}{2}\right)} \frac{\BatchSize}{\BatchSize -3} \, . 
\end{align}
Using Eq.~\eqref{eq:Var_b_i} and the fact that, by symmetry, 

\begin{align}
    \Einput{\left(\RCompHidNodeAAF{i}{1}\right)^2} = \frac{\CVar{}{\RCompHidNodeABN{1}{i}}}{2}  \approx \frac{ \CVar{}{\RCusCompHidNodeABN{1}{i}}}{2}\,,
\end{align}

we have 
\begin{align}\label{eq:VarY1}
    \CVar{\chi}{\RCompHidNodeAAF{i}{1}}  \approx \frac{\BatchSize}{2(\BatchSize -4)} - \left( \sqrt{\frac{1}{\BatchSize \pi}}  \frac{\Gamma \left( \frac{\BatchSize -1}{2}\right)}{\Gamma \left( \frac{\BatchSize -2}{2}\right)} \frac{\BatchSize}{\BatchSize -3} \right)^2\,.
\end{align}
Note that the r.h.s. of Eq.~\eqref{eq:VarY1} do not dependent from the node index $i$. We indicate with $\CVar{\chi}{\RCompHidNodeAAF{0}{1}}$ the value equal for each node, to highlight the absence of dependence from the index.
Repeating the analysis analogue to the one outlined in Ref.~\cite{pmlr-v235-francazi24a} (in particular see App.~D) we get, for a single hidden layer architecture:

\begin{align}
            \pdf{\ROutNode{c}{}}{(\Dataset)}{\OutNode{}{}} =&
    \NormalDens{\OutNode{}{}}{ \RMean{c}}{\StdKaiming^2 \CVar{\chi}{\RCompHidNodeAAF{0}{1}}}\,, \label{eq:out_1layer}\\
    \pdf{ \RMean{c}}{}{\ArgHidNodeAAF} =& \NormalDens{m}{0}{\StdKaiming^2 \Einput{\RCompHidNodeAAF{0}{1}{}}^2 } \,.\label{eq:out_1layer_mean}
\end{align}

\paragraph{Deep architectures}
We can then consider deeper architectures, increasing the number of hidden layers. Each hidden layer is defined by the same sequence of sub-modules: fully connection, BN, ReLU. Thus we can repeat, for each layer, the analysis of the single hidden layer case using as input the distribution of the output of the previous layer. For example for the second layer the we will repeat the same analysis using Eq.~\eqref{eq:out_1layer} as input distribution. Note that this is also a Gaussian distribution. We derived the p.d.f. of the set of random variables after passing through the BN transformation. In particular from Eq.~\eqref{eq:b_int_expr}
we can see that, assuming Gaussian i.i.d. random variables in input, the distribution of $\RCompHidNodeABN{l}{i}$ do not depend on the mean and variance of the input. Therefore, is easy to proof (\textit{e.g.} by induction) that the distribution after the batch norm will be the same for each layer.
This implies that the output nodes will follow Eq.~\eqref{eq:out_1layer} and Eq.~\eqref{eq:out_1layer_mean} irrespective to the depth of the network.

The presence of BN before the activation function keeps the level of IGB constant with the increasing of depth.

\newpage

\subsection{ReLU + BN}

We investigate the impact of applying Batch Normalization (BN) after the ReLU (ReLU + BN), in contrast to the more conventional placement before ReLU (BN + ReLU). We show that this modification significantly alters the statistical properties of the network's initial outputs, promoting more balanced and unbiased predictions at initialization. Following the analysis presented in Appendix~\ref{app:BN_B_A}, we first consider the full-batch regime and subsequently extend our results to the mini-batch setting.

\subsubsection{ReLU + BN: Full Batch} \label{relu + bn}
We start by analyzing the full-batch setting, deferring the generalization to the mini-batch case to App.~\ref{app:ReLU+BN:mb}. We show that applying BN after the activation function suppresses IGB, regardless of network depth. Specifically, we prove that the output nodes are identically distributed and centered at zero, resulting in class-balanced predictions at initialization.

The following is a formal formulation of Th. \ref{thm:bn-deep-after}.

\begin{theorembox}[ReLU + BN]
\label{thm:BN_A_A}
Consider the setting described in Sec.~\ref{sec:Settings} with an MLP where BN is applied \emph{after} the ReLU activation function, and assume full-batch normalization statistics. Then, in the limit of infinite width and dataset size, the output distribution \(\pdf{\ROutNode{c}{}}{(\Dataset)}{\OutNode{}{}}\) converges to:
\begin{align}
\pdf{\ROutNode{c}{}}{(\Dataset)}{\OutNode{}{}} \xrightarrow{| \WeightSet{} | \rightarrow \infty} \NormalDens{\OutNode{}{}}{0}{\StdKaiming^2},
\end{align}
with the corresponding mean distribution collapsing to:
\begin{align}
\pdf{\RMean{c}}{}{\Mean{}} \xrightarrow{| \WeightSet{} | \rightarrow \infty} \delta(\Mean{}),
\end{align}
indicating the suppression of IGB and the emergence of balanced initial predictions.
\end{theorembox}

To prove Thm.~\ref{thm:BN_A_A} we will follow once again the induction principle. In this case we want to show that the output nodes are identically distributed. In particular we will show that after each Dense layer, nodes are all centered in 0:
\begin{equation}
    \begin{cases}
      & \pdf{\RCompHidNodeBAF{i}{1}}{(\Dataset)}{\ArgHidNodeBAF}   = \NormalDens{\ArgHidNodeBAF}{0}{\StdKaiming^2}\,, \\ 
      &\pdf{\RCompHidNodeBAF{i}{l}}{(\Dataset)}{\ArgHidNodeBAF}   = \NormalDens{\ArgHidNodeBAF}{0}{\StdKaiming^2} \Longrightarrow \pdf{\RCompHidNodeBAF{i}{1+1}}{(\Dataset)}{\ArgHidNodeBAF}   = \NormalDens{\ArgHidNodeBAF}{0}{\StdKaiming^2}\,.
    \end{cases}
\end{equation}
\paragraph{Base step}
As stated above, our input data have distribution $\RCompInputValue{i}{}   \sim \NormalDens{\ArgHidNodeBAF}{0}{1} $ and they are IID. Also in this case it is straightforward to extend the analysis to arbitrary mean and variance for the input, reaching the same conclusions.\\
After passing through the weights, the data will undergo the following linear transformation:

\begin{equation}\label{h1 distribution}
     \RCompHidNodeBAF{i}{1} = \sum_{j=1}^{\DatasetSize} \Weights{0}{ij} \RCompInputValue{j}{}  , 
 \end{equation}
Now, the mean over the data of $\RCompHidNodeBAF{i}{1}$ is
\begin{equation}
    \Einput{\RCompHidNodeBAF{i}{1}} = \Einput{\sum_{j=1}^{\DatasetSize} \Weights{0}{ij} \RCompInputValue{j}{}  } \stackrel{\text{Ind.}}{=}  \sum_{j=1}^{\DatasetSize} \Weights{0}{ij}  \Einput{\RCompInputValue{j}{} } = 0,
\end{equation}
while the variance over the data  is
\begin{equation}
    \CVar{\Dataset}{\RCompHidNodeBAF{i}{1}} = \CVar{\Dataset}{ \sum_{j=1}^{\DatasetSize} \Weights{0}{ij} \RCompInputValue{j}{} } =  \sum_{j=1}^{\DatasetSize} \left(  \Weights{0}{ij} \right)^2 \CVar{\Dataset}{\RCompInputValue{j}{}} = \DatasetSize \frac{\StdKaiming^2 }{\DatasetSize} = \StdKaiming^2.
\end{equation}
After passing through the weights, the Dense output will distribute as
\begin{equation}
    \pdf{\RCompHidNodeBAF{i}{1}}{(\Dataset)}{\ArgHidNodeBAF}   = \NormalDens{\ArgHidNodeBAF}{0}{\StdKaiming^2}.
\end{equation}

\paragraph{Inductive step I: Dense $\rightarrow$ ReLU}
The effect of the ReLU is
\begin{equation}\label{relu_post_h}
    \pdf{\RCompHidNodeAAF{i}{l}}{(\Dataset)}{\ArgHidNodeAAF} = \frac{1}{2}\delta(\ArgHidNodeAAF) + \Theta(\ArgHidNodeAAF) \mathcal{N}(\ArgHidNodeAAF; 0,\,\StdKaiming^2).
\end{equation}
\\
\paragraph{Inductive step II. ReLU $\rightarrow$ BN}
We know that Batch Normalization has the following effect on the outputs:
\\
\begin{equation}
    \RCompHidNodeABN{l}{i}  = \frac{\RCompHidNodeAAF{i}{l} - \Einput{ \RCompHidNodeAAF{i}{l} } }{\CVar{\Dataset}{\RCompHidNodeAAF{i}{l}}} ,
\end{equation}
In order to evaluate the distribution of $\RCompHidNodeABN{l}{i}$, we need to compute the expected value and the variance of $\RCompHidNodeAAF{i}{l}$ as follows:

\begin{align}\label{mean-variance of ReLU-Normal}
    \Einput{\RCompHidNodeAAF{i}{l}}  &= \int_{\mathbb{R}}  \left( \frac{1}{2} \delta(\ArgHidNodeAAF) + \Theta(\ArgHidNodeAAF) \NormalDens{\ArgHidNodeAAF}{0}{\StdKaiming^2} \right)  \ArgHidNodeAAF  \,d\ArgHidNodeAAF = \notag \\ 
    &= \underbrace{\frac{1}{2}  \int_\mathbb{R} \delta(\ArgHidNodeAAF)\ArgHidNodeAAF\, d\ArgHidNodeAAF}_{=0}  + \frac{1}{\sqrt{2\pi\StdKaiming^2}}\int_0^\infty  \exp\left(-\frac{1}{2}\left(\frac{\ArgHidNodeAAF}{\StdKaiming}\right)^2\right)\ArgHidNodeAAF\, d\ArgHidNodeAAF = \frac{\StdKaiming}{\sqrt{2\pi}},
\end{align}

\begin{align}\label{second moment relu_bn}
    \Einput{\left(\RCompHidNodeAAF{i}{l}\right)^2}  &= \int_{\mathbb{R}}  \left( \frac{1}{2} \delta(\ArgHidNodeAAF) + \Theta(\ArgHidNodeAAF) \NormalDens{\ArgHidNodeAAF}{0}{\StdKaiming^2} \right)  \ArgHidNodeAAF^2  \,d\ArgHidNodeAAF = \notag \\ 
    &= \underbrace{\frac{1}{2}  \int_\mathbb{R} \delta(\ArgHidNodeAAF)\ArgHidNodeAAF^2 \, d\ArgHidNodeAAF}_{=0}  + \frac{1}{\sqrt{2\pi\StdKaiming^2}}\int_0^\infty  \exp\left(-\frac{1}{2}\left(\frac{\ArgHidNodeAAF}{\StdKaiming}\right)^2\right)\ArgHidNodeAAF^2 \, d\ArgHidNodeAAF = \frac{\StdKaiming^2}{2},
\end{align}

\begin{align}
    \CVar{\Dataset}{\RCompHidNodeAAF{i}{l}} = \Einput{\LR{\RCompHidNodeAAF{i}{l}}^2} - \Einput{\RCompHidNodeAAF{l}{i}}^2 = \frac{\StdKaiming^2}{2} - \frac{\StdKaiming^2}{2\pi} = \frac{\StdKaiming^2}{2} \left(1 - \frac{1}{\pi}\right).
\end{align}

So the expected value and standard deviation are $\Einput{\RCompHidNodeAAF{i}{l}}  = \frac{\StdKaiming }{\sqrt{2\pi}}$ and $\CVar{\Dataset}{\RCompHidNodeAAF{i}{l}} = \frac{\StdKaiming^2}{2 } - \frac{\StdKaiming^2}{2\pi}$, 

After the normalization, the output will have a distribution with zero mean and standard deviation equal to $1$.\\

For simplicity let's introduce the following notation: we define the parameters of the previous ReLU's output as $\Einput{\RCompHidNodeAAF{i}{l}}  = a$, $\sigma_{\RCompHidNodeAAF{i}{l}} = \alpha$.\\
\\
The distribution of the BN output is now the following:

\begin{align}
    \pdf{\RCompHidNodeABN{l}{i}}{(\Dataset)}{\ArgHidNodeABN} &= \frac{1}{2}\delta\left(\frac{\ArgHidNodeABN - a}{\alpha}\right)+ \Theta \left(\frac{\ArgHidNodeABN - a}{\alpha}\right) \mathcal{N}\left(\ArgHidNodeABN; a, \frac{\StdKaiming^2}{\alpha^2}\right) \notag \\
    &= \frac{\StdKaiming \sqrt{\pi - 1}}{2} \delta \left(\ArgHidNodeABN \sqrt{2\pi} - \StdKaiming \right) + \Theta \left(\ArgHidNodeABN\sqrt{2\pi} -\StdKaiming \right) \mathcal{N}\left(\ArgHidNodeABN; \frac{\StdKaiming}{\sqrt{2\pi}}, \frac{2\pi}{(\pi - 1)}\right).
\end{align}

From the above formula, one can easily verify that $\pdf{\RCompHidNodeABN{l}{i}}{(\Dataset)}{\ArgHidNodeABN}$ is a distribution centered at zero with unit variance.

\textbf{III. BN $\rightarrow$ Dense}\\
Note that the random variables $\{ \RCompHidNodeABN{l}{i} \}$'s are IID and have zero mean and unitary variance. \\
It means that, when we pass the output through the Dense Layer, the final output will have the following form:

\begin{equation}
    \RCompHidNodeAAF{i}{l+1} = \sum_{j = 1}^{\LayerNumNodes{l}} \Weights{l}{ij}\RCompHidNodeABN{l}{j},
\end{equation}
\\
therefore, from the CLT,
\begin{equation}
     \pdf{\RCompHidNodeBAF{i}{l+1}}{(\Dataset)}{\ArgHidNodeBAF}\xrightarrow{\text{$\LayerNumNodes{l} \rightarrow \infty$}} \mathcal{N}\left(\ArgHidNodeBAF; \Einput{ \sum_{j = 1}^{\LayerNumNodes{l}} \Weights{l}{ij}\RCompHidNodeABN{l}{j} }, \CVar{\Dataset}{ \sum_{j = 1}^{\LayerNumNodes{l}} \Weights{l}{ij}\RCompHidNodeABN{l}{j}} \right).
\end{equation}
\\
Given that what is stated in Eq. (\ref{mean_h}) and Eq. (\ref{var_h}) applies also to our case, then
\\
\begin{align}
    \Einput{\RCompHidNodeBAF{j}{l+1} } = \Einput{ \RCompHidNodeABN{l}{0} } \sum_{j = 1}^{N_1} \Weights{l}{ij} = 0, \\
    \CVar{\Dataset}{ \RCompHidNodeBAF{j}{l+1}} = \underbrace{\CE{\Dataset}{\left(\RCompHidNodeABN{l}{0} \right)^2}}_{=1} S_{w^2}^{(l)} = \StdKaiming^2.
\end{align}
\\

The reason why the expected value above is $1$ is that 

\begin{equation}
    \CE{\Dataset}{\left(\RCompHidNodeABN{l}{0} \right)^2} = \CVar{\Dataset}{ \RCompHidNodeABN{l}{0}} - \CE{\Dataset}{\left(\RCompHidNodeABN{l}{0} \right)}^2 = 1-0
\end{equation}

We can therefore write the distribution of the hidden units at any layer \(l+1\) as:
\begin{equation}
    \pdf{\RCompHidNodeBAF{i}{l+1}}{(\Dataset)}{\ArgHidNodeBAF} = \NormalDens{\ArgHidNodeBAF}{0}{\StdKaiming^2}.
\end{equation}
This conclusion extends to the output layer as well, since \(\ROutNode{c}{} = \RCompHidNodeBAF{c}{L+1}\). As a result, all output nodes are identically distributed and centered at zero. This symmetry implies that, at initialization, each class is predicted with equal probability, and therefore \emph{no initial guessing bias} is present. The model makes balanced predictions across the dataset, regardless of its depth.

\subsubsection{ReLU + BN: Mini Batch}\label{app:ReLU+BN:mb}
When BN is placed after the activation function we are in the following settings:
\begin{enumerate}[I.]
    \item Start from normally distributed datapoints.
    \item Pass inputs (divided in mini-batches) through a fully connected layer (Gaussian weights).
    \item Pass through the activation function (ReLU).
    \item Pass through the BN layer.
    \item (Repeat iteratively steps from II. to IV. in case of deep architectures).
    \item Repeat step II. to connect the last hidden layer to the output layer.
    \item Get the $\pdf{\RClassFraction{0}}{}{\ClassFraction{0}}$.
\end{enumerate}
In this case we don't need to explicitly compute the distribution. In fact at the end of each hidden layer--and in particular at the end of the last one-- we have the presence of BN that modify the distribution of each batch, centering the hidden representation in 0. The dataset is then given simply by the union of batches. Being the batches all centred in 0, the overall dataset average will be 0 as well. This means that, again according to the analysis of Ref.~\cite{pmlr-v235-francazi24a} (App.~D) we have
\begin{align}
    \pdf{\ROutNode{c}{}}{(\Dataset)}{\OutNode{}{}} =
    \NormalDens{\OutNode{}{}}{0}{\StdKaiming^2 \CVar{\Dataset}{\RCompHidNodeABN{L}{i}}}\,,
\end{align}
and therefore absence of IGB. 
\newpage

\section{Layer Normalization}
\label{app:LN}

LN is commonly applied in recurrent neural networks and Transformer models, typically positioned immediately before activation functions \cite{ba2016layer}. Unlike BN, LN computes statistics over individual samples, normalizing across all features within a single data point. Below, we analyze how the placement of LN within hidden layers impacts IGB in deep architectures.
\newline
Formally, LN computes normalization across the nodes within a layer, not across batches. Thus, given a layer with outputs from numerous nodes, LN standardizes these outputs individually for each data point. With the assumption that the number of nodes \( \LayerNumNodes{l} \) in each layer \( l \) is sufficiently large, \( \LayerNumNodes{l} \gg 1 \), we leverage the CLT in subsequent analyses.

In Fig. \ref{fig:static_Norm}, we illustrate the critical influence of LN placement. Specifically, applying \textbf{ReLU after LN} (LN + ReLU) maintains IGB, whereas applying \textbf{LN after ReLU} (ReLU + LN) eliminates IGB entirely.

To quantify fluctuations of layer outputs, we introduce \( \VarRatio \) (Definition~\ref{def:VarRatio}):
\begin{align}\label{eq:VarRatio_def}
   \VarRatio \equiv  \frac{\CVar{\WeightSet{}}{\RMean{c}}}{\CVar{\Dataset}{\ROutNode{c}{}}}.
\end{align}
An increased \( \VarRatio \) signifies greater IGB. Our analysis demonstrates that placing LN before ReLU proportionally scales both variances, preserving IGB, whereas positioning LN after ReLU sets \( \VarRatio \) consistently to zero.

To support subsequent analysis, we introduce the following lemma, elucidating the linearity properties of rectified normal distributions under scaling:

\begin{lemmabox}[Linearity of Mean and Variance of Rectified Normal Distributions]
\label{thm:rectified_normals}
Let \( X \sim \mathcal{N}(\mu, \sigma^2) \) and define \( Y = \alpha X \), with \(\alpha \in \mathbb{R}\). The ReLU activation function, defined as \(\text{ReLU}(z) = \max(0,z)\), satisfies:
\begin{align}
    \mathbb{E}[\text{ReLU}(Y)] &= \alpha \, \mathbb{E}[\text{ReLU}(X)], \\
    \text{Var}[\text{ReLU}(Y)] &= \alpha^2 \, \text{Var}[\text{ReLU}(X)].
\end{align}
\end{lemmabox}

\begin{proof}
Let \( X \sim \mathcal{N}(\mu, \sigma^2) \) and \( Y = \alpha X \sim \mathcal{N}(\alpha \mu, \alpha^2 \sigma^2) \). Consider the cumulative distribution function (CDF) \(\Phi\) of the standard normal distribution, defined by:
\[
\Phi(z) = \int_{-\infty}^{z} \frac{1}{\sqrt{2\pi}} e^{-t^2/2}\, dt.
\]

The distribution of the rectified variable \(\text{ReLU}(X)\) is given by:
\[
\text{ReLU}(X) \sim \Phi\left(-\frac{\mu}{\sigma}\right)\delta(x) + \Theta(x)\mathcal{N}(x; \mu, \sigma^2),
\]
where \(\delta(x)\) is the Dirac delta and \(\Theta(x)\) the Heaviside step function. Its expectation and variance are:
\begin{align}
\mathbb{E}[\text{ReLU}(X)] &= \mu\left[1 - \Phi\left(-\frac{\mu}{\sigma}\right)\right] + \frac{\sigma}{\sqrt{2\pi}}\exp\left(-\frac{\mu^2}{2\sigma^2}\right), \\
\text{Var}[\text{ReLU}(X)] &= (\mu^2 + \sigma^2)\left[1 - \Phi\left(-\frac{\mu}{\sigma}\right)\right] + \frac{\mu\sigma}{\sqrt{2\pi}}\exp\left(-\frac{\mu^2}{2\sigma^2}\right) - \mathbb{E}[\text{ReLU}(X)]^2.
\end{align}

Similarly, the rectified distribution \(\text{ReLU}(Y)\) is:
\[
\text{ReLU}(Y) \sim \Phi\left(-\frac{\alpha\mu}{|\alpha|\sigma}\right)\delta(y) + \Theta(y)\mathcal{N}(y; \alpha\mu, \alpha^2\sigma^2).
\]

Evaluating expectation and variance, we have:
\begin{align}
\mathbb{E}[\text{ReLU}(Y)] &= |\alpha|\left[ \frac{\alpha}{|\alpha|}\mu\left(1 - \Phi\left(-\frac{\alpha\mu}{|\alpha|\sigma}\right)\right) + \frac{|\alpha|\sigma}{\sqrt{2\pi}}\exp\left(-\frac{\alpha^2\mu^2}{2\alpha^2\sigma^2}\right) \right] \\
&= \alpha \left[ \mu\left(1 - \Phi\left(-\frac{\mu}{\sigma}\right)\right) + \frac{\sigma}{\sqrt{2\pi}}\exp\left(-\frac{\mu^2}{2\sigma^2}\right) \right] \\
&= \alpha \, \mathbb{E}[\text{ReLU}(X)], \\
\text{Var}[\text{ReLU}(Y)] &= \alpha^2\text{Var}[\text{ReLU}(X)],
\end{align}

thus confirming linearity and completing the proof.
\end{proof}

\subsection{LN + ReLU}\label{Sec:LN+ReLU}

We now analyze the impact on IGB of applying LN immediately before the ReLU activation function within settings depicted in Fig.~\ref{fig:NN_scheme}. A fundamental assumption throughout our analysis is that the width of each hidden layer is infinitely large, i.e., \(N_l \rightarrow \infty\) for every layer \(l\), with \(\LayerNumNodes{l}\) indicating the number of nodes in layer \(l\). Additionally, consistent with typical IGB analysis \citep{francazi2023theoretical}, we assume the initial input data distribution is standardized normal:
\begin{align}
     \RCompInputValue{i}{} &\sim \mathcal{N}(0,1).
\end{align}

The following is a formal formulation of Th. \ref{thm:ln-deep-before}.

\begin{theorembox}[LN + ReLU]
\label{thm:LN_before}
Consider the same setting as in Thm.~\ref{thm:out_dist}, but with LN applied before the activation function in each hidden layer. In the limit of infinite width and dataset size, the distribution of the output node \(\ROutNode{c}{}\) at initialization converges to:
\begin{align}
\pdf{\ROutNode{c}{}}{(\Dataset)}{ \OutNode{}{}} &\xrightarrow{| \WeightSet{} | \rightarrow \infty} \NormalDens{\OutNode{}{}}{\RMean{c}}{\LNVar{}{1} \CVar{\Dataset}{\ROutNode{c}{}}}, \\
\pdf{\RMean{c}}{}{\Mean{}} &\xrightarrow{| \WeightSet{} | \rightarrow \infty} \NormalDens{\Mean{}}{0}{\LNVar{}{1} \CVar{\WeightSet{}}{\RMean{c}}},
\end{align}
where \(\CVar{\Dataset}{\ROutNode{c}{}}\) and \(\CVar{\WeightSet{}}{\RMean{c}}\) denote variances of the output distribution and output means in the baseline scenario (ReLU-only, without normalization). Here, \(\LNVar{}{1}\) represents the variance of pre-activation values across the first layer prior to normalization.
\end{theorembox}

This theorem shows explicitly how Layer Normalization, when placed immediately before the ReLU activation, affects the distribution of layer outputs. Crucially, the distribution retains the same functional form as the baseline scenario described in Thm.~\ref{thm:out_dist}, with the only difference being a scaling of both variances by the same factor \(\LNVar{}{1}\). This scaling is constant and does not vary with network depth or across initialization instances. Consequently, when we compute \(\VarRatio\) (see Eq.~\eqref{eq:VarRatio_def}), which measures IGB as the ratio between variance of centers and variance of distributions around those centers, the scaling factor cancels out. Thus, the level of IGB remains exactly the same as in the unnormalized baseline scenario.

To formally demonstrate this equivalence, we now derive the distribution of the intermediate node \(\RCompHidNodeBAF{i}{l}\), representing the output of a deep neural network after \(l\) layers (Eq.~\ref{eq:DNN_prop}). By comparing the resultant distributions and variances with and without LN, we verify explicitly the invariance of the IGB measure \(\VarRatio\).

\textbf{Input $\rightarrow$ Dense}\\

After passing through the Dense Layer, the input is mapped following the same linear transformation as in Eq.~\eqref{h1 distribution} 
Now, averaging over the data distribution gives
\begin{equation}
    \Einput{\RCompHidNodeBAF{i}{1}} = \Einput{ \sum_{i = 1}^{\DatasetSize} \Weights{0}{ij} \RCompInputValue{j}{} } =  \sum_{i = 1}^{\DatasetSize}   \Weights{0}{ij}  \Einput{\RCompInputValue{j}{} } = 0,
\end{equation}
while the variance over the data  is
\begin{equation}
    \CVar{\Dataset}{\RCompHidNodeBAF{i}{1}} = \CVar{\Dataset}{ \sum_{i = 1}^{\DatasetSize} \Weights{0}{ij} \RCompInputValue{j}{}} =  \sum_{i = 1}^{\DatasetSize} \left(  \Weights{0}{ij} \right)^2\text{Var}\left(\RCompInputValue{j}{}\right) = \DatasetSize \frac{\StdKaiming^2 }{\DatasetSize} = \StdKaiming^2\,,
\end{equation}

Where we used the law of large numbers,
\begin{equation*}
    \frac1d \sum_{j=1}^d \left( w_{ij}^{(0)}\right)^2
    \stackrel{d\to\infty}{\longrightarrow}
    \CE{\mathcal{W}}{\left(w_{ij}^{(0)}\right)^2} = \frac{\sigma_w^2}d
    \,.
\end{equation*}

After passing through the weights, the Dense output will distribute as
\begin{equation}\label{eq:ph1}
    \pdf{\RCompHidNodeBAF{i}{1}}{(\Dataset)}{\ArgHidNodeBAF}   = \NormalDens{\ArgHidNodeBAF}{0}{\StdKaiming^2}.
\end{equation}
\textbf{Dense $\rightarrow$ LN}\\
The second step is to normalize the Dense output along the layer.\\
The transformation performed in LN standardizes $\RCompHidNodeBAF{i;a}{1}$ over all the elements of the respective layer as follows:
\begin{equation}
    \RCompHidNodeALN{1}{i;a} = \frac{\RCompHidNodeBAF{i;a}{1} - \LNMean{a}{1}}{\LNStd{a}{1}},
\end{equation}
where\\
\begin{itemize}
    \item $\RCompHidNodeALN{1}{i;a}$ represents the $i^{th}$ output (node) of the   input \(\RVecInputValue{a}{}\) after passing through  LN in the first block
    \item $\LNMean{a}{1} = \sum_i^{\LayerNumNodes{1}} \RCompHidNodeBAF{i;a}{1}/\LayerNumNodes{1}$ is the average along the layer.   $\LNMean{a}{1}  \xrightarrow[]{\LayerNumNodes{1} \rightarrow\infty}\CE{\WeightSet{}}{\RCompHidNodeBAF{i;a}{1}}  $.
    \item $\LNVar{a}{1} = \sum_j^{\LayerNumNodes{1}} \left(\RCompHidNodeBAF{j}{1} - \LNMean{a}{1}  \right)^2/\LayerNumNodes{1}$ is the variance along the layer.
\end{itemize}
Note the dependence on the sample index \( a \) in \( \LNMean{a}{1}\) and \( \LNStd{a}{1}\) as the two quantity are estimated over the sample node components. In the limit of $\LayerNumNodes{1} \rightarrow \infty$, however, these two random quantities concentrate around their mean value, according to the CLT, \textit{i.e.}:
\begin{align}
    \LNMean{a}{1}  &\rightarrow 0\,, \\
    \LNStd{a}{1}  &\rightarrow \LNStd{}{1} \,,
\end{align}
where $\LNVar{}{1} $ indicate the variance over the whole layer $l$, not anymore of a single node. 
We will assume this wide network regime in the following, avoiding the sample dependece on the estimators.
Then we can re-write the transformation as
\begin{equation}
    \RCompHidNodeALN{1}{i} = \frac{\RCompHidNodeBAF{i}{1}}{\LNStd{}{1}}.
\end{equation}

The distribution can now be written as 
\begin{equation}
   \pdf{\RCompHidNodeALN{1}{i}}{(\Dataset)}{\ArgHidNodeALN}  = \mathcal{N}\left(\ArgHidNodeALN; 0,  \frac{\sigma_{w}^2}{\LNVar{}{1}}  \right).
\end{equation}
Note that we are applying just a scaling, and not a shifting to the normal distribution. This means that the mean does not vary, but the variance gets scaled.\\

\textbf{LN $\rightarrow$ ReLU}\\
We then observe what happens after passing the LN output through the ReLU: we refer to the $i^{th}$ output of the ReLU activation of the $1^{st}$ layer as $\RCompHidNodeAAF{i}{1}$.\\
The output of the ReLU will then follow the distribution below:\\

\begin{equation}
    \pdf{\RCompHidNodeAAF{i}{1}}{(\Dataset)}{\ArgHidNodeAAF} = \frac{1}{2}\delta(\ArgHidNodeAAF) + \Theta(\ArgHidNodeAAF)\mathcal{N}\left(\ArgHidNodeAAF; \, 0,  \frac{\sigma_{w}^2}{\LNVar{}{1}}  \right).
\end{equation}

We will show in the following that this setting generates the same level of IGB as in the absence of LN, \textit{i.e.}, we get the same value for \(\VarRatio\) at each depth. Note that, in the absence of LN, we have an MLP with a ReLU activation function. As shown in \cite{pmlr-v235-francazi24a}, in this setting we have amplifying IGB, \textit{i.e.} IGB emerges and gets amplified with the depth.
To show the equivalence, in terms of IGB, between the two cases we will compare the two settings. We will call from now on with $\RNNCompHidNodeAAF{i}{1}  $ the node after passing through the ReLU in a MLP without LN. Applying the ReLU to the distribution in Eq.~\eqref{eq:ph1}  we will have 

\begin{equation}
 \pdf{\RNNCompHidNodeAAF{i}{1} }{(\Dataset)}{\ArgHidNodeAAF}  = \frac{1}{2}\delta(\ArgHidNodeAAF) + \Theta(\ArgHidNodeAAF)\mathcal{N}\left(\ArgHidNodeAAF; 0, \sigma_{w}^2 \right).
\end{equation}
As demonstrated in Theorem \ref{thm:rectified_normals}, the expected value and the variance of the two distributions are 
\begin{align}
\begin{split}
    \Einput{\RCompHidNodeAAF{i}{1}}  = \frac{ \Einput{ \RNNCompHidNodeAAF{i}{1} }}{\LNStd{}{1}},
    \end{split}
\end{align}
\begin{align}
\begin{split}
    \CVar{\Dataset}{ \RCompHidNodeAAF{i}{1}} = \frac{\CVar{\Dataset}{ \RNNCompHidNodeAAF{i}{1} }}{\LNVar{}{1}}.
\end{split}
 \end{align}

The latter will be useful in the next steps.\\

\textbf{ReLU $\rightarrow$ Dense}\\
The transformation after the Dense Layer is
\begin{equation}
    \RCompHidNodeBAF{i}{2} = \sum_{j = 1}^{\LayerNumNodes{1}} \Weights{1}{ij}\RCompHidNodeAAF{j}{1}.
\end{equation}
Then the distribution is

\begin{align}\label{h ditribution after 1 layer}
    \pdf{\RCompHidNodeBAF{i}{2} }{(\Dataset)}{\ArgHidNodeBAF} = \mathcal{N}\left(\ArgHidNodeBAF; \sum_j^{N_1} \Weights{1}{ij}  \Einput{\RCompHidNodeAAF{j}{1} },  \sum_j^{N_1} \left(\Weights{1}{ij}\right)^2 \CVar{\Dataset}{\RCompHidNodeAAF{j}{1}} \right) 
\end{align}

We just saw what happens to the distribution of the output after the first set of layers (Dense $\rightarrow$ LN $\rightarrow$ ReLU $\rightarrow$ Dense). \\
The next step is to study the behavior of the variance of the centers and the behavior of the variance around the centers: if the ratio of the latter increases with the number of layers, it means that there is strong IGB.\\
\\
First of all, we evaluate the probability distribution of the centers after the first layer.\\

\begin{align}
\begin{split}\label{eq:p<h2>}
\pdf{\Einput{ \RCompHidNodeBAF{i}{2}}}{}{\ArgHidNodeBAF}   = \NormalDens{\ArgHidNodeBAF}{ \sum_j^{\LayerNumNodes{1}} \CE{\WeightSet{}}{\Weights{1}{ij}}  \Einput{ \RCompHidNodeAAF{j}{1} }}{\sum_j^{\LayerNumNodes{1}}\text{Var}_{\mathcal{W}}\left( \Weights{1}{ij} \right) \Einput{ \RCompHidNodeAAF{j}{1} }^2} ,
   \end{split}  
\end{align}

Where $\CE{\WeightSet{}}{\cdot}$ is the expected value of the weights randomness.
\\
The expected value of the weights is zero, so the expected value of $\Einput{ \RCompHidNodeBAF{i}{2} }$ becomes zero, since the weights are centered in zero:
\\
\begin{equation}
    \CE{\WeightSet{}}{ \Einput{ \RCompHidNodeBAF{i}{2} }} = \sum_j^{\LayerNumNodes{1}} \CE{\WeightSet{}}{\Weights{1}{ij}}  \Einput{ \RCompHidNodeAAF{j}{1} } = 0,
\end{equation}
The variance is instead
\\
\begin{align}\label{eq:gam_num}
     \CVar{\WeightSet{}}{ \Einput{ \RCompHidNodeBAF{i}{2}} }&=\sum_j^{\LayerNumNodes{1}}\text{Var}_{\mathcal{W}}\left( \Weights{1}{ij} \right) \Einput{ \RCompHidNodeAAF{j}{1} }^2  =  \StdKaiming^2 \frac{1}{N_1} \sum_j^{\LayerNumNodes{1}} \Einput{ \RCompHidNodeAAF{j}{1} }^2 \notag \\
     &= \StdKaiming^2 \frac{1}{\LayerNumNodes{1}} \sum_j^{\LayerNumNodes{1}} \frac{\Einput{ \RNNCompHidNodeAAF{j}{1} }^2}{\LNVar{}{1}} \xrightarrow[]{\LayerNumNodes{1} \rightarrow\infty} \StdKaiming^2 \frac{\CE{\WeightSet{}}{\Einput{ \RNNCompHidNodeAAF{j}{1} }^2}}{\LNVar{}{1}} \, ,
\end{align}
where $\CE{\WeightSet{}}{\Einput{ \RNNCompHidNodeAAF{j}{1} }^2}$ is the average of $\Einput{ \RNNCompHidNodeAAF{j}{1}}^2$ over the weights randomness. 
\\
Substituting in Eq.~\eqref{eq:p<h2>} we have:
\begin{equation}\label{eq:f_EZ^2_LN}
    \pdf{\Einput{ \RCompHidNodeBAF{i}{2}}}{}{\ArgHidNodeBAF}   \xrightarrow[]{\LayerNumNodes{1} \rightarrow\infty} \mathcal{N}\left(\ArgHidNodeBAF; \, 0, \StdKaiming^2 \frac{\CE{\WeightSet{}}{\Einput{\RNNCompHidNodeAAF{j}{1}  }^2}}{\LNVar{}{1}} \right).
\end{equation}
We now pass to the denominator of the r.h.s. of Eq.~\eqref{eq:VarRatio_def}.
We saw in Eq.~\eqref{h ditribution after 1 layer} that the output of the dense layer $\RCompHidNodeBAF{i}{2}$ distributes as
\begin{equation}
    \pdf{\RCompHidNodeBAF{i}{2} }{(\Dataset)}{\ArgHidNodeBAF} = \NormalDens{\ArgHidNodeBAF}{\sum_j^{N_1} \Weights{1}{ij}  \Einput{\RCompHidNodeAAF{j}{1}{}}}{\sum_j^{N_1} \left(\Weights{1}{ij}\right)^2\CVar{\Dataset}{\RCompHidNodeAAF{j}{1}}}  .
\end{equation}
Now we need to study the expected value of the random variable 
$\text{Var}_\chi\left(\RCompHidNodeBAF{i}{2}\right)$ to compare it with the ones of the centers from Eq.~\eqref{eq:f_EZ^2_LN}:\\
\begin{align}
\begin{split}\label{eq:gam_den}
   \CE{\WeightSet{}}{ \CVar{\Dataset}{\RCompHidNodeBAF{i}{2}} } =& \CE{\WeightSet{}}{ \sum_j^{N_1}\left(\Weights{1}{ij}\right)^2\text{Var}_\chi\left(\RCompHidNodeAAF{j}{1}\right) } = \sum_j^{N_1}\CE{\WeightSet{}}{ \left(\Weights{1}{ij}\right)^2 } \CE{\WeightSet{}}{\CVar{\Dataset}{\RCompHidNodeAAF{j}{1}}}  \\
    =& \StdKaiming^2\frac{1}{N_1}\sum_j^{N_1} \CE{\WeightSet{}}{ \CVar{\Dataset}{  \RCompHidNodeAAF{j}{1} } }=\StdKaiming^2\frac{1}{N_1}\sum_j^{N_1}  \frac{\CE{\WeightSet{}}{\text{Var}_\chi  \left(\RNNCompHidNodeAAF{j}{1} \right)}}{\LNVar{}{l}}  \\ 
    =& \StdKaiming^2 \frac{\CE{\WeightSet{}}{\text{Var}_\chi  \left(\RNNCompHidNodeAAF{j}{1} \right)}}{\LNVar{}{l}} \, .
\end{split}
\end{align}\label{variance around the centers}

Finally, from Eq.~\eqref{eq:gam_num} and Eq.~\eqref{eq:gam_den} we get
\begin{equation}\label{eq:h2_var_ratio}
    \frac{\CVar{\WeightSet{}}{\Einput{\RCompHidNodeBAF{0}{2}}}}{\CVar{\Dataset}{\RCompHidNodeBAF{0}{2}}} = \frac{\StdKaiming^2 \frac{\CE{\WeightSet{}}{\Einput{ \RNNCompHidNodeAAF{j}{1} }^2}}{\LNVar{}{l}}}{\StdKaiming^2 \frac{\CE{\WeightSet{}}{\text{Var}_\chi  \left(\RNNCompHidNodeAAF{j}{1}  \right)}}{\LNVar{}{l}}} = \frac{\CE{\WeightSet{}}{\Einput{ \RNNCompHidNodeAAF{j}{1} }^2}} {\CE{\WeightSet{}}{\text{Var}_\chi  \left(\RNNCompHidNodeAAF{j}{1} \right)} } .
\end{equation}
Note that  $\RCompHidNodeBAF{c}{L} = \ROutNode{c}{}$ for a DNN with L hidden layers (see Eq.~\eqref{eq:DNN_prop}).
The ratio in the last term of Eq.~\eqref{eq:h2_var_ratio} is formally equal to the one we would get in a ReLU MLP, as shown in Ref.~\cite{pmlr-v235-francazi24a}.\\
We now proceed with the same computation for the following layers.\\

\textbf{Dense $\rightarrow$ LN}\\

In order to evaluate the expected value and the variance over the output layers, we need to study the probability distribution over the entire layer, not just over the single node's output.
To do so, we can compute the distribution of the layer preactivation as a convolution over the generic single node distribution. To distinguish the random variable associated with the specific node $i$, \textit{i.e.}, $\RCompHidNodeBAF{i}{2}$, from the generalized layer pre-activation (whose statistics are computed over different layer nodes), we indicate the latter with $\RCompHidNodeBAF{}{2}$. Note that there is a crucial difference between $\RCompHidNodeBAF{i}{2}$ and $\RCompHidNodeBAF{}{2}$ as, in the presence of IGB, nodes of the same layer are not identically distributed. In particular, since the difference between the nodes' distributions lies in the mean, we can derive the distribution of the layer by expressing the node distribution as a conditional probability (conditioned on the mean value) and marginalizing over the mean distribution, \textit{i.e.}

\begin{equation}\label{eq:dln}
    \pdf{\RCompHidNodeBAF{}{2}}{}{\ArgHidNodeBAF}= \int_{\mathbb{R}} \pdf{\RCompHidNodeBAF{i}{2}}{}{\ArgHidNodeBAF \mid \Einput{\RCompHidNodeBAF{i}{2}}=y} \pdf{\Einput{\RCompHidNodeBAF{i}{2}}}{}{y} \, dy\,.
\end{equation}

The left-hand side of Eq. \eqref{eq:dln} is the marginal distribution of the second-layer preactivations (not conditioned on a specific node). Conceptually, nodes in layer 2 are independent but not identically distributed: each node has a center
that depends on the fixed weights. The integral builds the layerwise distribution as a mixture over these node distributions, so we are effectively marginalizing over the weight initialization through the distribution of
(note that
is an expectation over dataset randomness but still a random variable with respect to the weights). 

In the following we refer to the variances of the two distributions respectively as $\CVar{\Dataset}{\RCompHidNodeBAF{i}{2}} = \sigma_{\ArgHidNodeBAF}^2$ and $\CVar{\WeightSet{}}{\Einput{\RCompHidNodeBAF{i}{2}}} = \sigma_{\Einput{\ArgHidNodeBAF}}^2$. We will have then:
\begin{align}
    \pdf{\RCompHidNodeBAF{}{2}}{}{\ArgHidNodeBAF} &= \int_{\mathbb{R}} \frac{1}{\sqrt{2\pi\sigma_{\ArgHidNodeBAF}^2}}\exp \left( -\frac{1}{2} \frac{(x - y)^2}{\sigma_{\ArgHidNodeBAF}^2}\right) \cdot  \frac{1}{\sqrt{2\pi\sigma_{\Einput{\ArgHidNodeBAF}}^2}}\exp \left( -\frac{1}{2} \frac{(y )^2}{\sigma_{\Einput{\ArgHidNodeBAF}}^2}\right)  dy \notag \\
    &=
    \int_{\mathbb{R}} \frac{1}{2\pi\sigma_{\ArgHidNodeBAF}\sigma_{\Einput{\ArgHidNodeBAF}}} \exp \left(-\frac{1}{2}\left( \frac{(x - y)^2}{\sigma_{\ArgHidNodeBAF}^2} + \frac{y^2}{\sigma_{\Einput{\ArgHidNodeBAF}}^2}\right) \right) \, dy \notag\\
    &= \frac{1}{\sqrt{2\pi(\sigma_{\ArgHidNodeBAF}^2 + \sigma_{\Einput{\ArgHidNodeBAF }}^2)}}\exp \left( -\frac{ x^2}{2 (\sigma_{\ArgHidNodeBAF}^2 + \sigma_{\Einput{\ArgHidNodeBAF}}^2)}\right)
      = \NormalDens{x}{0}{\sigma_{\ArgHidNodeBAF}^2 + \sigma_{\Einput{\ArgHidNodeBAF}}^2}
\end{align}

We now need to normalize our output over the Layer's expected value and standard deviation. \\
We already know the distribution of the layer's output, so in our case the normalization will have the form:
\begin{equation}\label{eq:DistL2}
    \RCompHidNodeALN{2}{i} = \frac{\RCompHidNodeBAF{i}{2} - 0}{\sqrt{\sigma_{\ArgHidNodeBAF}^2 + \sigma_{\Einput{ \ArgHidNodeBAF }}^2}} = \frac{\RCompHidNodeBAF{i}{2}}{\sqrt{\sigma_{\ArgHidNodeBAF}^2 + \sigma_{\Einput{ \ArgHidNodeBAF }}^2}} \sim  \mathcal{N}\left(\frac{\sum_j^{N_1} \Weights{1}{ij}  \Einput{ \RCompHidNodeAAF{j}{1} }}{\sqrt{\sigma_{\ArgHidNodeBAF}^2 + \sigma_{\Einput{ \ArgHidNodeBAF }}^2}} ,  \frac{\sum_j^{N_1} \left(\Weights{1}{ij}\right)^2\CVar{\Dataset}{\RCompHidNodeAAF{j}{1}}}{\sigma_{\ArgHidNodeBAF}^2 + \sigma_{\Einput{ \ArgHidNodeBAF }}^2}  \right)
\end{equation}

\textbf{LN $\rightarrow$ ReLU}\\
Now we will demonstrate that, even if the normally-distributed input is not centered in zero, the effect of LN is vain on IGB.\\
Also in this case, we will compare the parameters of the ReLU when LN is not applied  with our actual parameters.\\
\\
In order to simplify the notation, let us substitute the terms as follows:
\begin{itemize}
    \item $\sum_j^{N_1} \Weights{1}{ij}  \Einput{ \RCompHidNodeAAF{j}{1}} = \mu$ is the numerator of the mean of $\pdf{\RCompHidNodeALN{2}{i}}{(\Dataset)}{\ArgHidNodeALF}$ in Eq.~\eqref{eq:DistL2}
    \item $\sum_j^{N_1} \left(\Weights{1}{ij}\right)^2\CVar{\Dataset}{\RCompHidNodeAAF{j}{1}}= \sigma^2$ is the numerator of the variance of $\pdf{\RCompHidNodeALN{2}{i}}{(\Dataset)}{\ArgHidNodeALF}$ in Eq.~\eqref{eq:DistL2}
    \item $\sqrt{\sigma_{\ArgHidNodeBAF}^2 + \sigma_{\langle \ArgHidNodeBAF \rangle}^2} = \alpha $ is the denominator of the mean (and the standard deviation's denominator) of $\pdf{\RCompHidNodeALN{2}{i}}{(\Dataset)}{\ArgHidNodeALF}$ in Eq.~\eqref{eq:DistL2}
\end{itemize}
We can now re-write the distribution of $\RCompHidNodeALN{2}{i} | \mathcal{W}$ as follows:
\begin{equation}
    \pdf{\RCompHidNodeALN{2}{i}}{(\Dataset)}{\ArgHidNodeALF} = \mathcal{N}\left(\ArgHidNodeALF; \, \frac{\mu}{\alpha}, \, \frac{\sigma^2}{\alpha^2} \right).
\end{equation}
while - if the LN is not applied - the distribution would remain
\begin{equation}
   \pdf{\RCompHidNodeBAF{i}{2}}{(\Dataset)}{\ArgHidNodeBAF}  = \mathcal{N}\left(\ArgHidNodeBAF; \, \mu, \, \sigma^2 \right).
\end{equation}
The output of the rectified $\RNNCompHidNodeAAF{i}{2}$ would be
\begin{equation}
  \pdf{\RNNCompHidNodeAAF{i}{2} }{(\Dataset)}{\ArgHidNodeAAF} = \Phi \left(\frac{\mu}{\sigma} \right)\delta (\ArgHidNodeAAF) + \Theta (\ArgHidNodeAAF) \mathcal{N}\left(\ArgHidNodeAAF; \, \mu, \, \sigma^2 \right),
\end{equation}
while the distribution of the rectified $\RCompHidNodeALN{2}{i}$ will be
\begin{equation}
   \pdf{\RCompHidNodeAAF{i}{2}}{(\Dataset)}{\ArgHidNodeAAF} = \Phi \left(\frac{\mu/\alpha}{\sigma/\alpha} \right)\delta (\ArgHidNodeAAF) + \Theta (\ArgHidNodeAAF) \mathcal{N}\left(\ArgHidNodeAAF; \, \frac{\mu}{\alpha}, \, \frac{\sigma^2}{\alpha^2} \right).
\end{equation}
From Theorem \ref{thm:rectified_normals} about linearity of expected value and variance of rectified normal distributions, we can see that - if LN is applied - the mean and the variance of $\RCompHidNodeAAF{i}{2}$ will be proportional to the ones of $\RNNCompHidNodeAAF{i}{2} $:
\begin{align}
\begin{split}
    \Einput{ \RCompHidNodeAAF{i}{2} } =  \frac{1}{\sqrt{\sigma_{\ArgHidNodeBAF}^2 + \sigma_{\langle \ArgHidNodeBAF \rangle}^2}}\Einput{ \RNNCompHidNodeAAF{i}{2} },
\end{split}
\end{align}
\begin{align}
    \CVar{\Dataset}{\RCompHidNodeAAF{i}{2}} =  \frac{1}{\sigma_{\ArgHidNodeBAF}^2 + \sigma_{\langle \ArgHidNodeBAF \rangle}^2} \CVar{\Dataset}{\RNNCompHidNodeAAF{i}{2}}.
\end{align}
This means that the procedure to find the ratio of the variance of the centers over the variance around the centers is exactly the same as the one starting from Eq. \ref{h ditribution after 1 layer}.\\
After the next Dense layer, the ratio of the two variances of interest will be
\begin{equation}
    \frac{\CVar{\WeightSet{}}{\Einput{\RCompHidNodeBAF{0}{3}}}}{\CVar{\Dataset}{\RCompHidNodeBAF{0}{3}}}= \frac{\StdKaiming^2 \frac{\CE{\WeightSet{}}{\Einput{ \RNNCompHidNodeAAF{0}{2} }^2}}{(\sigma_{\ArgHidNodeBAF}^2 + \sigma_{\langle \ArgHidNodeBAF \rangle}^2)}}{\StdKaiming^2 \frac{\CE{\WeightSet{}}{\text{Var}_\chi  \left(\RNNCompHidNodeAAF{0}{2} \right)}}{(\sigma_{\ArgHidNodeBAF}^2 + \sigma_{\langle \ArgHidNodeBAF \rangle}^2)}} = \frac{\CE{\WeightSet{}}{\Einput{ \RNNCompHidNodeAAF{0}{2}  }^2} }{\CE{\WeightSet{}}{\text{Var}_\chi  \left(\RNNCompHidNodeAAF{0}{2} \right)} }, 
\end{equation}
which again is formally equal to the expression of $\VarRatio$ in absence of LN, computed in \cite{pmlr-v235-francazi24a}. We can proceed in the same way for the generic layer $l$, arriving to the same conclusion.\\
We have now demonstrated that, the presence of LN before ReLU leave the level of IGB unchanged.
Therefore, as in the ReLU MLP settings, we will have a level of IGB increasing with the depth of the DNN.\\
The same procedure can be iterated for all the following layers.

\begin{figure}[]
\centering
\includegraphics[width=0.7\textwidth]{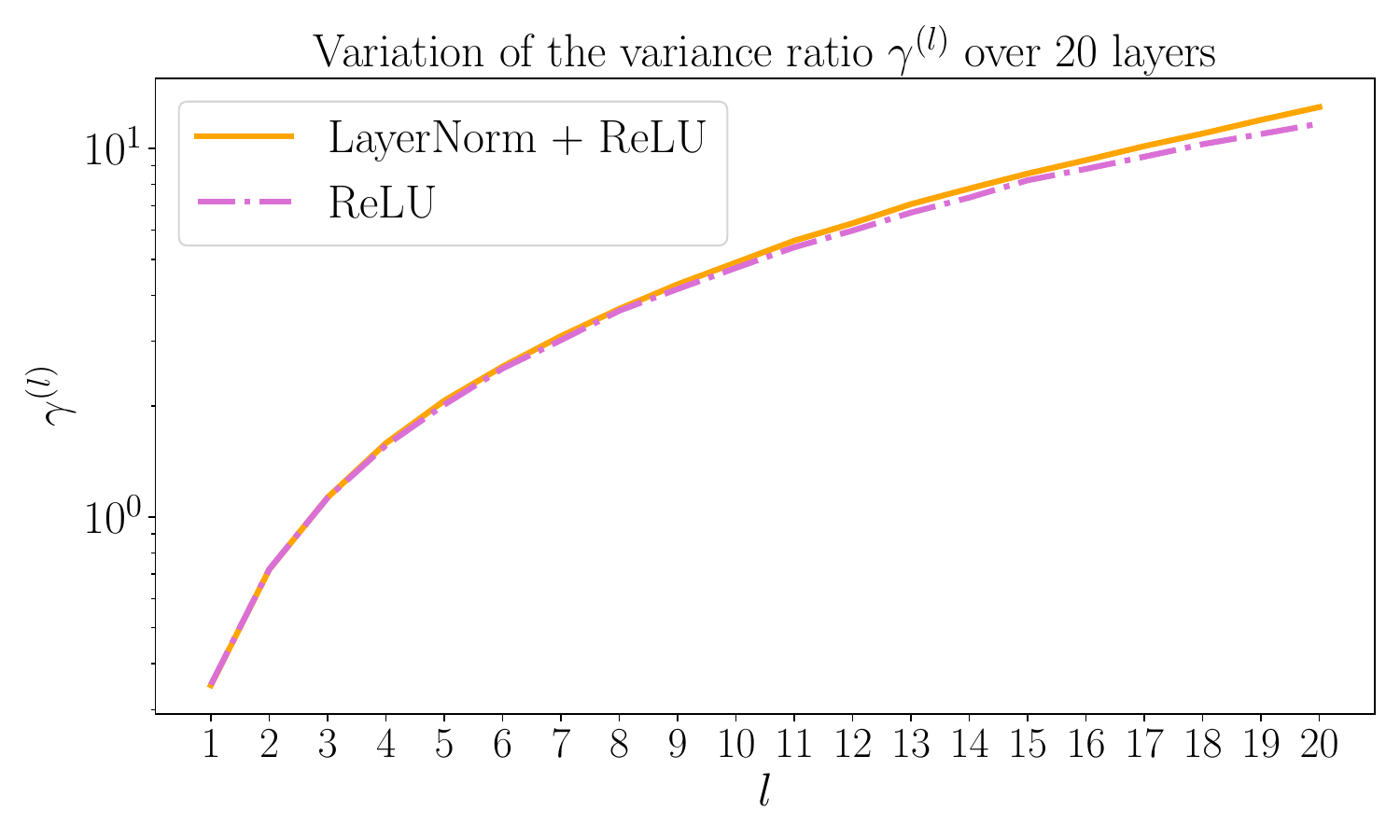}
\caption{Empirical comparison of the variance ratio $\frac{\CVar{\WeightSet{}}{\Einput{\RCompHidNodeBAF{0}{l}}}}{\CVar{\Dataset}{\RCompHidNodeBAF{0}{l}}}$ between the ReLU case and the LN+ReLU case for 20 consecutive layers. The two curves are overlapped, meaning that the effect of LN before ReLU is vain on IGB. }
\label{fig:ln-relu infinite layer}
\end{figure}

\newpage

\subsection{ReLU + LN}
We now study what happens when LN is applied after ReLU, with a neural  network within the setting illustrated in Fig.~\ref{fig:NN_scheme}.
We again assume that our input data distributes as $\NormalDistr{0}{1}$, as the typical setting employed in IGB analysis \citep{francazi2023theoretical}.\\ 
The input data will then pass through the first dense layer: the weights are initialized with \textbf{Kaiming Initialization}.
An important assumption that will be valid for all the passages, is that $N_l \rightarrow \infty$ for every $l$.

The following is a formal formulation of Th. \ref{thm:ln-deep-after}.

\begin{theorembox}[ReLU + LN]
\label{thm:LN_A_A}
Consider the setting described in Sec.~\ref{sec:Settings} with an MLP where LN is applied \emph{after} the ReLU activation function in each hidden layer. Then, in the limit of infinite width and dataset size, the output distribution \(\pdf{\ROutNode{c}{}}{(\Dataset)}{\OutNode{}{}}\) converges to:
\begin{align}
\pdf{\ROutNode{c}{}}{(\Dataset)}{\OutNode{}{}} \xrightarrow{| \WeightSet{} | \rightarrow \infty} \NormalDens{\OutNode{}{}}{0}{\StdKaiming^2},
\end{align}
with the corresponding mean distribution collapsing to:
\begin{align}
\pdf{\RMean{c}}{}{\Mean{}} \xrightarrow{| \WeightSet{} | \rightarrow \infty} \delta(\Mean{}),
\end{align}
indicating the suppression of Initial Guessing Bias and the emergence of fully balanced predictions at initialization.
\end{theorembox}

\textbf{Input $\rightarrow$ Dense}\\
The first step is to observe the distribution of the output after passing through the Dense Layer.\\
The transformation performed is the following:
\begin{equation}
   \RCompHidNodeBAF{i}{1} = \sum_{j = 1}^{\LayerNumNodes{0}} w_{ij}^{(0)} \RCompInputValue{j}{}.
\end{equation}
where \(\RCompHidNodeBAF{i}{1}\) represents the $i^{th}$ output (node) of the Dense Layer in the first block.\\
Now, the mean over the data of $\RCompHidNodeBAF{i}{1}$ is
\begin{equation}
    \Einput{ \RCompHidNodeBAF{i}{1}} = \Einput{ \sum_{i = 1}^{\LayerNumNodes{0}} w_{ij}^{(0)} \RCompInputValue{j}{} } =  \sum_{i = 1}^{\LayerNumNodes{0}}  w_{ij}^{(0) } \Einput{ \RCompInputValue{j}{} } = 0,
\end{equation}
While the variance over the data  is
\begin{equation}
    \text{Var}\left(\RCompHidNodeBAF{i}{1} \right) = \text{Var}\left( \sum_{i = 1}^{N_1} w_{ij}^{(0)} \RCompInputValue{j}{}  \right) \stackrel{\text{Ind.}}{=}  \sum_{i = 1}^{N_1} \left(  w_{ij}^{(0)} - 0 \right)^2\text{Var}\left(\RCompInputValue{j}{}\right) = N_1 \frac{\sigma_w^2 }{N_1} = \sigma_w^2.
\end{equation}
After passing through the weights, the Dense output of every node will distribute, independently, as
\begin{equation}\label{h_1_relu_ln}
   \pdf{\RCompHidNodeBAF{i}{1}}{(\Dataset)}{\ArgHidNodeBAF} = \mathcal{N}\left(\ArgHidNodeBAF; \,0, \sigma_{w_i}^2\right).
\end{equation}
\textbf{Dense $\rightarrow$ ReLU}\\
After passing every node's output through the ReLU activation function, we will have the same distribution for every node, defined as 
\begin{equation}
     \pdf{\RCompHidNodeAAF{i}{1}}{(\Dataset)}{\ArgHidNodeAAF}  = \frac{1}{2}\delta(\ArgHidNodeAAF) + \Theta(\ArgHidNodeAAF) \mathcal{N}\left(\ArgHidNodeAAF; 0,\,\sigma_w^2\right),
\end{equation}

Given that all the $\RCompHidNodeAAF{i}{1}$ are identically distributed, it means that 
\begin{align}
    \Einput{ \RCompHidNodeAAF{i}{1} }  &= \Einput{ \RCompHidNodeAAF{j}{1} } \,\forall \, i,j \in \mathbb{N},\\
    \CVar{\Dataset}{\RCompHidNodeAAF{i}{1}}  &= \CVar{\Dataset}{\RCompHidNodeAAF{j}{1}}\,\forall \, i,j \in \mathbb{N}.
\end{align}
\textbf{ReLU $\rightarrow$ LN}\\
We now normalize the ReLU output over the layer. The elements needed for LN have already been defined as: 
\begin{itemize}
    \item $\RCompHidNodeALN{1}{i;a}$ represents the $i^{th}$ output (node) of the  LN in the first block
    \item $ \LNMean{a}{1} = \sum_i^{N_1} \RCompHidNodeAAF{i}{1}/N_1$ is the average along the layer.  $\LNMean{a}{1}  \xrightarrow[]{N_1 \rightarrow\infty} \CE{\WeightSet{}}{\RCompHidNodeAAF{i}{1}} $
    \item $\LNVar{a}{1} = \sum_i^{N_1} \left(\RCompHidNodeAAF{i}{1} - \LNMean{a}{1} \right)/N_1$ is the variance along the layer.
\end{itemize}
As in Sec.~\ref{Sec:LN+ReLU} we work in the limit of $\LayerNumNodes{l} \to \infty \,\,\, \forall l$ where the two estimators $\LNMean{a}{1}$ and $\LNVar{a}{1}$ converge, according to the CLT, to their expectation value. Note that in the first layer the nodes $\left\{ \RCompHidNodeAAF{i}{1} \right\}$ are independent and identically distributed \citep{pmlr-v235-francazi24a}.
Therefore, we can see the LN transformation as equivalent to a BN transformation: 
\begin{equation}
    \RCompHidNodeALN{1}{i;a} = \frac{\RCompHidNodeAAF{i}{1} - \CE{\WeightSet{}}{\RCompHidNodeAAF{i}{1}}}{\sqrt{\CVar{\WeightSet{}}{\RCompHidNodeAAF{i}{1}}}} = \frac{\RCompHidNodeAAF{i}{1} -  \Einput{ \RCompHidNodeAAF{i}{1} }}{\sqrt{\CVar{\Dataset}{ \RCompHidNodeAAF{i}{1} }}},
\end{equation}
This means that the distribution of every node of $l^{(1)}$ has zero mean and unitary variance; the expected value and the variance over the dataset are standardized:
\begin{align}
    \Einput{ \RCompHidNodeALN{1}{i} } &= 0, \, \forall i \\ 
    \CVar{\Dataset}{\RCompHidNodeALN{1}{i}} &= 1, \, \forall i.
\end{align}
\textbf{LN $\rightarrow$ Dense}\\
The transformation performed by the fully connected layer is the following:
\begin{equation}
    \RCompHidNodeBAF{i}{2} = \sum_{j = 1}^{N_1} w_{ij}^{(1)} \RCompHidNodeALN{1}{j}.
\end{equation}
So the distribution of $ \RCompHidNodeBAF{i}{2}$ will be
\begin{equation}
    \pdf{\RCompHidNodeBAF{i}{1}}{(\Dataset)}{\ArgHidNodeBAF} =  \mathcal{N}\left(\ArgHidNodeBAF; \, \sum_j^{N_1} w_j^{(1)}  \Einput{\RCompHidNodeALN{1}{j} },  \sum_j^{N_1} \left(w_{j}^{(1)}\right)^2\CVar{\Dataset}{\RCompHidNodeALN{1}{j}} \right) 
\end{equation}
Given that we demonstrated that $\Einput{ \RCompHidNodeALN{1}{i} } = 0, \, \forall i$ and $\CVar{\Dataset}{\RCompHidNodeALN{1}{i} } = 1, \, \forall i$ the distribution will be 
\begin{equation}
    \pdf{\RCompHidNodeBAF{i}{1}}{(\Dataset)}{\ArgHidNodeBAF} =  \mathcal{N}\left(\ArgHidNodeBAF; \, 0,  \sum_j^{N_1} \left(w_{j}^{(1)}\right)^2 \right)  = \mathcal{N}\left(\ArgHidNodeBAF; \, 0,  \sum_j^{N_1} \left(w_{j}^{(1)} - 0\right)^2 \right)  \xrightarrow[]{N_1 \rightarrow\infty} \mathcal{N}\left(\ArgHidNodeBAF; \, 0,  \sigma_w^2 \right)
\end{equation}
We are, again, in the same situation described for $h_i^{(1)}$ in Eq. \ref{h_1_relu_ln}.\\ This means that the null amount of IGB is the same as $\RCompHidNodeBAF{i}{1}$ and it will not increase in the next blocks' output.

\newpage

\section{Experiments}\label{app:Exp}
In this section, we present complementary experiments supporting our analysis; implementation details, including model specifications and datasets used, are provided in App.~\ref{sec:reprod}.

\subsection{Reproducibility}\label{sec:reprod}
Here we provide technical details about the experiments, to allow for reproducibility. The code used for the experiments presented in this work is available at \url{https://github.com/EmanueleFrancazi/IGB-and-Normalization}

\paragraph{Datasets}
\begin{itemize}
\item \textbf{Gaussian Blob} (\GB): Consistent with the presented analysis, for most of the experiments shown we used a Gaussian blob as input. Specifically, all elements of the dataset are \textit{i.i.d.}, and each individual element consists of a random vector of dimension $d = 1000$ with normally distributed components. For binary experiments, we generate two Gaussian blobs centered at $\pm \mu$, where the mean $\mu$ scales with the input dimension to ensure a non-vanishing class overlap even in the high-dimensional limit. In particular, we set $\mu = 1/\sqrt{d}$, such that:
\begin{equation}\notag
    \RCompInputValue{i}{} \sim 
    \begin{cases}
        \NormalDistr{-\mu}{\mathbf{I}} & \text{if } y_i = 0\,, \\
        \NormalDistr{+\mu}{\mathbf{I}} & \text{if } y_i = 1\,.
    \end{cases}
\end{equation}
This setup provides a controlled and analytically tractable scenario for evaluating binary classification in high dimensions.

   \item \textbf{CIFAR10} (\CIFAR): We use \texttt{CIFAR10} (\url{https://www.cs.toronto.edu/~kriz/cifar.html})~\citep{cifar} as an example of a real multi-class dataset. Before the start of the simulation, we perform standardization: pixel values are rescaled to the interval $[0, 1]$, followed by channel-wise mean subtraction and standard deviation normalization. 

While we present multi-class experiments on more advanced architectures, for MLP-based experiments we focus on binary classification tasks for greater interpretability. To this end, we define two different binary versions of CIFAR10:

\begin{itemize}
    \item \textbf{Cat vs Dog} (\CatVsDog): A binary classification task where we extract only the \texttt{cat} and \texttt{dog} classes from the full dataset.
    \item \textbf{Vehicles vs Animals} (\VehVsAni): A broader binary task where we aggregate multiple classes into two macro categories: vehicles (\texttt{airplane}, \texttt{automobile}, \texttt{truck}) and animals (\texttt{cat}, \texttt{deer}, \texttt{dog}, \texttt{horse}).
\end{itemize}

These setups allow us to assess the behavior of the models under both narrowly and broadly defined binary distinctions.

\item \textbf{Tiny ImageNet} (\TinyImageNet): A subset of the ImageNet dataset, consisting of $200$ classes with $500$ training images, $50$ validation images, and $50$ test images per class, each downsampled to $64 \times 64$ pixels. The dataset is publicly available at \url{https://www.kaggle.com/c/tiny-imagenet}.

    \item \textbf{MNIST} (\MNIST): We use \texttt{MNIST} (\url{http://yann.lecun.com/exdb/mnist/})~\citep{deng2012mnist} to reproduce binary experiments on real data. The binary dataset is defined by merging the digit classes into two macro groups based on parity: the \texttt{even} class $\{0,2,4,6,8\}$ and the \texttt{odd} class $\{1,3,5,7,9\}$.
\end{itemize}

\paragraph{Models}
We provide details on the architectures used in our experiments. Our description does not include loss functions, as they are irrelevant in the untrained networks analyzed. 
\begin{itemize}
    \item \textbf{MLP}: Our analysis provides theoretical predictions for MLPs. To support our study, we consider two different MLP configurations:
    \begin{itemize}
        \item \textbf{MLP with a single hidden layer} (\MLPA):  The hidden layer contains $\LayerNumNodes{1}=100$ nodes.
        \item \textbf{MLP with multiple hidden layers} (\MLPB): This network has $L=20$ hidden layers, each with $N=100$ nodes.
    \end{itemize}
    Each MLP model employs ReLU activation, which is a widespread example of an activation function known to induce IGB and to amplify it with network depth~\citep{pmlr-v235-francazi24a}. This setting serves as our baseline, previously analyzed in~\citep{pmlr-v235-francazi24a}. In our experiments, we compare this baseline to variants that incorporate normalization either before or after the activation, as discussed in Appendix~\ref{sec:Notation}.

    \item \textbf{MLP-mixer}: We include the MLP-mixer (\MLPmix), introduced in~\cite{tolstikhin2021mlp}, as an example of a more advanced MLP architecture. We adopt the design described at \url{https://github.com/omihub777/MLP-Mixer-CIFAR/blob/main/README.md}.

    \item \textbf{ResNet34}: We consider ResNet34 (\ResNet), proposed in~\citet{he2016deep}, as an example of an architecture with skip connections. We use the implementation available at \url{https://www.kaggle.com/code/kmldas/cifar10-resnet-90-accuracy-less-than-5-min}.
    
    \item \textbf{ResNet101} (\ResNetHundredOne): As a deeper residual network, we also consider ResNet101, following the implementation available at \url{https://github.com/mbk2103/ResNet101-Implementation/blob/main/resnet_model.py}.

    \item \textbf{Swin Transformer (Large)} (\SwinL): Finally, we include the Swin Transformer (Large) model~\citep{liu2021swin}, representative of modern transformer-based architectures. We rely on the implementation available at \url{https://github.com/WangFeng18/Swin-Transformer/blob/master/SwinTransformer.py}.

\end{itemize}

We emphasize that these architectures were not chosen to represent the state of the art, but rather as well-documented, accessible examples capable of achieving good performance with relatively low computational cost. This makes our findings regarding IGB more broadly applicable and facilitates reproducibility.

\subsection{Experiments on MLPs}\label{sec:exp_mlp}
We begin by analyzing MLP architectures, where theoretical predictions regarding IGB are most directly applicable. In all experiments, random initializations are filtered based on the initial class bias $\RClassFraction{0}$, allowing us to compare \textit{neutral} initializations ($\RClassFraction{0} \approx 0.5$) to \textit{prejudiced} ones ($\RClassFraction{0} \approx 1$). This filtering enables us to isolate how class bias at initialization affects training.
\newline
A central aim of our experimental analysis is to examine how well the qualitative behavior predicted by our theory—formulated under unstructured data assumptions—translates to real-world scenarios. While our theoretical framework is developed for random, high-dimensional inputs, we empirically assess whether similar dynamics are observed in practical settings, and identify where significant deviations emerge. This comparison offers insight into which behaviors are driven primarily by architectural design and which arise from interactions with data structure. Although a full analysis of these effects is beyond the scope of this work, our findings highlight promising directions for future study into the complex interplay between architecture-induced and data-induced biases.
To this end, we extend our analysis beyond the Gaussian blob setting and consider \VehVsAni{} and \MNIST{} as two representative examples of structured real datasets. The comparison allows us to underline both the consistency and the divergence between theoretical predictions and empirical behavior in more realistic conditions.
All dynamic experiments are repeated over multiple random seeds; plots report the mean and standard error across runs to account for variability.

\begin{figure*}[h]
    \centering
    \includegraphics[width=0.45\textwidth]{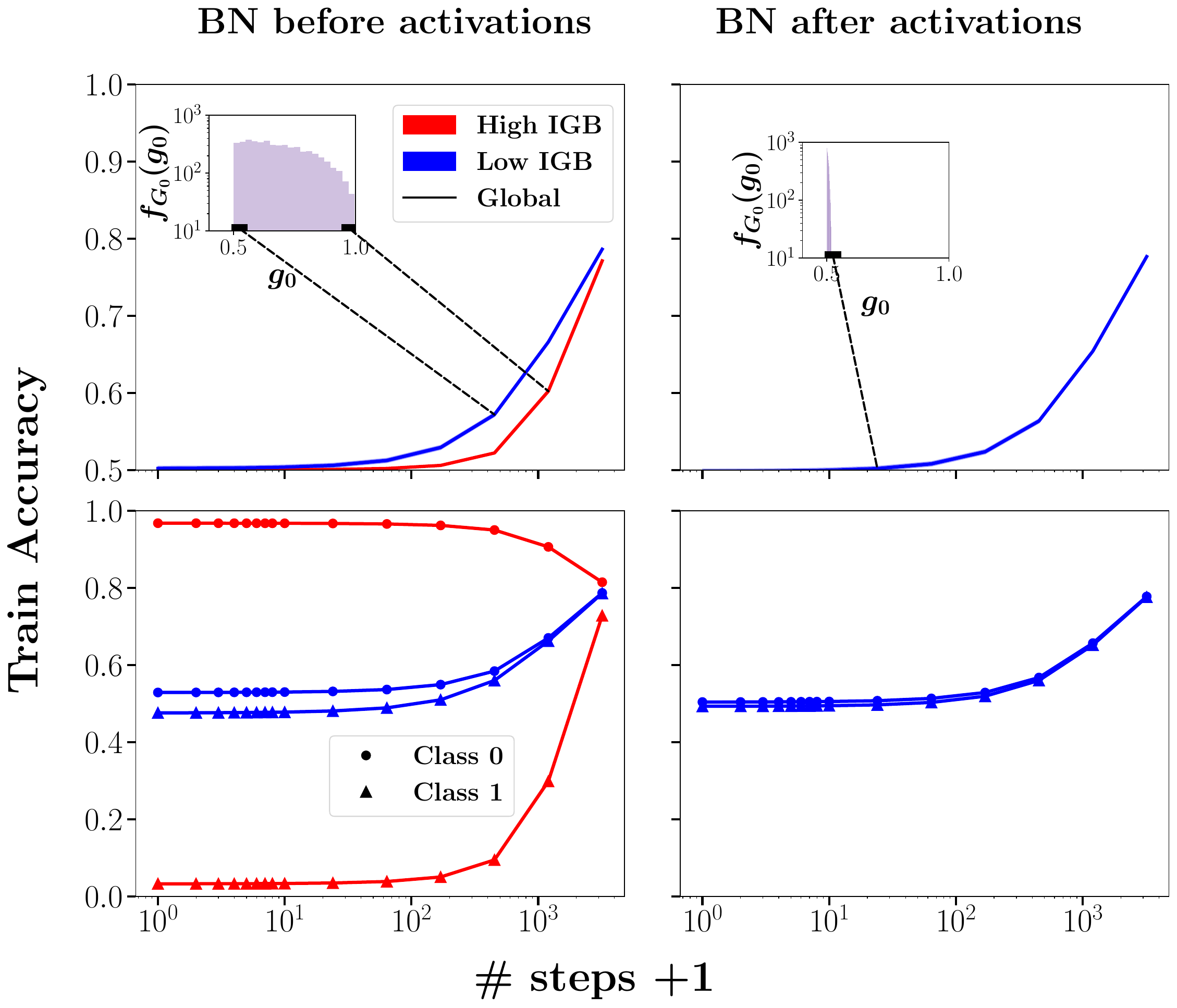}
    \includegraphics[width=0.45\textwidth]{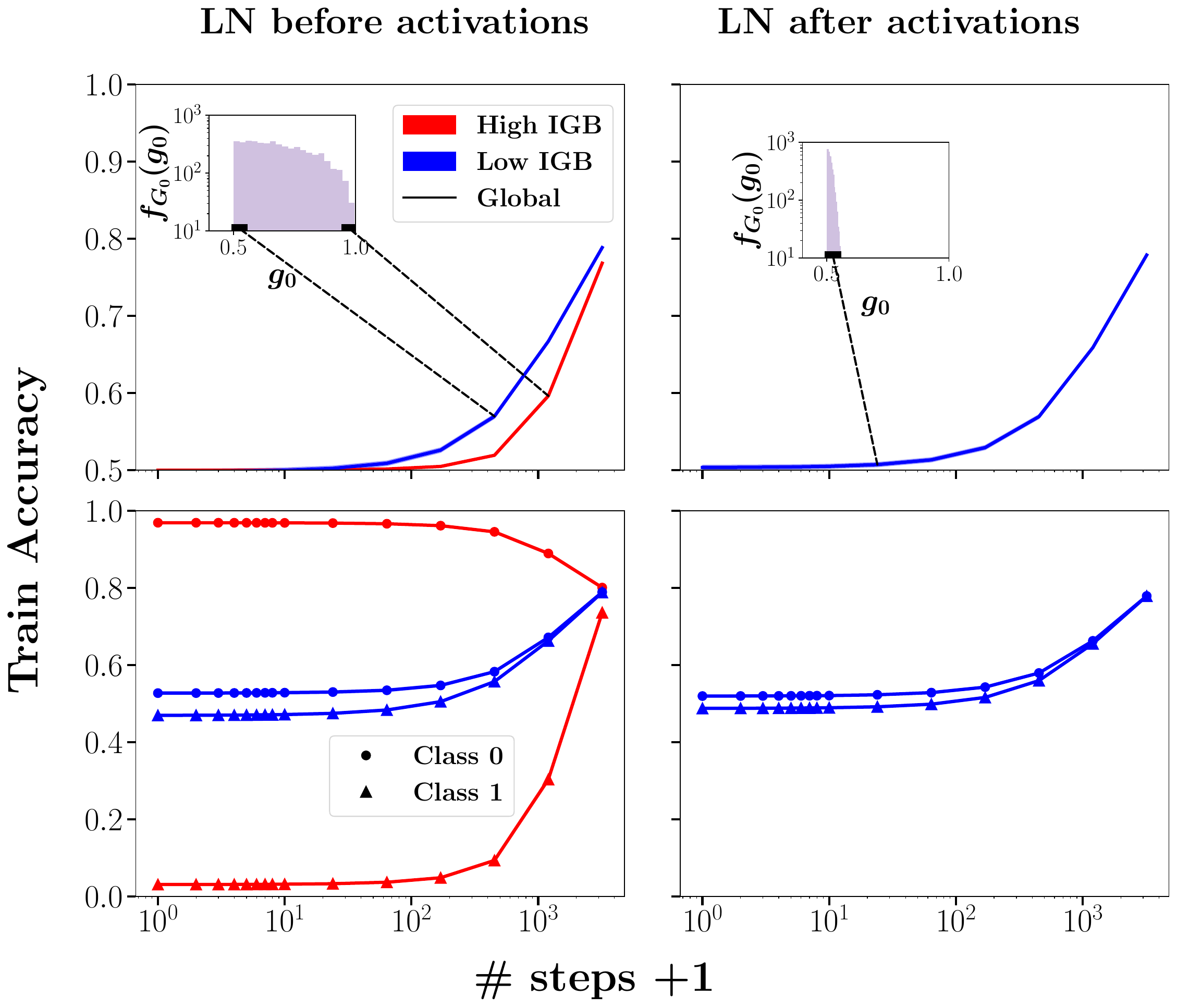}
    \caption{
    Training accuracy on \GB{} using the shallow \MLPA{} model. Left: BN; Right: LN. Each panel compares normalization applied \textit{before} vs \textit{after} ReLU. Within each configuration, curves are averaged over multiple runs and grouped by initialization bias (\textit{neutral}: $\RClassFraction{0} \approx 0.5$, \textit{prejudiced}: $\RClassFraction{0} \approx 1$). Top: global accuracy; bottom: per-class accuracy. See App.~\ref{sec:reprod} for model \MLPA{} and dataset \GB{} details. Learning rate: \texttt{1e-5}, batch size: \texttt{512}.
    }
    \label{fig:gb_1hl_train}
\end{figure*}

\begin{figure*}[h]
    \centering
    \includegraphics[width=0.45\textwidth]{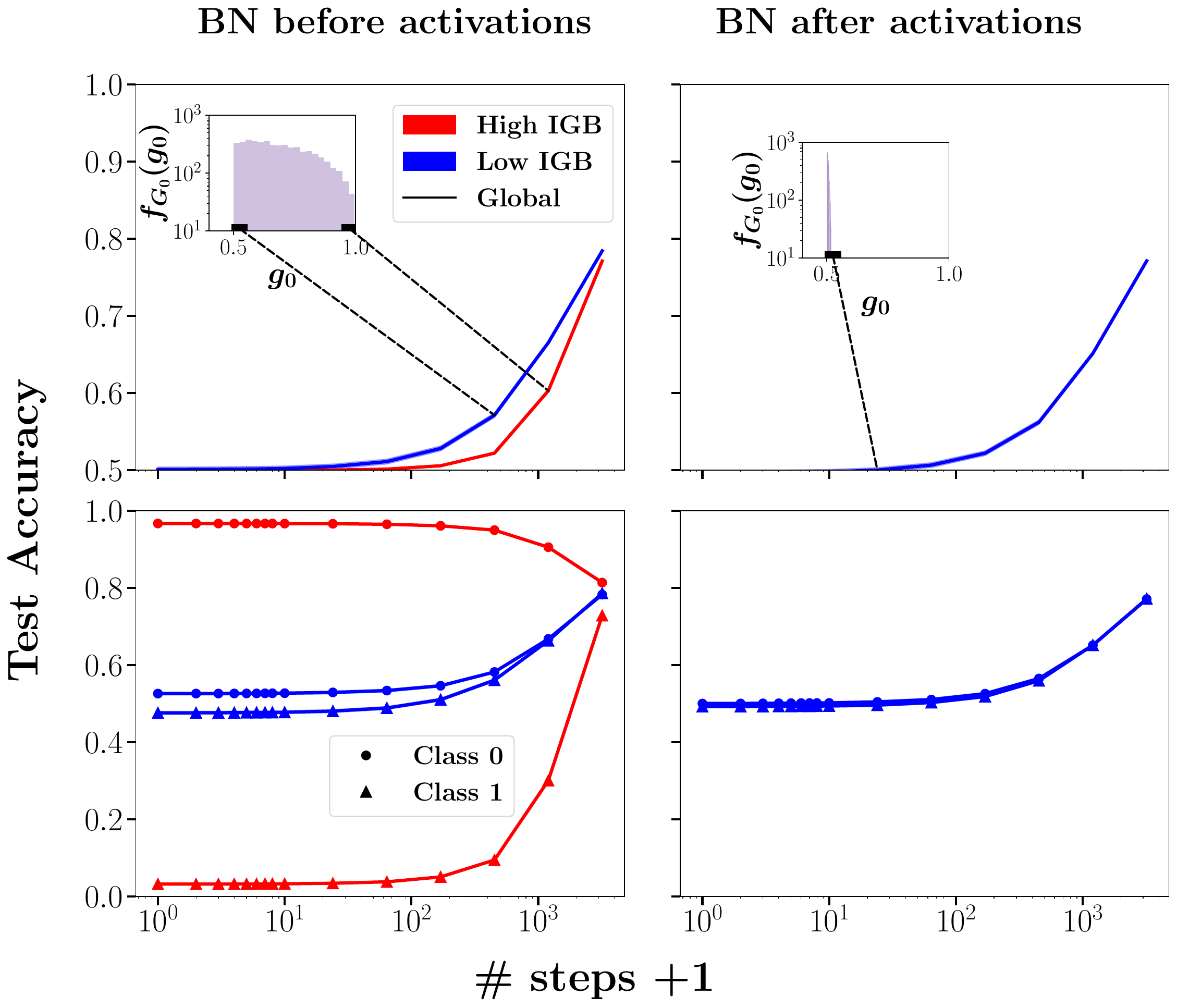}
    \includegraphics[width=0.45\textwidth]{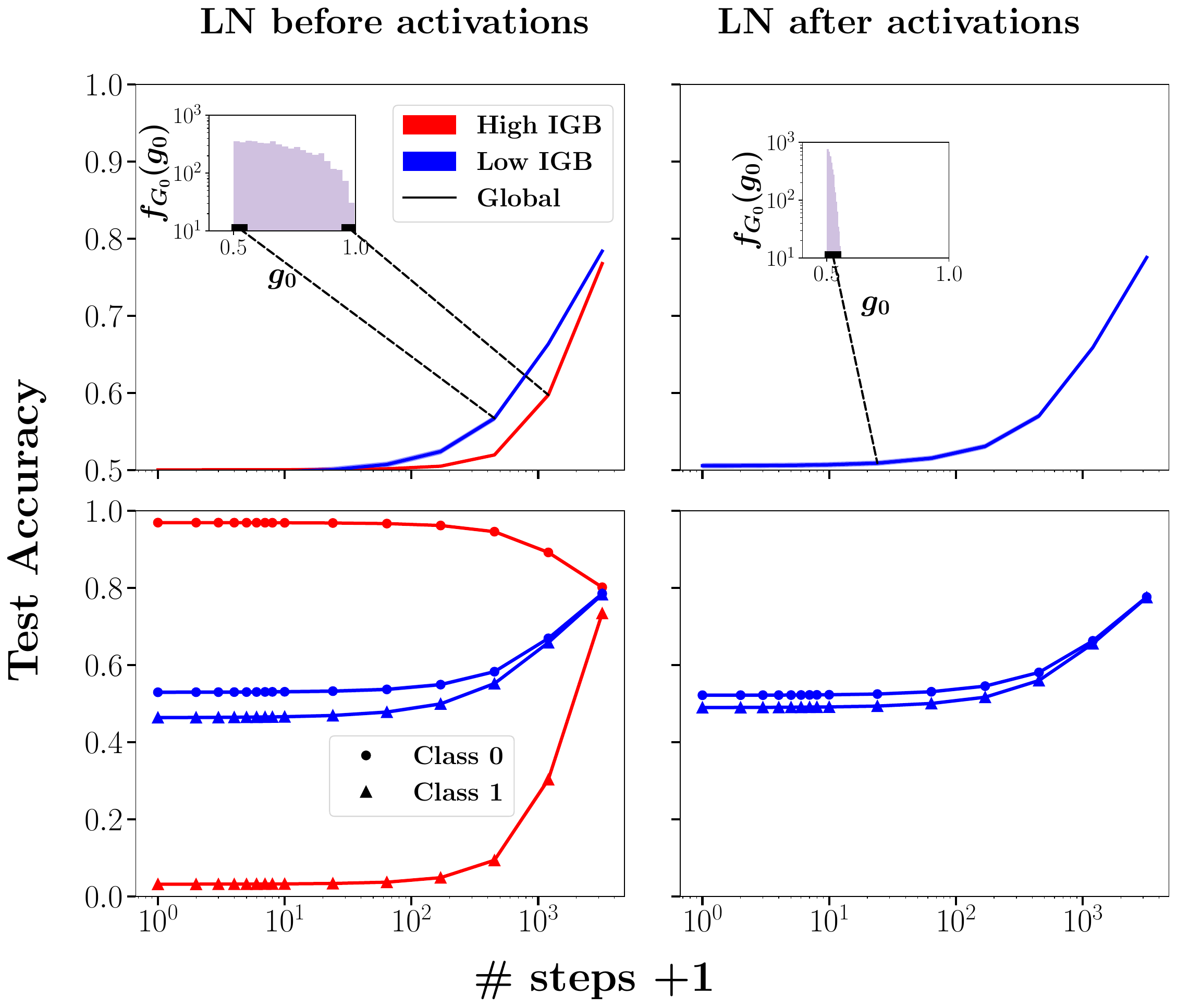}
    \caption{
    Test accuracy on \GB{} using the shallow \MLPA{} model. Left: BN; Right: LN. Each panel compares normalization applied \textit{before} vs \textit{after} ReLU. Top: global accuracy; bottom: per-class accuracy. Curves are grouped by initialization bias. See App.~\ref{sec:reprod} for model \MLPA{} and dataset \GB{} details. Learning rate: \texttt{1e-5}, batch size: \texttt{512}.
    }
    \label{fig:gb_1hl_test}
\end{figure*}

\vspace{0.3em}
\noindent\textbf{Gaussian Blob – random data.}  
We begin with \GB{}, a high-dimensional synthetic dataset without meaningful correlations. Results (Figs.~\ref{fig:gb_1hl_train},~\ref{fig:gb_1hl_test}) replicate theoretical predictions: normalization placement critically determines the level of IGB, which in turn shapes both convergence speed and per-class imbalance. Models with normalization after ReLU exhibit faster reabsorption of initial bias and improved convergence, especially under LN.

\begin{figure*}[h]
    \centering
    \includegraphics[width=0.45\textwidth]{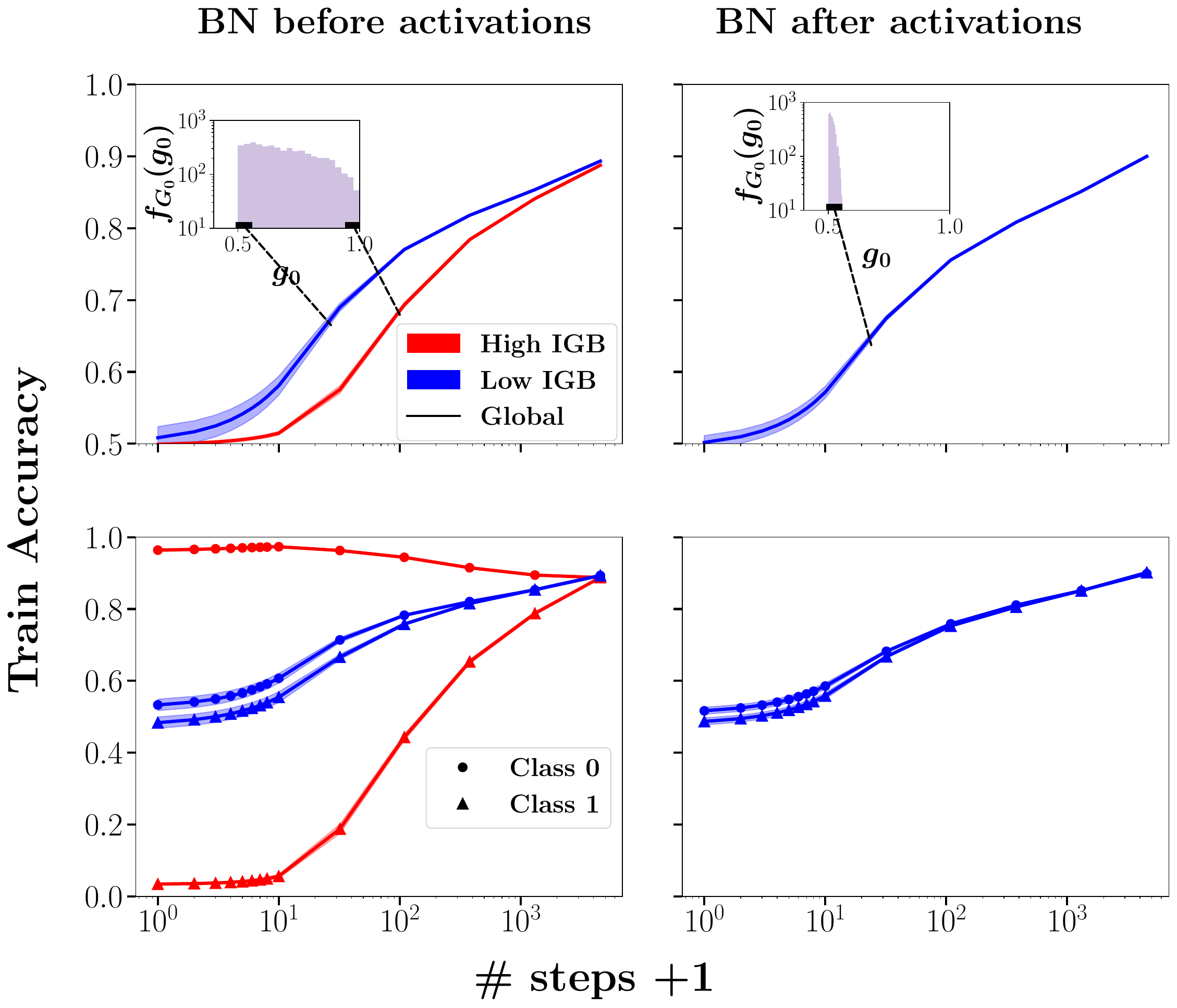}
    \includegraphics[width=0.45\textwidth]{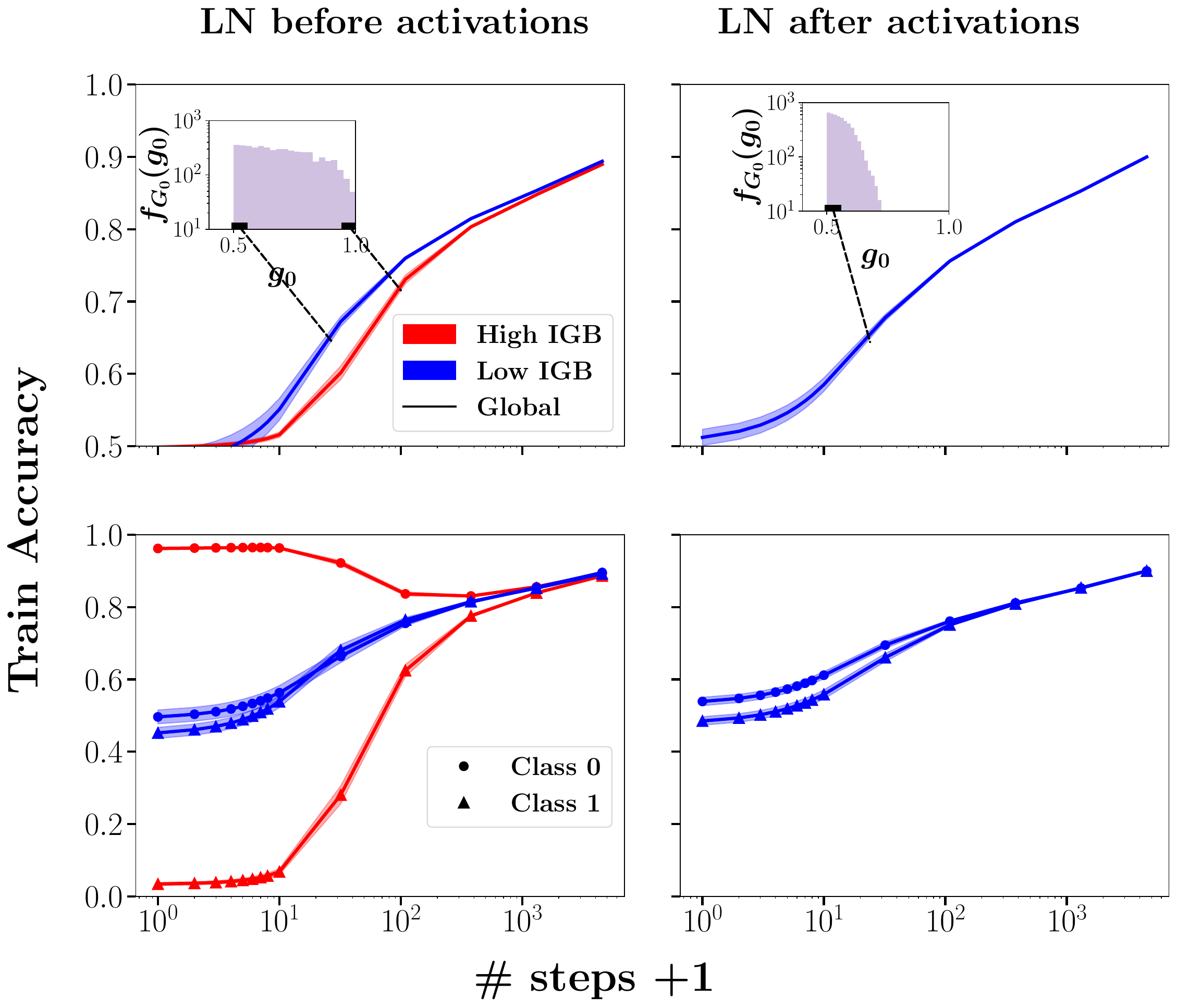}
    \caption{
    Training accuracy on \VehVsAni{} using the shallow \MLPA{} model. Left: BN; Right: LN. Each panel compares normalization \textit{before} vs \textit{after} ReLU. Top: global accuracy; bottom: per-class accuracy. See App.~\ref{sec:reprod} for model \MLPA{} and dataset \VehVsAni{} details. Learning rate: \texttt{1e-5}, batch size: \texttt{512}.
    }
    \label{fig:cifar_1hl_train}
\end{figure*}

\begin{figure*}[h]
    \centering
    \includegraphics[width=0.45\textwidth]{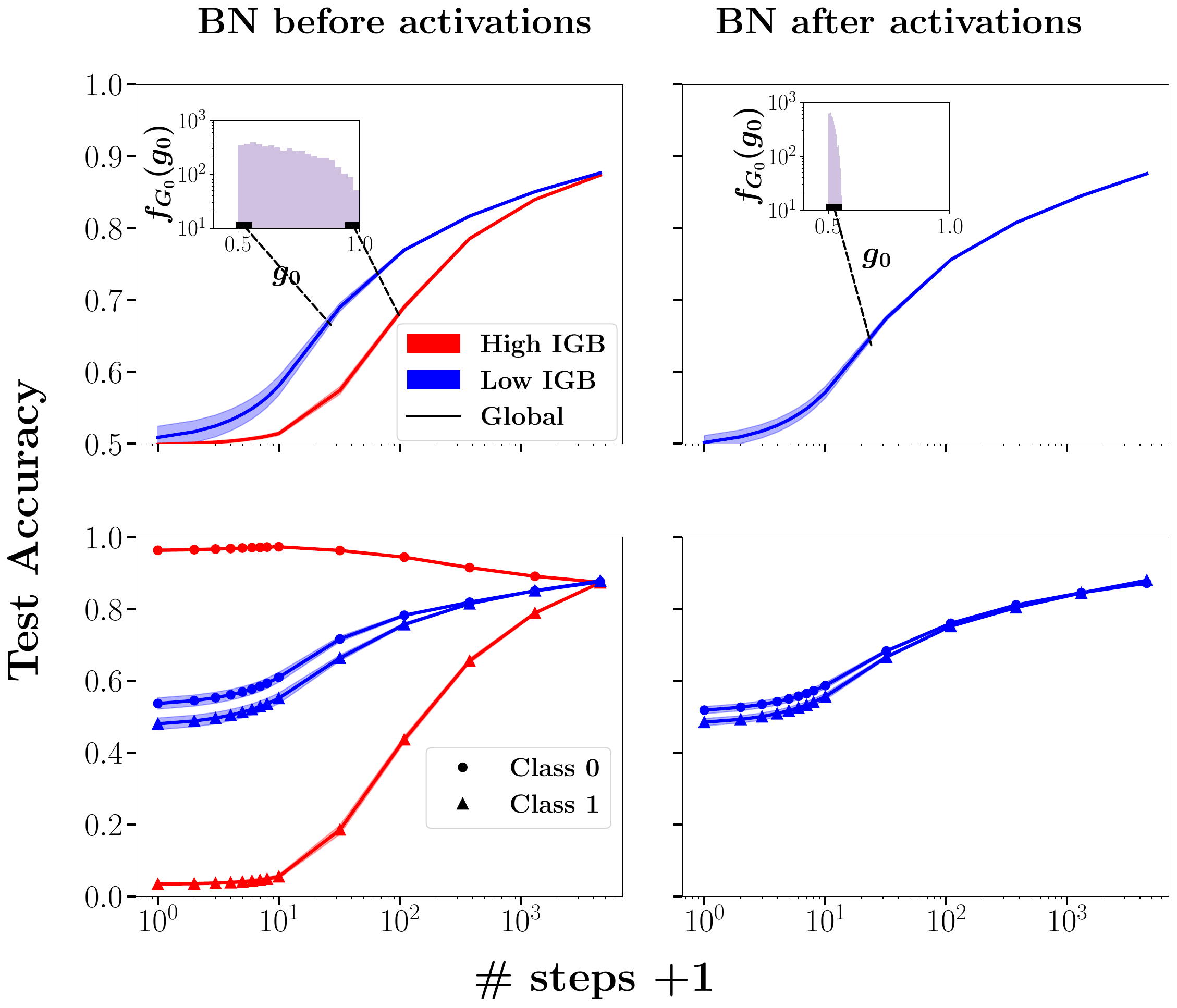}
    \includegraphics[width=0.45\textwidth]{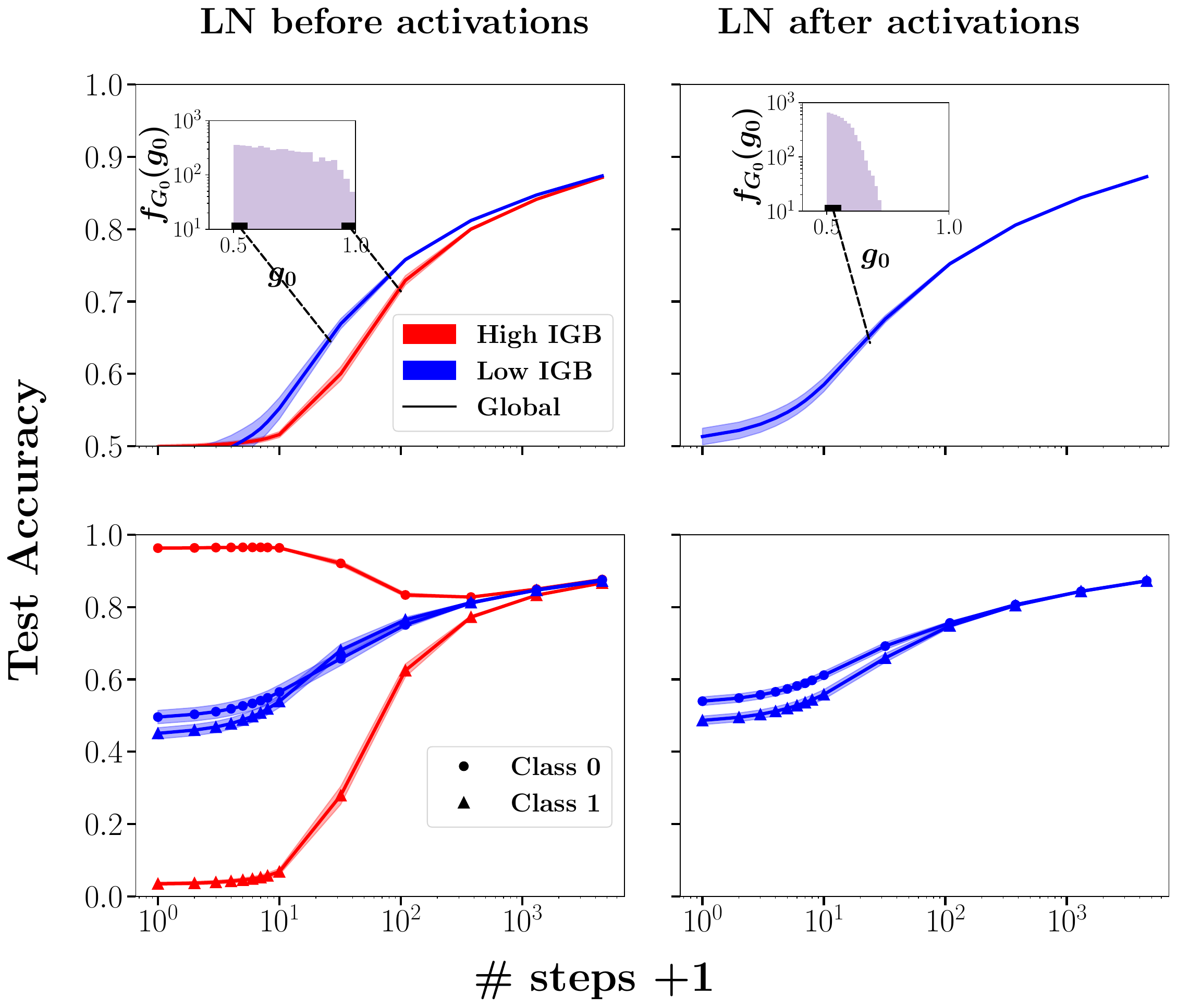}
    \caption{
    Test accuracy on \VehVsAni{} using the shallow \MLPA{} model. Left: BN; Right: LN. Global (top) and per-class (bottom) performance trends mirror those observed on random data. See App.~\ref{sec:reprod} for model \MLPA{} and dataset \VehVsAni{} details. Learning rate: \texttt{1e-5}, batch size: \texttt{512}.
    }
    \label{fig:cifar_1hl_test}
\end{figure*}

\vspace{0.3em}
\noindent\textbf{CIFAR10 – empirical evaluation on structured inputs.}  
Next, we evaluate \VehVsAni{}, a binary subset of CIFAR10. Despite being a structured real dataset, results (Figs.~\ref{fig:cifar_1hl_train},~\ref{fig:cifar_1hl_test}) remain qualitatively consistent with those on \GB{}. This suggests that our theory captures dominant effects even outside the regime of its assumptions. Normalization placement still controls IGB level, and bias-driven convergence effects remain visible.

\begin{figure*}[h]
    \centering
    \includegraphics[width=0.45\textwidth]{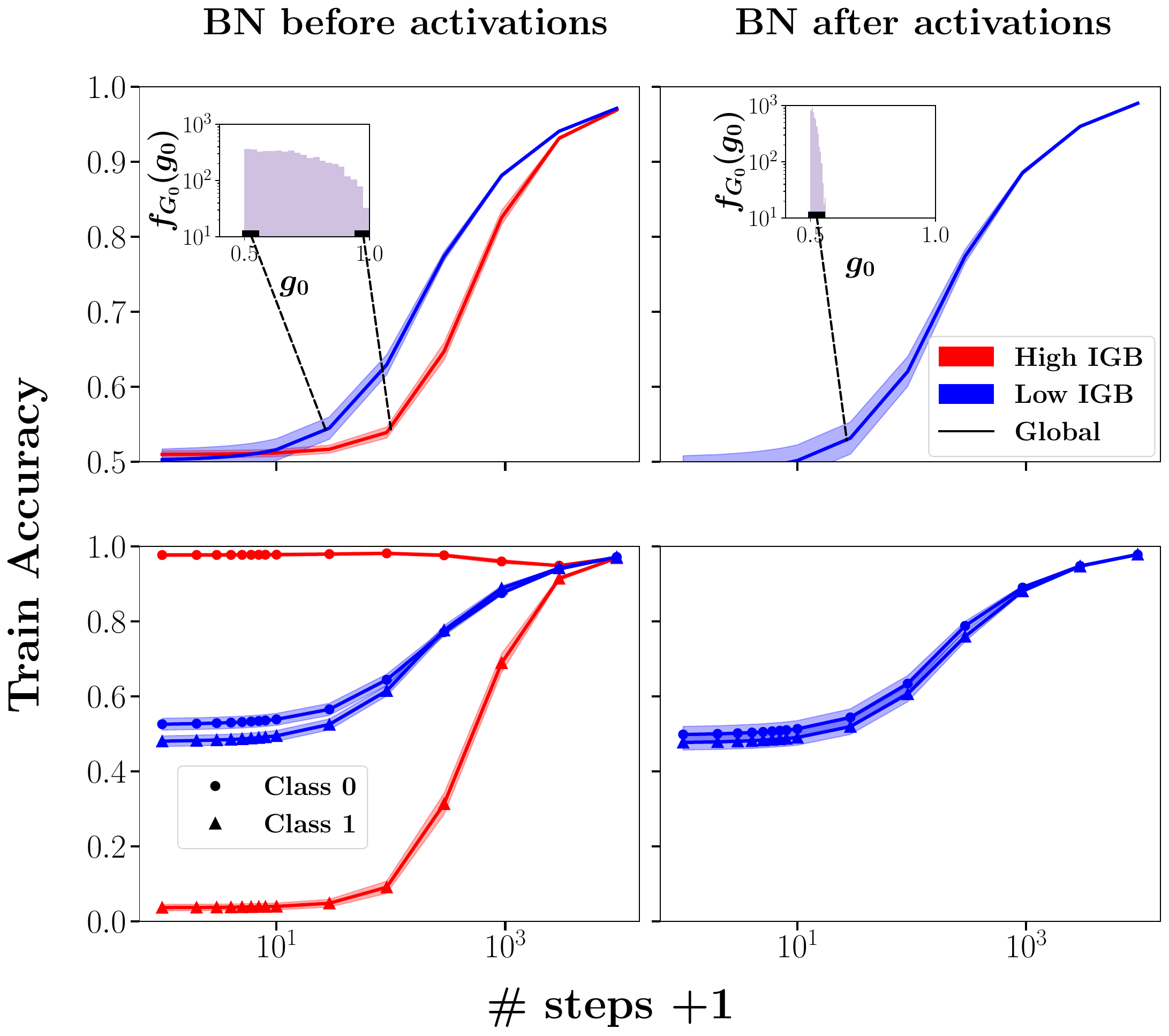}
    \includegraphics[width=0.45\textwidth]{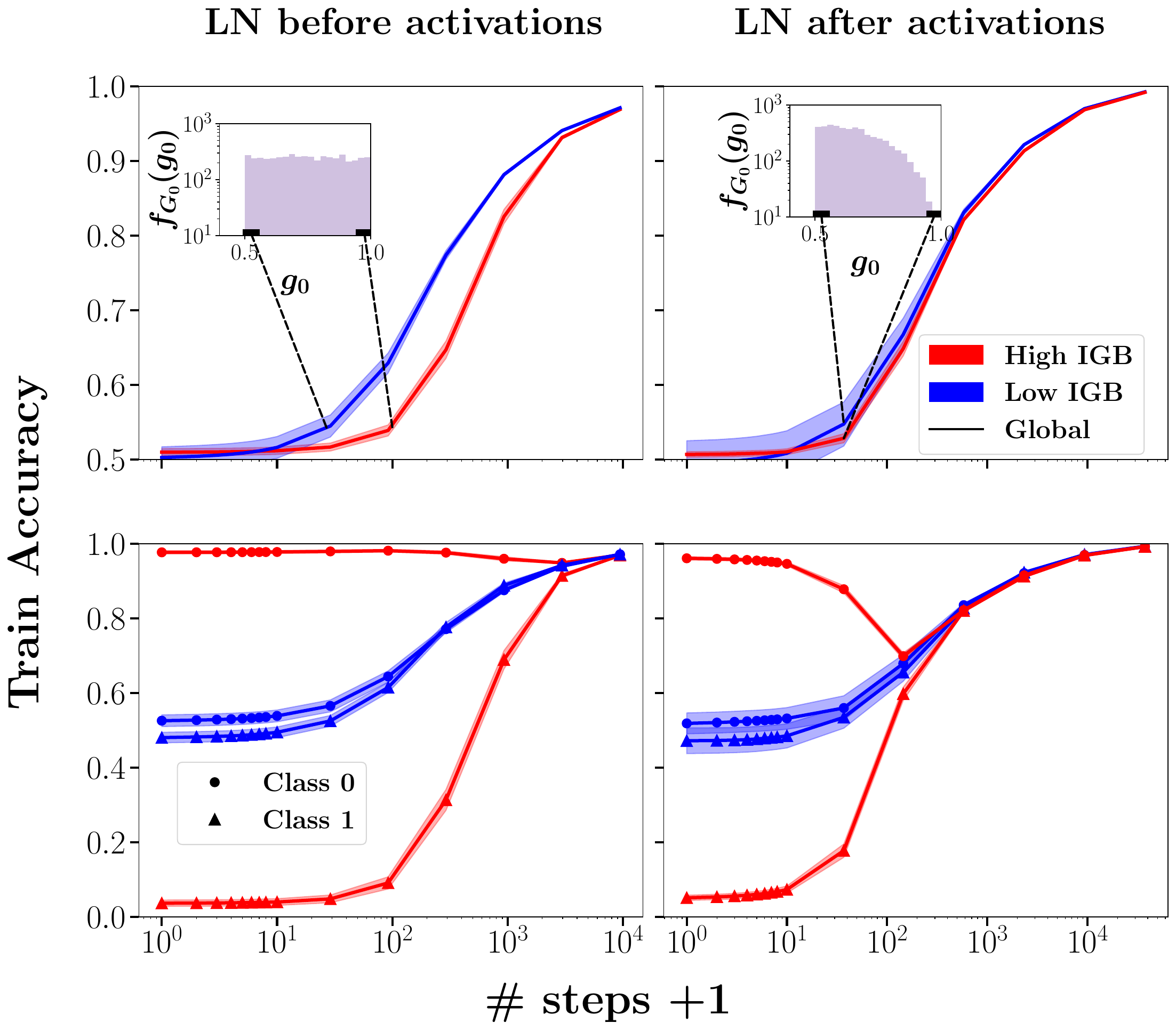}
    \caption{
    Training accuracy on \MNIST{} with the shallow \MLPA{} model. Left: BN; Right: LN. LN-after-ReLU reduces IGB (as can be seen from the insets), though it does not eliminate it. Top: global accuracy; bottom: per-class accuracy. See App.~\ref{sec:reprod} for model \MLPA{} and dataset \MNIST{} details. Learning rate: \texttt{1e-5}, batch size: \texttt{512}.
    }
    \label{fig:mnist_1hl_train}
\end{figure*}

\begin{figure*}[h]
    \centering
    \includegraphics[width=0.45\textwidth]{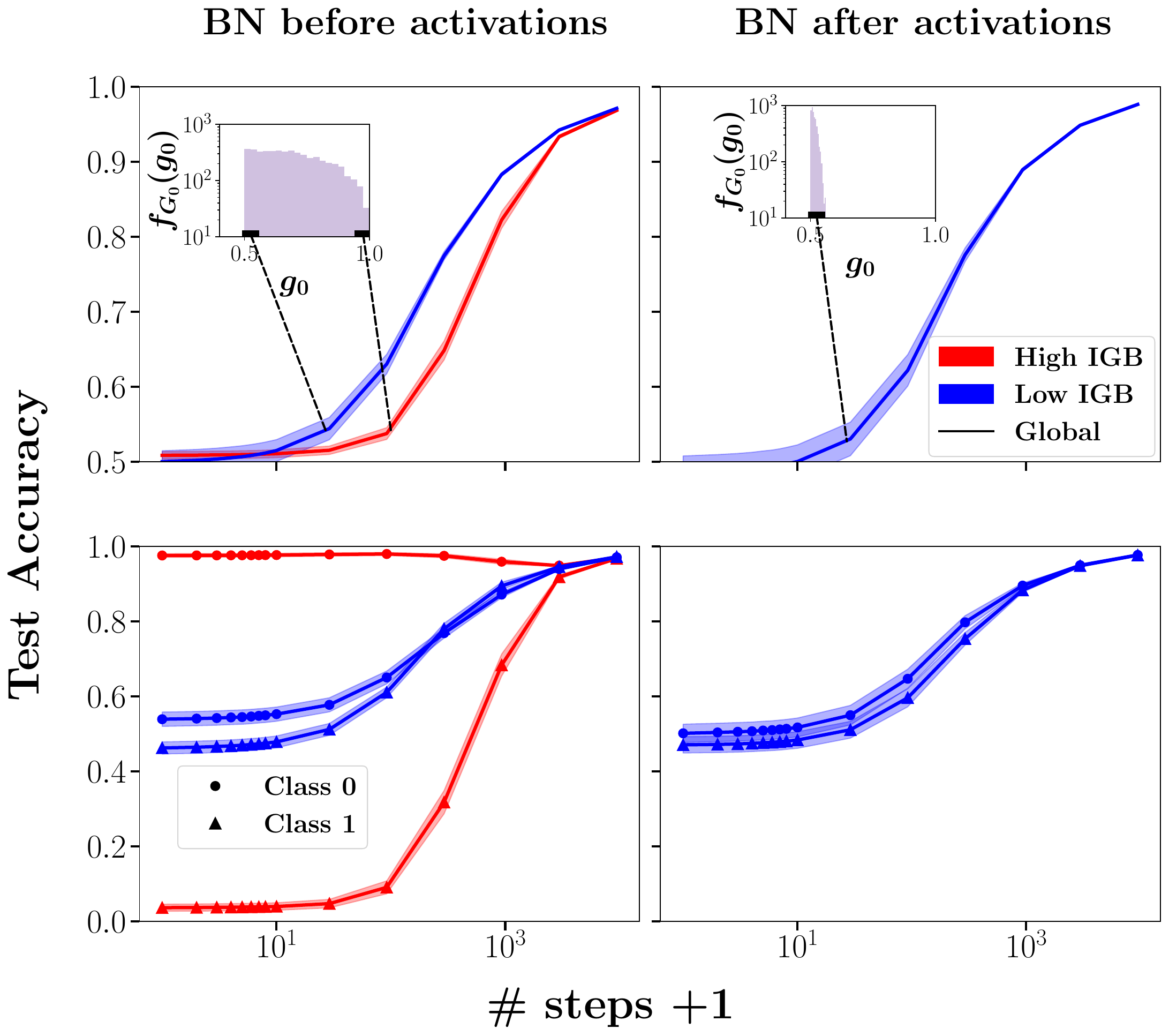}
    \includegraphics[width=0.45\textwidth]{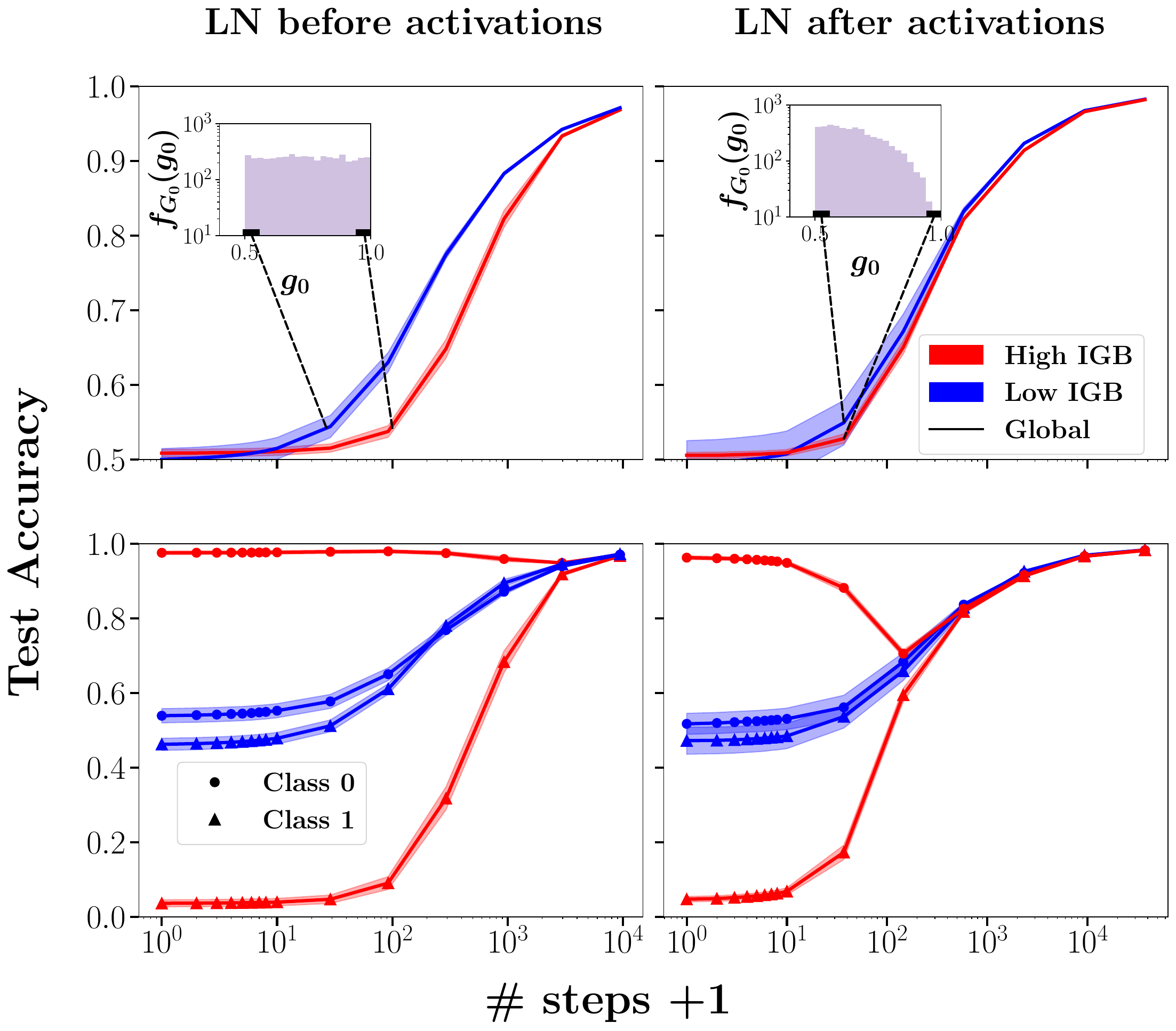}
    \caption{
    Test accuracy on \MNIST{} with the shallow \MLPA{} model. Left: BN; Right: LN. LN-after-ReLU results in broader bias distributions and persistent asymmetries. See App.~\ref{sec:reprod} for model \MLPA{} and dataset \MNIST{} details. Learning rate: \texttt{1e-5}, batch size: \texttt{512}.
    }
    \label{fig:mnist_1hl_test}
\end{figure*}

\vspace{0.3em}
\noindent\textbf{MNIST – highly correlated data.}  
To explore the effect of data structure more explicitly, we include experiments on \MNIST{}. We select MNIST as a prototype of strongly correlated inputs, since prior work on IGB~\citep{pmlr-v235-francazi24a} argued that datasets, where the correlations between different instances are strong (such as MNIST, where pixels at the corners are always white), should exhibit more IGB than what is predicted by a theory based on random uncorrelated inputs. Here (Figs.~\ref{fig:mnist_1hl_train},~\ref{fig:mnist_1hl_test}), we observe a qualitative departure: LN-after-ReLU reduces IGB without completely eliminating it—contrary to the behavior on \GB{} and \VehVsAni{}—as predicted by the arguments in Ref.~\citep{pmlr-v235-francazi24a}. As we can see by comparing the inset of LN plots in Figs.~\ref{fig:mnist_1hl_train},~\ref{fig:mnist_1hl_test} with the corresponding insets in Figs.~\ref{fig:gb_1hl_train},~\ref{fig:gb_1hl_test}, the distribution profiles differ in detail; nevertheless, the behaviour induced by the normalization placement remains consistent with our theory: placing the normalization \emph{before} the activation leads to more biased behaviour than placing it \emph{after}. \newline
Interestingly, BN-after-ReLU continues to exhibit near-neutral initial states, consistent with our theory and highlighting BN’s stronger ability to reduce IGB, as our theorems also indicate. We stress that one does not necessarily always want to completely eliminate IGB~\citep{bassi2025normplacement}, so the optimal configuration depends on the details of the problem. Nevertheless, our analysis provides a clearer understanding of the mechanisms underlying the initial bias and identifies practical design levers—most notably the placement of normalization relative to activation—that allow one to tune and shape the initial predictive behaviour through principled network design, rather than merely mitigate it.

\begin{figure*}[h]
    \centering
    \includegraphics[width=0.45\textwidth]{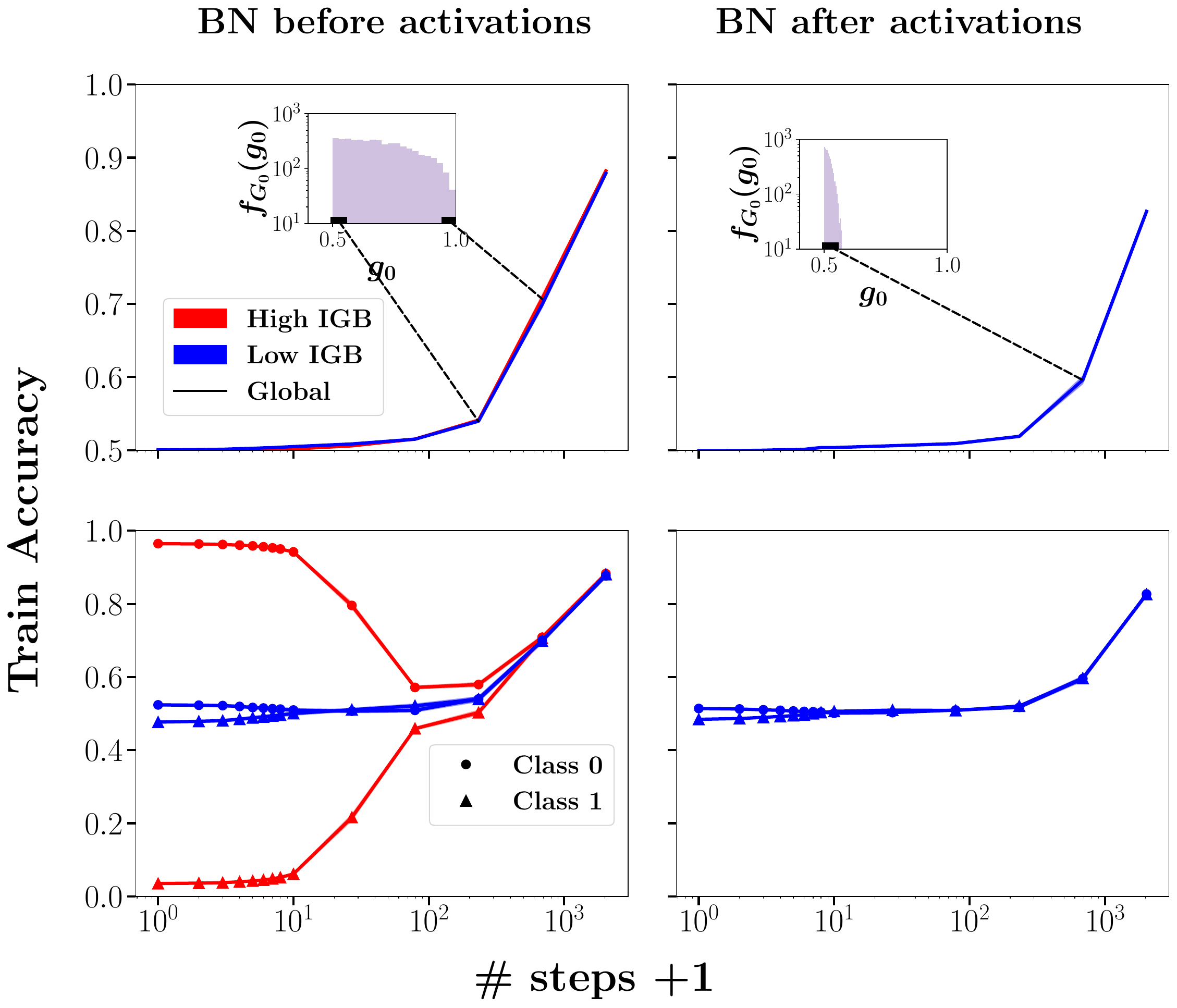}
    \includegraphics[width=0.45\textwidth]{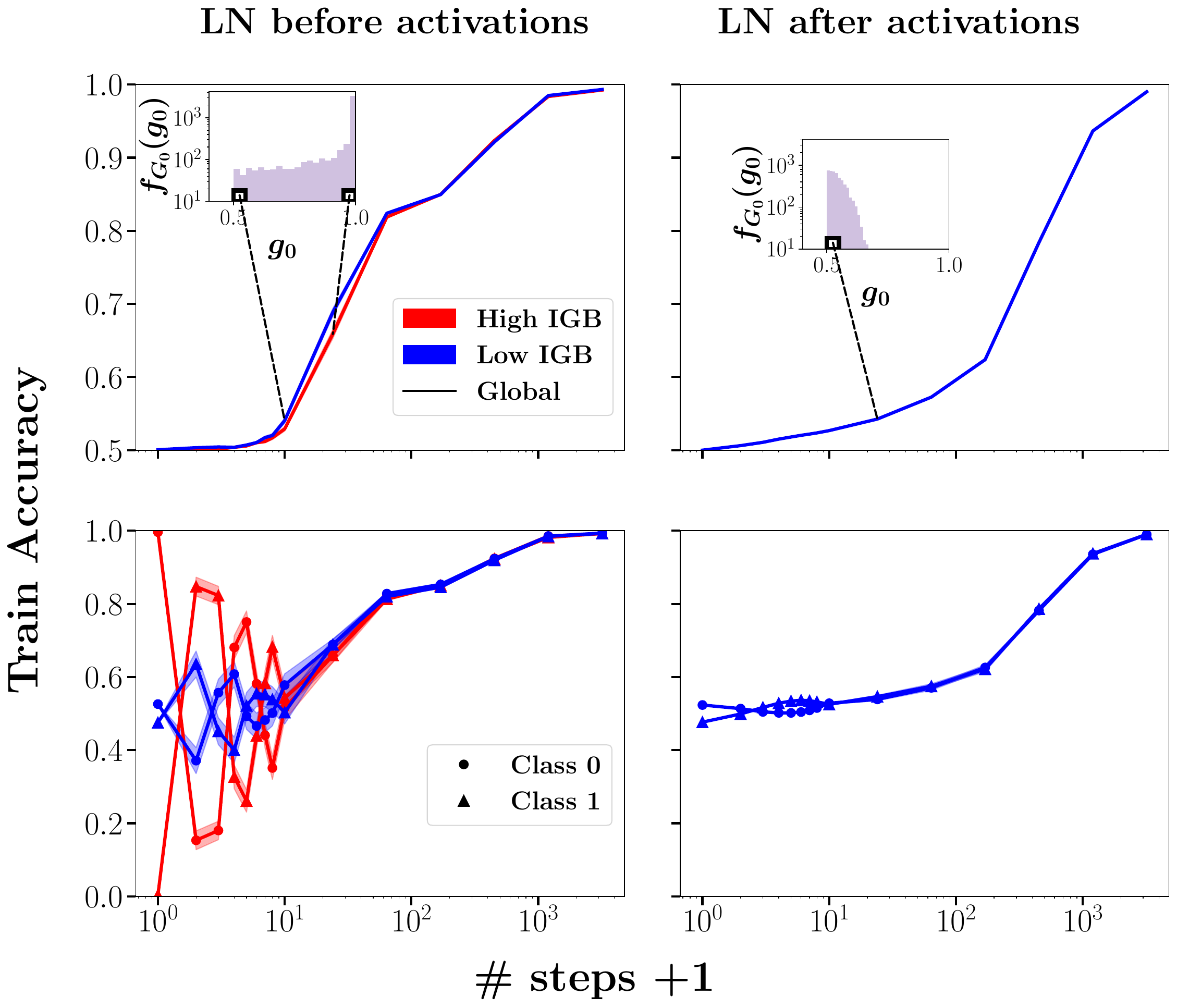}
    \caption{
    Training accuracy on \GB{} using the deep \MLPB{} model. Left: BN; Right: LN. Per-class accuracy reflects IGB differences, but global convergence speeds remain similar. See App.~\ref{sec:reprod} for model \MLPB{} and dataset \GB{} details. Learning rate: \texttt{1e-3} for LN, \texttt{5e-4} for BN. Batch size: \texttt{512}.
    }
    \label{fig:gb_20hl_train}
\end{figure*}

\begin{figure*}[h]
    \centering
    \includegraphics[width=0.45\textwidth]{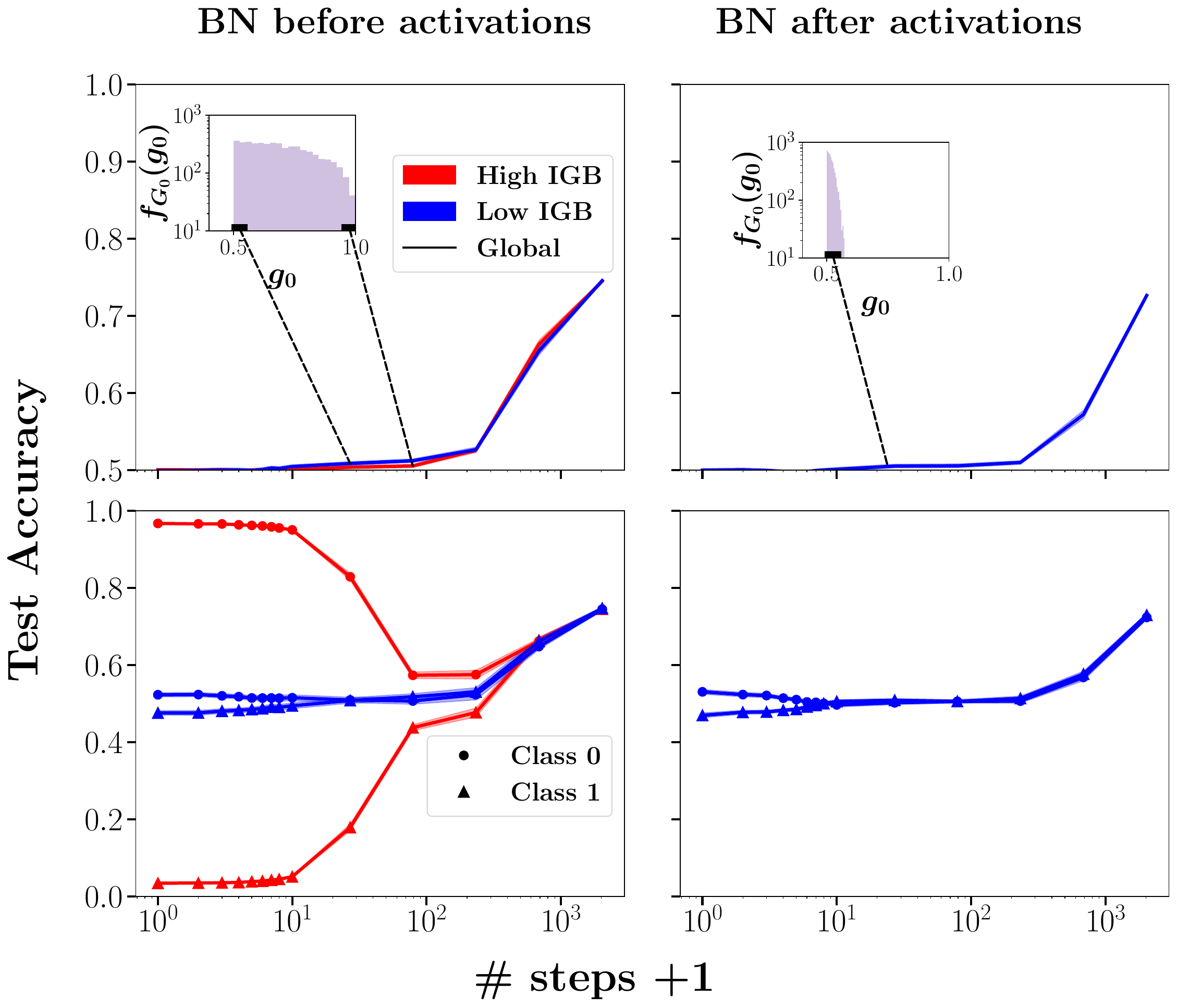}
    \includegraphics[width=0.45\textwidth]{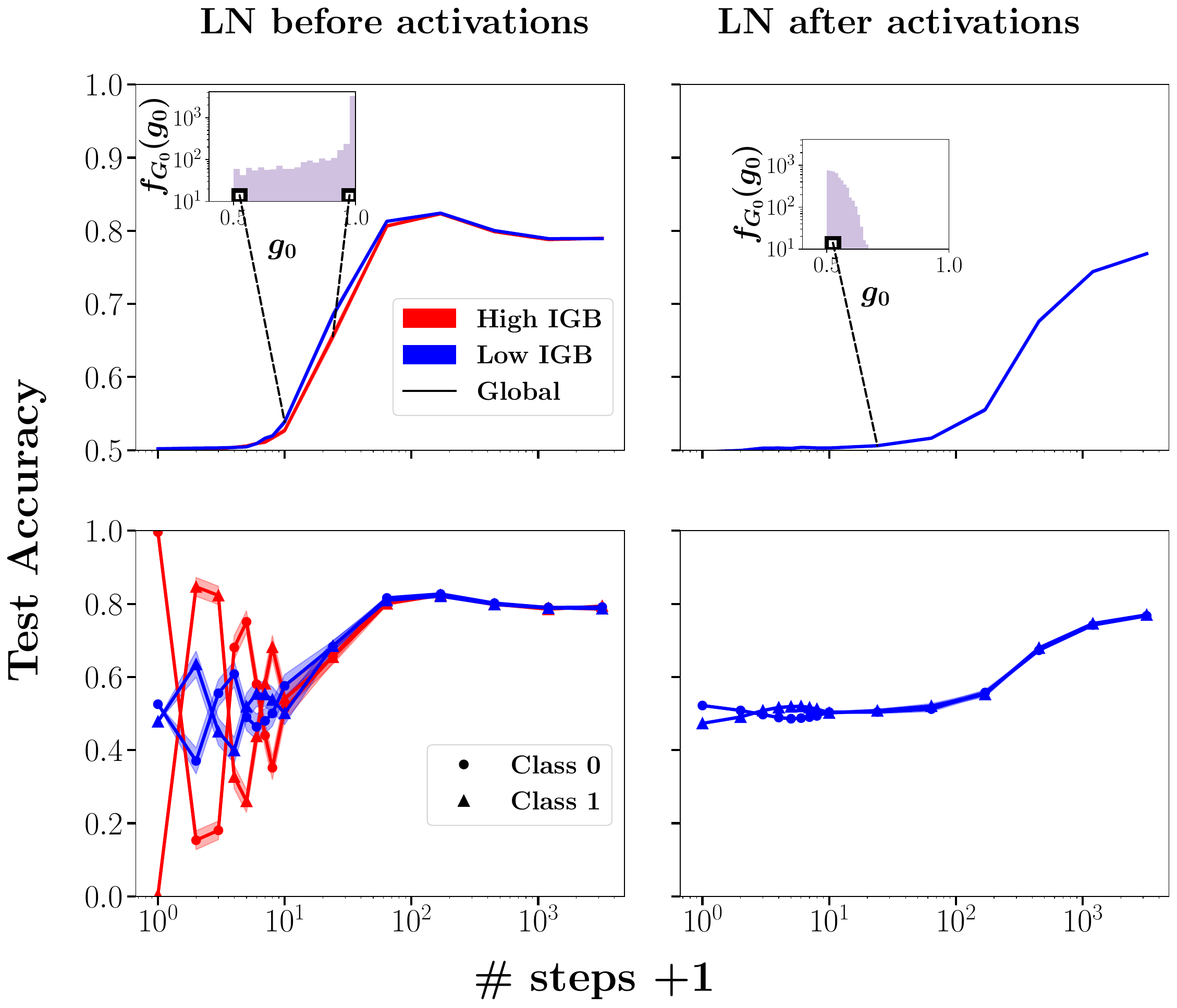}
    \caption{
    Test accuracy on \GB{} with the deep \MLPB{} model. Left: BN; Right: LN. Despite clear differences in per-class behavior, global accuracy curves largely overlap. See App.~\ref{sec:reprod} for model \MLPB{} and dataset \GB{} details. Learning rate: \texttt{1e-3} for LN, \texttt{5e-4} for BN. Batch size: \texttt{512}.
    }
    \label{fig:gb_20hl_test}
\end{figure*}

\vspace{0.3em}
\noindent\textbf{Robustness: Deep MLPs.}  
Finally, we repeat the \GB{} experiments on a deep \MLPB{} architecture to test the robustness of our findings (Figs.~\ref{fig:gb_20hl_train},~\ref{fig:gb_20hl_test}). While per-class accuracy still reflects IGB differences, global accuracy curves largely overlap across normalization placements. This suggests that, in deep models, early-stage bias may still shape local dynamics but not necessarily affect global convergence speed.

\subsection{Experiments on Other Architectures}\label{app:exp_Other_arch}
While our theory focuses on MLPs, we now investigate how its insights transfer to more complex, widely used architectures. A comprehensive study across all architectures and datasets is outside the scope of this paper; instead, we provide complementary evidence in realistic settings. We report initialization results on large, multi-class settings (\ResNetHundredOne{}, \SwinL{}) and training-dynamics experiments on (\ResNet{}, \MLPmix{}). This choice of settings allows us to present results on advanced models and realistic tasks while avoiding massive compute requirements, thereby supporting reproducibility. Across all experiments, we rely on established implementations, using the documented tuned choices of hyperparameters as baselines; the only \emph{variant} introduced for comparison is the placement of the normalization layer.

\begin{figure*}[t!]
    \centering
    \includegraphics[width=0.4\textwidth]{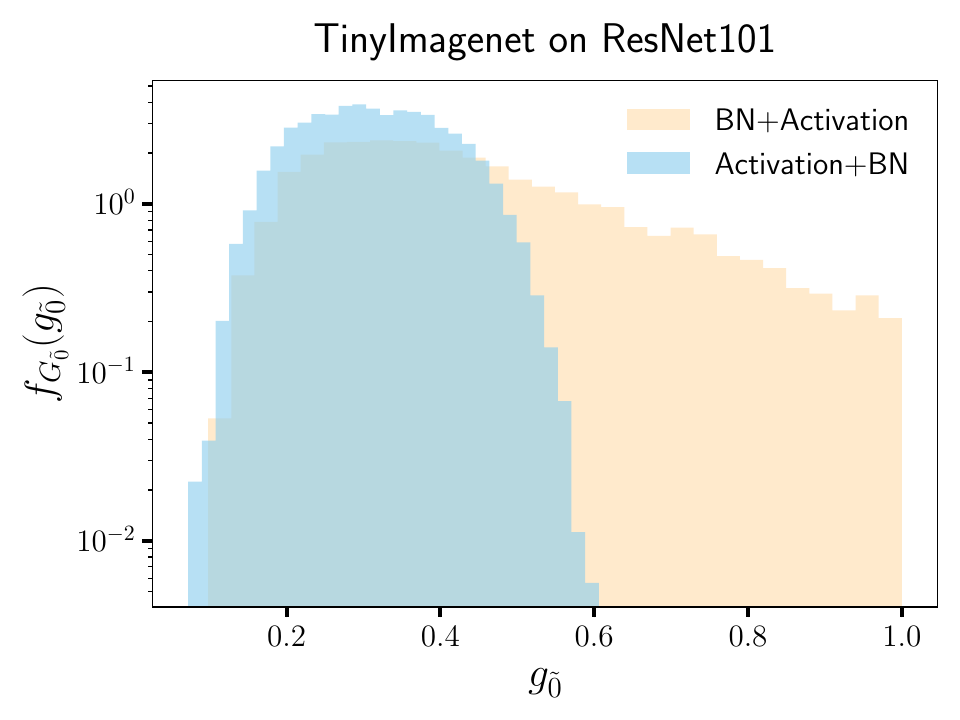}
    \includegraphics[width=0.4\textwidth]{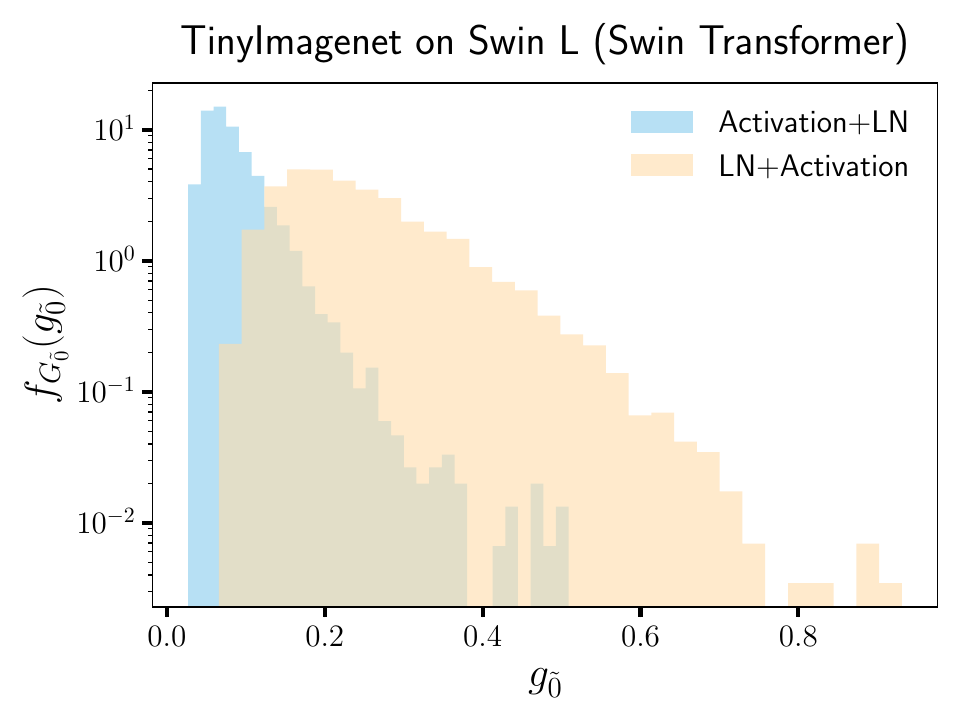}
    \caption{
    \textbf{Initialization bias at scale.} Distributions of the maximum per-class guess fraction $\RMultiClassRankedFraction{0}{}$ under baseline (\emph{norm + act.}) vs.\ \emph{act. + norm} for \ResNetHundredOne{} (left) and \SwinL{} (right) on high-cardinality datasets (\TinyImageNet). In both cases, moving normalization after the activation produces a right-shift, i.e., larger typical $\RMultiClassRankedFraction{0}{}$, indicating increased initial IGB. 
    }
    \label{fig:init_scale_resnet101_swinL}
\end{figure*}

\vspace{0.4em}
\noindent\textbf{Normalization placement and initialization bias.}
For regimes with many classes ($\NumberClasses$ large), the neutral fraction $1/\NumberClasses$ is small, making the maximum per-class guess fraction
$\RMultiClassRankedFraction{0}{}$ a sensitive summary of bias at initialization. We compare the distribution of $\RMultiClassRankedFraction{0}{}$ between the baseline configuration (\emph{normalization + activation}) and a variant with normalization moved \emph{after} the activation ((\emph{activation + normalization})), for \ResNetHundredOne{} and \SwinL{}. In both models, placing normalization after the activation yields a right-shift of the $\RMultiClassRankedFraction{0}{}$ distribution, indicating an increased typical fraction of datapoints absorbed by the most-guessed class at initialization. This shift signals a higher initial IGB level and is coherent with our theoretical expectations (Fig.~\ref{fig:init_scale_resnet101_swinL}; additional details in App.~\ref{sec:reprod}).

\begin{figure*}[t!]
    \centering
    \includegraphics[width=0.7\textwidth]{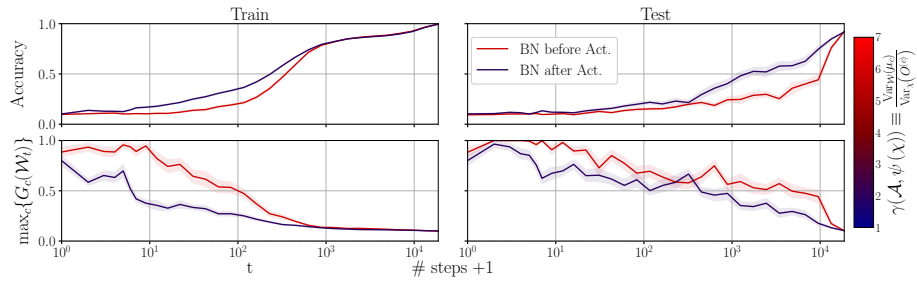}
    \caption{
    Training dynamics of \ResNet{} on \CIFAR{}, comparing the standard configuration with BN placed before the activation (BN + Act.) and a modified variant with BN applied after the activation (Act. + BN). The level of Initial Guessing Bias (IGB) is measured using the variance ratio \(\VarRatio{}\), computed from the network’s initial outputs. As predicted by our theoretical analysis, applying BN after the activation reduces \(\VarRatio{}\) in this setting, yielding a more balanced initial predictive state. Results are consistent with trends observed in MLP-based architectures.
    }
    \label{fig:ResNet_exp_app}
\end{figure*}

\begin{figure*}[t!]
    \centering
    \includegraphics[width=0.9\textwidth]{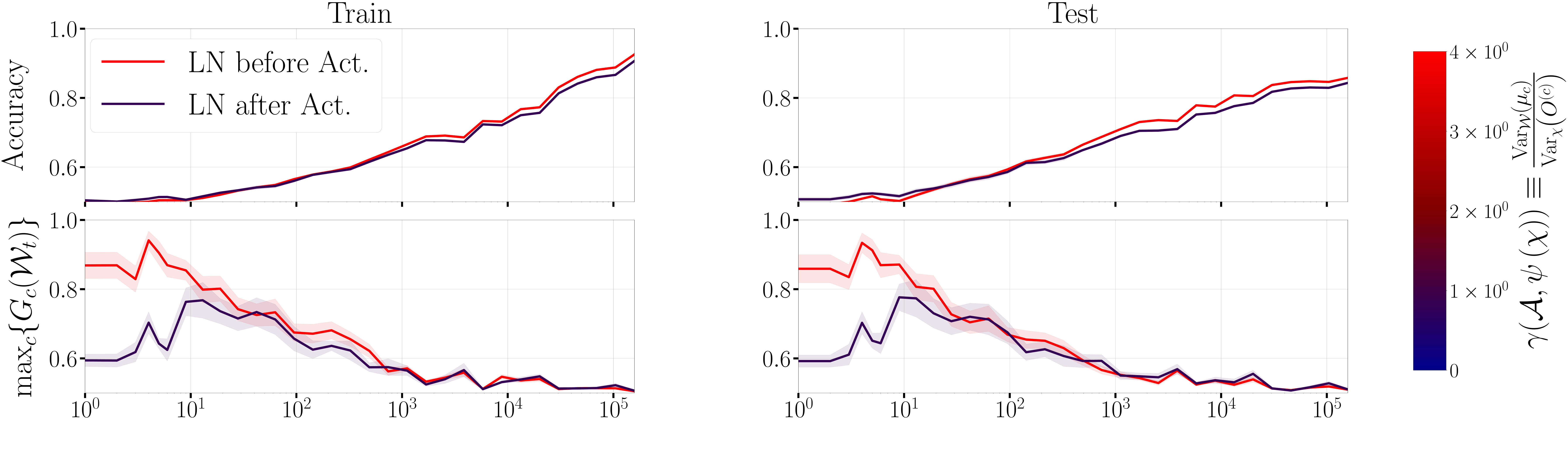}
    \includegraphics[width=0.9\textwidth]{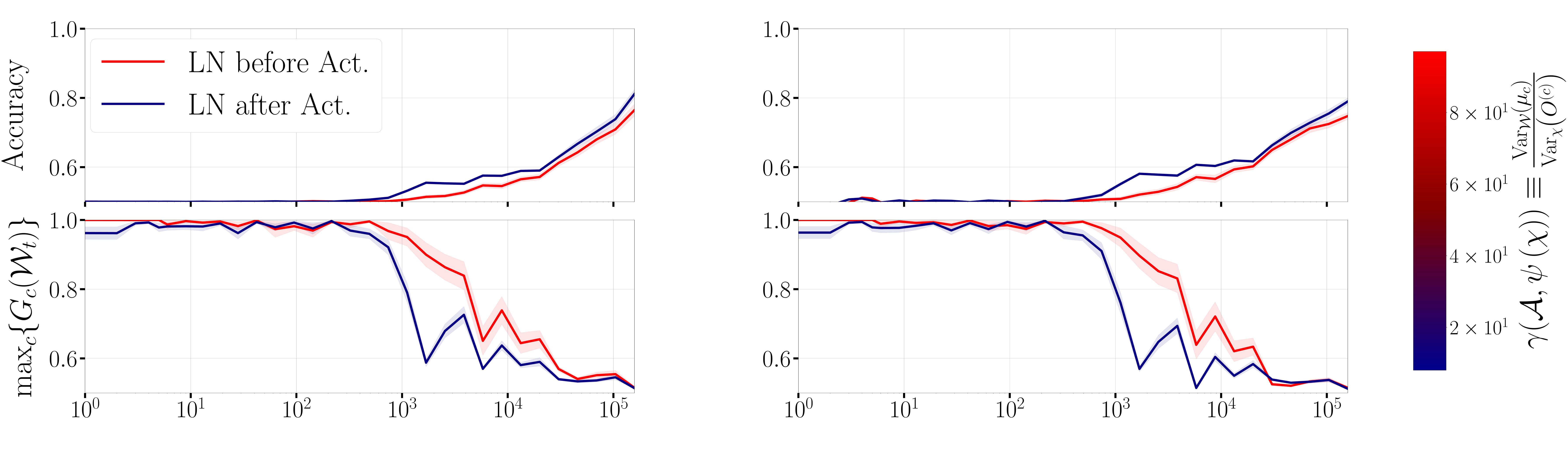}
    \includegraphics[width=0.9\textwidth]{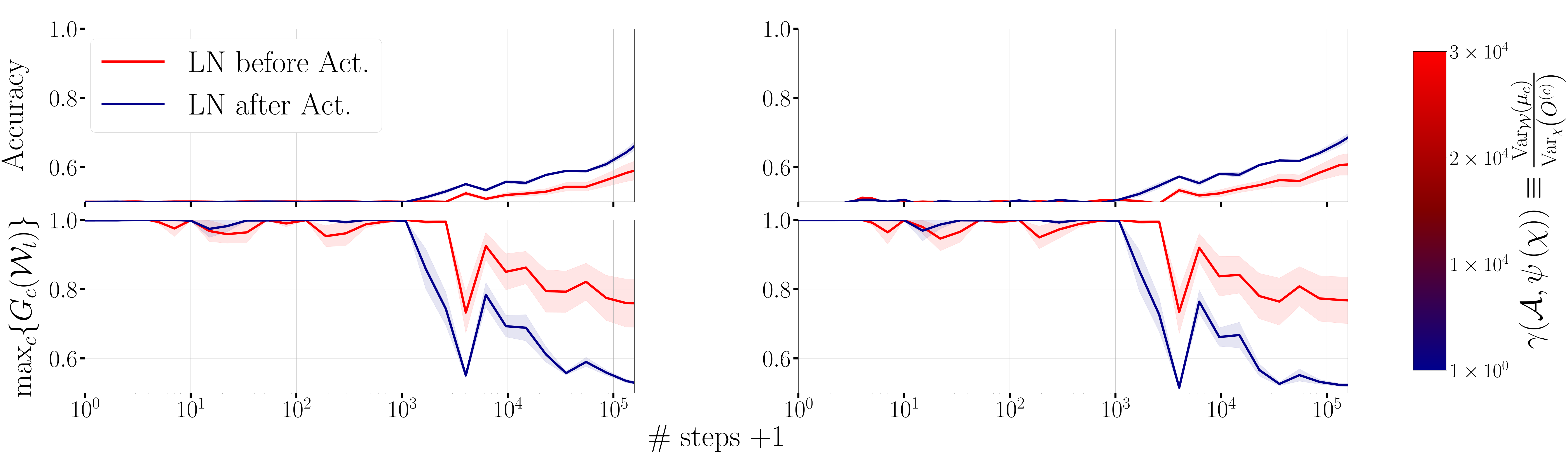}
    \includegraphics[width=0.9\textwidth]{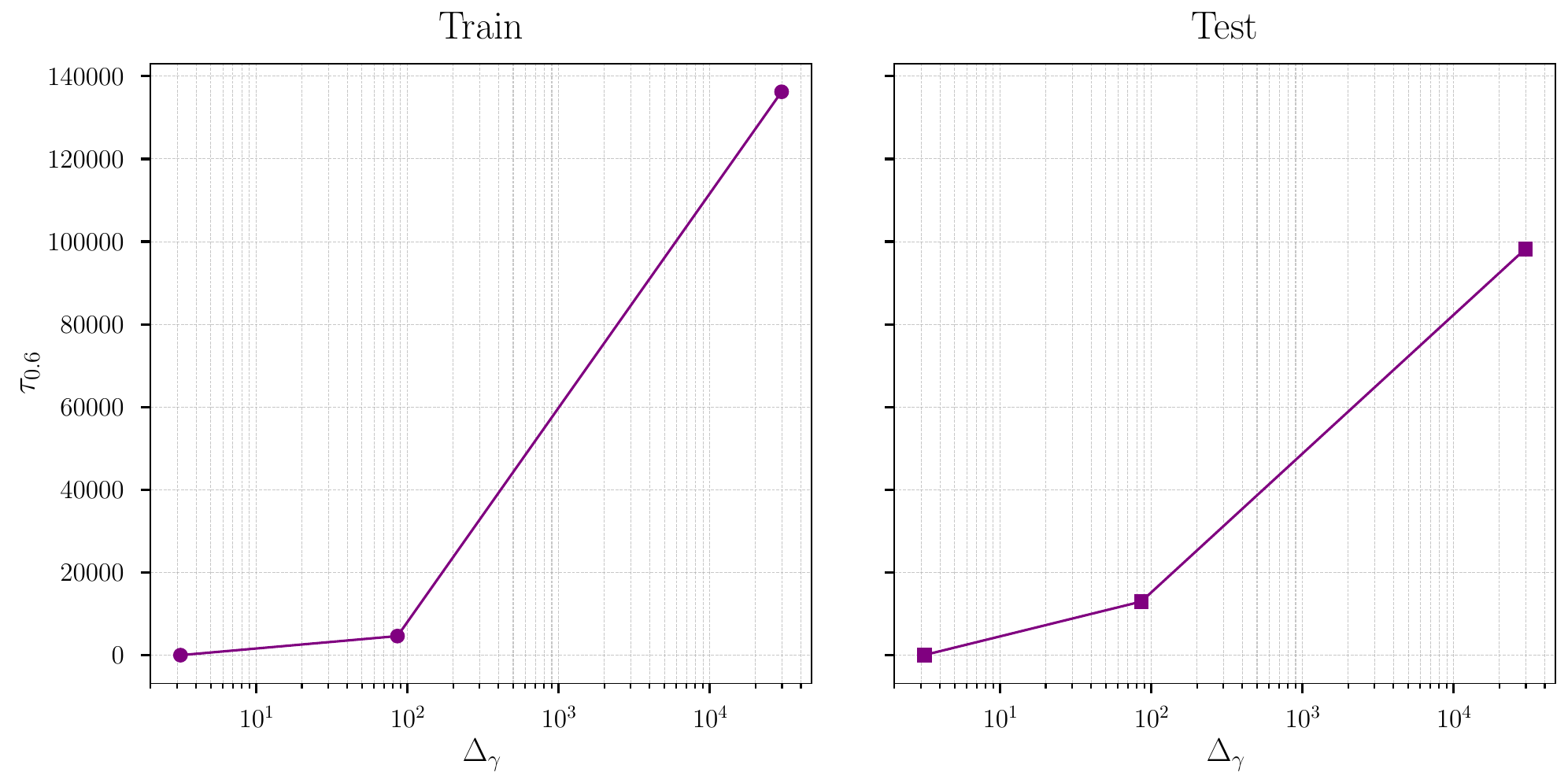}
    \caption{
    Learning dynamics of \MLPmix{} on \CatVsDog{} with increasing levels of IGB (top to bottom), induced by progressively shifting input distributions via standardization (adding a constant to all pixels). Each panel compares the canonical configuration with LN applied \textit{before} the activation (LN + Act.) to a variant with LN applied \textit{after} the activation (Act. + LN). For each setting, we report accuracy (left) and the maximum class prediction bias $\max_c \{ \RClassFraction{c} \}$ (right), averaged over multiple runs (mean $\pm$ standard error). Left column: training data; right column: test data. In the baseline setting (top), both configurations perform similarly, but as IGB increases, LN-before-activation results in slower convergence, consistent with theoretical predictions. Panel (d) summarizes the convergence delay ($\tau_{0.6}$) versus $\Delta_{\gamma}$, the difference in $\VarRatio$ between the two configurations.
    }
    \label{fig:dyn_MLP_mix}
\end{figure*}

\vspace{0.4em}
\noindent\textbf{Normalization placement and training dynamics.}
We next study the impact that the normalization placement has on the learning process, using \ResNet{} and \MLPmix{}.
 We emphasize that across all experiments the architecture and hyperparameters follow well-documented baselines; the only intervention on the model is the positioning of the normalization layer to define a variant for the comparison with the baseline.
To tune the IGB level without altering architectural details, we apply a simple pre-processing step to the input data, adding a constant value to all pixels, as prescribed in Ref.~\citep{pmlr-v235-francazi24a}. We compare the default setting with a variant in which normalization layers are moved \textit{after} the activation functions (see Figs.~\ref{fig:ResNet_exp_app},~\ref{fig:dyn_MLP_mix}).

Our experiments show consistent results with theoretical expectations. Moving normalization after the nonlinearity reduces IGB and accelerates convergence. For the MLP-Mixer (Fig.~\ref{fig:dyn_MLP_mix}), we further vary the IGB level by adjusting the data standardization procedure. We observe that as IGB increases, overall training slows in both configurations, but the gap in convergence speed between the two normalization placements widens. This suggests a compounding interaction between IGB strength and normalization structure, reinforcing the role of architectural choices in determining early dynamics.
All experiments on the dynamics are repeated over multiple random seeds; plots report the mean and standard error across runs to account for variability.

\clearpage

\end{document}